\documentclass{article} %
\pdfoutput=1
\usepackage{iclr2019_conference,times}

\usepackage{amsmath,amsfonts,bm}

\def\eqref#1{equation~\ref{#1}}

\def\1{\bm{1}}

\DeclareMathAlphabet{\mathsfit}{\encodingdefault}{\sfdefault}{m}{sl}
\SetMathAlphabet{\mathsfit}{bold}{\encodingdefault}{\sfdefault}{bx}{n}

\newcommand{\R}{\mathbb{R}}

\usepackage{hyperref}
\usepackage{url}
\usepackage{graphicx}
\usepackage{gensymb}
\usepackage{mathtools}
\usepackage{subfig}
\usepackage{booktabs}
\usepackage{float}
\usepackage{siunitx}
\usepackage{numprint}
\usepackage{wrapfig}
\usepackage{pdflscape} %

\graphicspath{{Figures/}}

\renewcommand{\S}{\mathcal{S}}
\newcommand{\D}{\mathcal{D}}

\newcommand{\xmark}{\text{\sffamily X}}

\title{Efficient Augmentation via Data Subsampling}

\author{%
Michael Kuchnik \& Virginia Smith \\
Carnegie Mellon University \\
\{mkuchnik,smithv\}@cmu.edu%
}

\renewcommand{\headrulewidth}{0pt} %
\newcommand{\customheader}{%
  \text{}%
}
\iclrfinalcopy%

\ificlrfinal%
\hypersetup{%
  pdftitle=Efficient Augmentation via Data Subsampling,
  pdfauthor=Michael Kuchnik \& Virginia Smith,
}
\fi

\begin{document}

\maketitle

\begin{abstract}
  Data augmentation is commonly used to encode invariances in learning methods.
  However, this process is often performed in an inefficient manner, as
  artificial examples are created by applying a number of transformations to
  all points in the training set. The resulting explosion of the
  dataset size can be an issue in terms of storage and training costs, as well
  as in selecting and tuning the optimal set of transformations to apply. In
  this work, we demonstrate that it is possible to significantly reduce the
  number of data points included in data augmentation while realizing the same
  accuracy and invariance benefits of augmenting the entire dataset.
  We propose a novel set of subsampling policies, based on model influence and
  loss, that can achieve a $90\%$ reduction in augmentation set size while
  maintaining the accuracy gains of standard data augmentation.
\end{abstract}

\section{Introduction}
Data augmentation is a process in which the training set is expanded by applying
class-preserving transformations, such as rotations or crops for images, to the
original data points. This process has become an instrumental tool in achieving
state-of-the-art accuracy in modern machine learning pipelines. Indeed, for
problems in image recognition, data augmentation is a key component in achieving
nearly all state-of-the-art results~\citep{cirecsan2010deep, dosovitskiy2016discriminative, graham2014fractional, sajjadi2016regularization}. Data augmentation is also a popular technique because of its simplicity, particularly in deep learning applications, where applying a set of known invariances to the data is often more straightforward than trying to encode this knowledge directly in the model architecture.

However, data augmentation can be an expensive process, as applying a number of transformations to the entire dataset may increase the overall size of the dataset by orders of magnitude.
For example, if applying just 3 sets of augmentations (e.g., translate, rotate, crop), each with 4 possible configurations, the dataset can easily grow by a factor of 12 (if applied independently), all the way to 64x (if applied in sequence). 
While this may have some benefits in terms of overfitting, augmenting the entire training set can also significantly increase data storage costs and training time, which can scale linearly or superlinearly with respect to the training set size. 
Further, selecting the optimal set of transformations to apply to a given data point is often a non-trivial task. Applying transformations not only takes processing time, but also frequently requires some amount of domain expertise.
Augmentations are often applied heuristically in practice, and small perturbations are
expected (but not proven) to preserve classes.
If more complex augmentations are applied to a dataset, they may have to be
verified on a per-sample basis.

In this work, we aim to make data augmentation more efficient and user-friendly by identifying subsamples of the full dataset that are good candidates for augmentation.
In developing policies for subsampling the data, we draw inspiration from the virtual support vector (VSV) method, which has been used for this purpose in the
context of SVMs~\citep{burges1997improving, decoste2002training}. The VSV method attempts to create a more robust decision surface by augmenting
only the samples that are close to the margin---i.e., the support vectors.
The motivation is intuitive: if a point does not affect the margin, then any
small perturbation of that point in data space will likely yield a point that is
again too far from the margin to affect it. The method proceeds by applying class-preserving data augmentations (e.g., small
perturbations) to all support vectors in the training set.
The SVM is then retrained on the support vector dataset concatenated with the
augmented dataset, and the end result is a decision surface that has been
encoded with transformation invariance
while augmenting many fewer samples than found in the full training set.

Although proven to be an effective approach for SVMs, methods utilizing support
vectors may not
generalize well to other classifiers.
Therefore, in this work, we aim to develop policies that can effectively reduce the augmentation set size while applying to a much broader class of models.
A key step in developing these policies is to determine some metric by which to rank the importance of data points for augmentation.
We build policies based on two key metrics.
First, we make a natural generalization of the VSV method by measuring the loss induced by a training point.
Second, we explore using the \textit{influence} of a point as an
indicator of augmentation potential.
Influence functions, originating from robust statistics, utilize more
information than loss (i.e., residuals) alone, as they take into account both leverage and residual information.

The contributions of this paper are as follows. First, we demonstrate that it is typically
unnecessary to augment the entire dataset to achieve high accuracy---for example, we can
maintain $99.86\%$ or more of the full augmentation accuracy while only augmenting $10\%$ of the dataset in the case
of translation augmentations, and we observe similar behavior for other
augmentations.
Second, we propose several policies to select the subset of points to augment.
Our results indicate that policies based off of training loss or model influence are an effective strategy over simple baselines, such as random sampling.
Finally, we propose several modifications to these approaches, such as sample reweighting and online learning, that can further improve performance. Our proposed policies are simple and straightforward to implement, requiring only a few lines of code. We perform experiments throughout on common benchmark datasets,
such as MNIST~\citep{lecun1998gradient}, CIFAR10~\citep{krizhevsky2009learning},
and NORB~\citep{lecun2004learning}\@.

\vspace{-.15em}
\section{Related Work}
In the domain of image classification, most state-of-the-art pipelines use some form
of data augmentation~\citep{cirecsan2010deep, dosovitskiy2016discriminative, graham2014fractional, sajjadi2016regularization}. This typically consists of applying crops, flips, or small affine transformations to all the data points in the training set, with parameters drawn randomly from hand-tuned ranges. Beyond image classification, various studies have applied data augmentation techniques to modalities such as audio~\citep{uhlich2017improving} and text~\citep{lu2006enhancing}. The selection of these augmentation strategies can have large performance impacts, and thus can require extensive selection and tuning~\citep{ratner2017learning}.

Motivated by the ubiquity of data augmentation and the difficulty in selecting
augmentations, there has been a significant amount of work in selecting and
tuning the best \textit{transformations} to use when performing augmentation.
For example,~\citet{fawzi2016adaptive} use adaptive data augmentation to choose
transformations that maximize loss for the
classifier;~\citet{ratner2017learning} propose learning a sequence model of
composed transformations; and~\citet{cubuk2018autoaugment} suggest a reinforcement learning approach. In contrast to these works, our aim is instead to select which \textit{data points} to augment while holding transformations fixed.
Our subsampling policies are therefore complementary to many of the described
approaches, and in fact, could be quite beneficial for approaches such as reinforcement learning that can quickly become infeasible for large datasets and transformation spaces.
Finally, we note that several recent works have proposed augmentation strategies based on adversarial training approaches, such as robust optimization frameworks or generative adversarial networks (GANs)~\citep{goodfellow2014generative,antoniou2017data,volpi2018generalizing}. These approaches generate artificial points from some target distribution, rather than by directly transforming the original training points.
We view these works as orthogonal and
complementary approaches to the proposed work, which is designed in concert with more traditional data augmentation strategies.

The area of work most closely related to our own is that of the Virtual Support Vector (VSV) method~\citep{burges1997improving, decoste2002training}. This method was proposed the support vector machine literature as a way to reduce the set of points for augmentation by limiting transformations to only support vectors. In the context of SVMs, the motivation is straightforward, as points that are far from the margin are unlikely to affect future models if they are transformed via small perturbations. However, to the best of our
knowledge, there has been no work extending these ideas to methods beyond SVMs,
where the notion of support vectors is not directly applicable.

Inspired by the VSV work, we similarly seek ways to downsample the set of
candidate points for augmentation, though through metrics beyond support
vectors. We begin by generalizing the notion of a support vector by simply
measuring the loss at each training point\footnote{We also investigate a more
direct generalization of the VSV method---sampling points according to their
distance from the margin---in Appendix~\ref{sec:margin}, although this method
generally underperforms the other metrics.}. We also explore model influence, which has been rigorously studied in the field of robust statistics as a way to determine which data points are most impactful on the model.
Model influence has been studied extensively in the regression
literature~\citep{hoaglin1978hat,pregibon1981logistic,cook1986assessment,walker1988influence}, and more recently, in non-differentiable (SVMs) and non-convex (deep networks) settings~\citep{koh2017understanding}.
We provide additional details on these metrics in Section~\ref{sec:policies}.

Finally, we note that this work is closely related to work in subsampling for general dataset
reduction (i.e., not in the context of data augmentation).
For example, works using gradients \citep{zhu2016gradient},
leverage \citep{drineas2011faster, drineas2012fast, ma2015statistical},
and influence functions \citep{mcwilliams2014fast, NIPS2018_7623, wang2018optimal} have shown better
results than uniform sampling of data samples in the original dataset.
Our scenario differs from the subsampling scenarios in these works as we
ultimately anticipate increasing the size of the dataset through augmentation, rather than decreasing
it as is the case with subsampling.
Subsampling methods are motivated by being unable to train models on entire
datasets due to the datasets being too large.
Our motivation is instead that the full \textit{augmented dataset} may be too large,
but the original training set is sufficiently small to be handled without
special consideration.
We therefore assume it is possible to obtain information (e.g.,
influence, loss, etc.) by fitting a model to the original data.
Further, the interpretation of our scenario differs, as the subsampling is performed with the ultimate aim being to retain the accuracy of some yet-to-be-determined fully augmented dataset, as opposed to the original dataset.

\section{Motivation: On the Effectiveness of Subsampling}
In this work, we seek to make data augmentation more efficient by providing
effective policies for subsampling the original training dataset. To motivate
the effect of subsampling prior to augmentation, we begin with a simple example.
In Table~\ref{table:best_vs_baseline}, we report the effect that performing
translation augmentations has on the final test accuracy for several datasets
(MNIST, CIFAR10, NORB). In the second column, we provide the final test accuracy
assuming \textit{none} of the training data points are augmented, and in the
last column, the final test accuracy after augmenting \textit{all} of the
training data points (i.e., our desired test accuracy). Note that the test
dataset in these examples has also been augmented with translation to better highlight
the effect of augmentation; we provide full experimental details in
Section~\ref{sec:exps}. In columns 3--8, we report test accuracies from
augmenting $5$, $10$, and $25$ percent of the data, where these subsamples are either derived using simple random sampling or via our proposed policies (to be discussed in Section~\ref{sec:policies}). 

An immediate take-away from these results is that, even in the case of simple
random sampling, it is often unnecessary to augment the entire
dataset to achieve decent accuracy gains. For example, augmenting just $25\%$ of
the dataset selected at random can yield more than half of the total accuracy
gain from full augmentation. However, it is also evident that subsampling can be
done more effectively with the appropriate policy. Indeed, as compared to random
sampling, when augmenting just $10\%$ of the data, these optimal policies can
achieve almost identical results to full augmentation (within $.1\%$ for CIFAR10
and higher accuracy than full augmentation for MNIST and NORB).
These results aim to serve as a starting point for the remaining paper.
We describe our proposed policies in detail in Section~\ref{sec:policies}, and
we provide full experiments and experimental details in Section~\ref{sec:exps}.

\begin{table}[h!]
\begin{center}
  \begin{tabular}{*{9}c}
  \toprule
  \multicolumn{1}{c}{Dataset}
  & \multicolumn{1}{c}{No Aug.}
  & \multicolumn{3}{c}{Baseline Random Policy}
  & \multicolumn{3}{c}{Best Policy}
  & \multicolumn{1}{c}{Full Aug.} \\
    \cmidrule(lr){3-5}
    \cmidrule(lr){6-8}
  & \multicolumn{1}{c}{$0\%$}
  & \multicolumn{1}{c}{$5\%$}
  & \multicolumn{1}{c}{$10\%$}
  & \multicolumn{1}{c}{$25\%$}
  & \multicolumn{1}{c}{$5\%$}
  & \multicolumn{1}{c}{$10\%$}
  & \multicolumn{1}{c}{$25\%$}
  & \multicolumn{1}{c}{$100\%$} \\
    \midrule
    MNIST
    & $93.2\%$
    & $99.0\%$
    & $99.3\%$
    & $99.5\%$
    & $99.7\%$
    & $99.8\%$
    & $99.7\%$
    & $99.6\%$
    \\
    CIFAR10
    & $96.3\%$
    & $96.6\%$
    & $96.8\%$
    & $97.0\%$
    & $97.0\%$
    & $97.2\%$
    & $97.3\%$
    & $97.3\%$
    \\
    NORB
    & $87.3\%$
    & $88.0\%$
    & $88.3\%$
    & $88.4\%$
    & $89.9\%$
    & $89.8\%$
    & $89.7\%$
    & $89.7\%$
    \\
    \bottomrule
\end{tabular}
\caption{Best observed policy vs.\ expected baseline with translate augmentations
  for various percentages of the training set being augmented.
The best policies are capable of reaching near full augmentation performance with a
  small augmentation budget.
}%
\label{table:best_vs_baseline}
\end{center}
\end{table}

\section{Augmentation Set Selection Policies}%
\label{sec:policies}

In this section, we provide details on our augmentation policies, including their general structure (described below), the metrics they utilize (Section~\ref{sec:metrics}), and improvements such as reweighting or online learning (Section~\ref{sec:improvements}).

\textbf{Setup.} The aim in this work is to find some subset $\S := \{(x_i, y_i), \dots (x_j, y_j)\}$ of the full training set $\D := \{(x_1, y_1), \dots (x_n, y_n)\}$, such that augmenting only the subset $\S$ results in similar performance to augmenting the entire dataset $\D$. More precisely, the goal is to minimize the size of $\S$, $|\S|$, subject to the constraint that $\text{perf}(\S_{aug}) \approx \text{perf}(\D_{aug})$, where $\S_{aug}$ and $\D_{aug}$ represent the dataset after appending augmented examples generated from the original examples in $\S$ or $\D$, respectively.
We note that while the performance measure $\text{perf}(\cdot)$ may be broadly
construed, we specifically focus on measuring performance based on test accuracy
in our experiments.

\textbf{General Policies.}
Our proposed policies consist of two parts: (i) an \textit{augmentation score}
which maps each training point $(x_i, y_i)$ to a value $s \in \R$,
and (ii) a \textit{policy} by which to sample points based on these augmentation scores.
In Section~\ref{sec:metrics}, we describe two metrics, loss and model influence, by which augmentation scores are generated.
In terms of
policies for subset selection based on these scores, we first explore two simple
policies---deterministic and random. In particular, given a set of augmentation
scores $\{s_1, \dots, s_n\}$ for the $n$ training points, we select a subset
$\mathcal{S} \subseteq \mathcal{D}$ either by ordering the points based on their
scores and taking the top $k$ values (in a deterministic fashion), or by
converting each augmentation score $s_i$ to a probability
$\pi_{i}(z) \in [0, 1]$,
and then sampling according to this distribution without replacement.
As the augmentation scores (and resulting policies) may be affected by updates to the model after each augmentation, we additionally explore in Section~\ref{sec:improvements} the effect of iteratively updating or re-weighting scores to adjust for shifts in the underlying model.
A non-exhaustive overview of the various augmentation policies is provided in
Table~\ref{table:policies_overview}.

\begin{table}[h!]
\begin{center}
  \begin{tabular}{*{4}c}
  \toprule
    \multicolumn{1}{c}{Policy Type}
  & \multicolumn{1}{c}{Selection Function}
  & \multicolumn{1}{c}{Update Scores}
  & \multicolumn{1}{c}{Downweight Points} \\
    \midrule
    Baseline
    & $P(z_i) = \tfrac{1}{n}$
    & \xmark%
    & \xmark%
    \\
    Random Prop.
    & $P(z_i) = \tfrac{s_i}{\sum_j s_j}$
    & \xmark%
    & \xmark%
    \\
    Deterministic Prop.
    & $\text{Rank}(z_i) = \texttt{SELECT}_S^{-1}(s_i)$
    & \xmark%
    & \xmark%
    \\
    Random Prop. Update
    & $P(z_i) = \tfrac{s_i}{\sum_j s_j}$
    & \checkmark%
    & \xmark%
    \\
    Rand. Prop. Downweight
    & $P(z_i) = \tfrac{s_i}{\sum_j s_j}$
    & \xmark%
    & \checkmark%
    \\
    \bottomrule
\end{tabular}
\caption{%
  Overview of the augmentation policies and their parameters, where $s_i$ is the augmentation score given to point $z_i = (x_i, y_i)$.
  The $\texttt{SELECT}_S^{-1}$ function corresponds to the inverse of an order
  statistic function. As a baseline, we compare to sampling data points at random, ignoring the augmentation scores. Note that the notation here is simplified to allow sampling with replacement, though in practice we perform sampling without replacement.
 }%
\label{table:policies_overview}
\end{center}
\end{table}

\subsection{Metrics: Loss and Influence}%
\label{sec:metrics}

We propose two metrics to determine our augmentation scores: training loss and model influence. 

\textbf{Training loss.} One method to obtain augmentation scores is the loss at a point in the training set. This can be viewed as a more direct generalization of the virtual support vector (VSV) method, as support vectors are points with non-zero loss.
However, studying loss directly will allow us: (i) to extend to methods beyond SVMs, and (ii) to expand the augmented set to data points beyond just the fixed set of support vectors.

\textbf{Influence.} We also explore policies based on Leave-One-Out (LOO) influence, which measures
the influence that a training data point has against its own loss when it is removed from the
training set.
We follow the notation used in \citet{koh2017understanding}.
Let $\hat{\theta}$ be the minimizer of the loss, which is assumed to be twice
differentiable and strictly convex in $\theta$.
Let $H_{\hat{\theta}}$ be the Hessian of the loss with respect to $\theta$
evaluated at the minimizer.
We define
the influence of upweighting a point, $z$, on the loss at a test point,
$z_\text{test}$, as
$\mathcal{I}_\text{up,loss}(z,z_\text{test}) :=
-\nabla_{\theta}L(z_\text{test},\hat{\theta})^\top H_{\hat{\theta}}^{-1}
\nabla_{\theta}L(z,\hat{\theta})$.
It follows that if the test point is $z$, then the LOO influence can be calculated as:
\begin{equation}
\label{eq:infl}
\mathcal{I}_\text{LOO}(z) := \mathcal{I}_\text{up,loss}(z,z) = -\nabla_{\theta}L(z,\hat{\theta})^\top H_{\hat{\theta}}^{-1} \nabla_{\theta}L(z,\hat{\theta})\,.
\end{equation}
For our augmentation scores, we care only about the magnitude of the LOO influence, so it
can be assumed that the sign is dropped.

To understand the potential of using training loss and model influence for
scoring, we provide a histogram of model influence across the CIFAR10 and NORB datasets in Figure~\ref{fig:inf_hist_translate}. Full results for all datasets and for training loss are provided in Appendix~\ref{sec:histograms}. In Figure~\ref{fig:inf_hist_translate}, we see that while most of the mass is centered around $0$ (which we
utilize to avoid points), there is sufficient variability to allow for
ranking points by preference.
Further, as seen in Figure~\ref{fig:inf_joint_translate}, these values are
correlated before and after augmentation, indicating that these metrics are a reliable measure of the future impact of a data point after augmentation.
We observe Spearman's rank correlations~\citep{spearman1904proof} between $0.5$ and $0.97$ with p-values less than
$0.001$ (and usually orders of magnitude less).

\vspace{-.5em}
\begin{figure}[ht]
\begin{minipage}{.47\textwidth}
\begin{center}
\subfloat[CIFAR10]{%
  \includegraphics[trim=5 0 10 0, clip,width=0.5\linewidth]{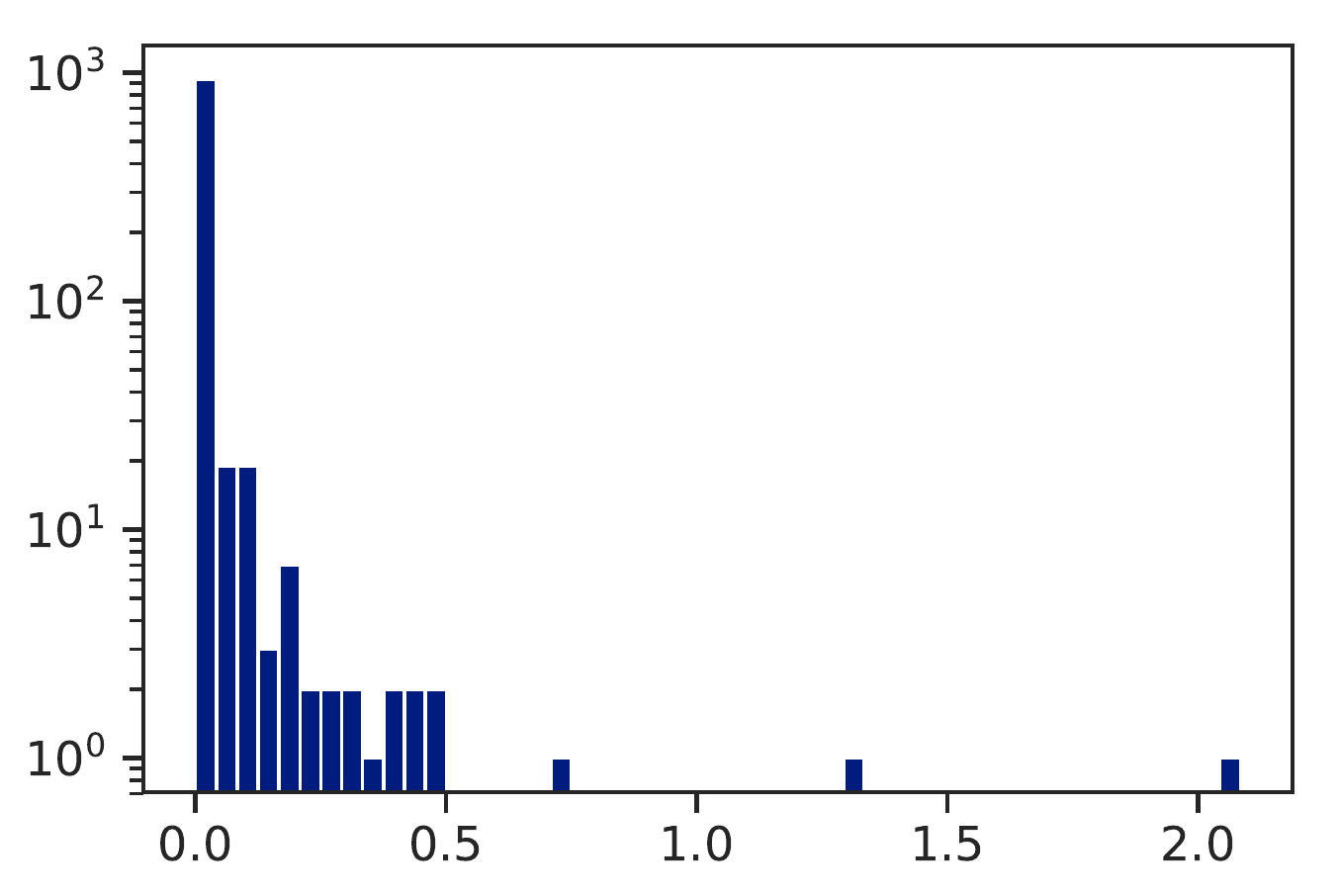}%
  }%
  \subfloat[NORB]{%
  \includegraphics[trim=5 0 10 0, clip,width=0.5\linewidth]{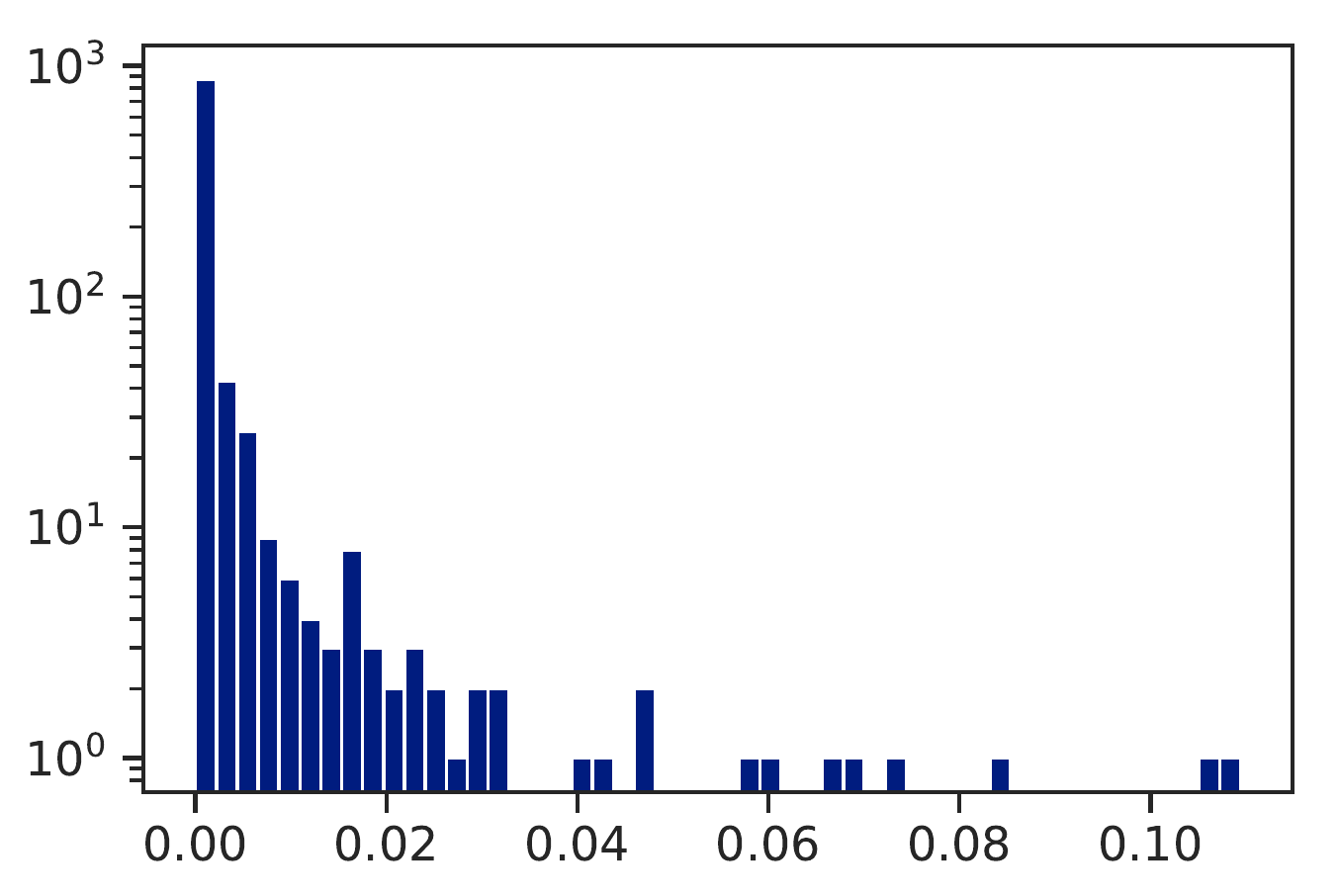}%
  }%
  \captionof{figure}{Distribution of influence function values on initial training set for translate
  augmentations. Most values are not influential and can therefore be augmented
  with low priority. We find similar results when measuring training loss
  (Appendix~\ref{sec:histograms}).}%
  \label{fig:inf_hist_translate}%
\end{center}
\end{minipage}%
\hfill
\begin{minipage}{.47\textwidth}
\vspace{-1em}
\begin{center}
\subfloat[CIFAR10]{%
  \includegraphics[trim=20 10 10 20, clip, width=0.49\linewidth]{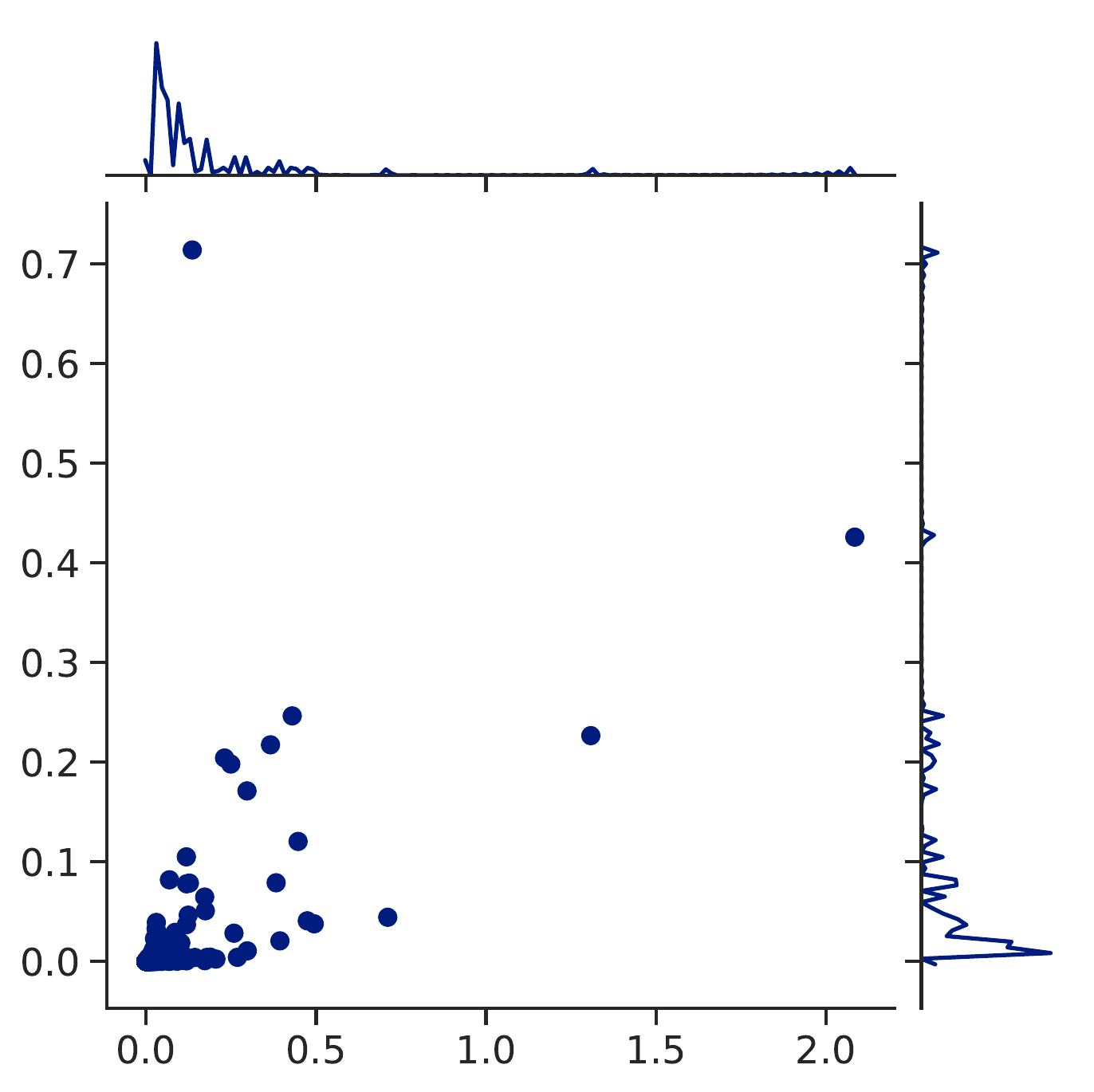}%
  }%
  \subfloat[NORB]{%
  \includegraphics[trim=20 10 10 20, clip, width=0.51\linewidth]{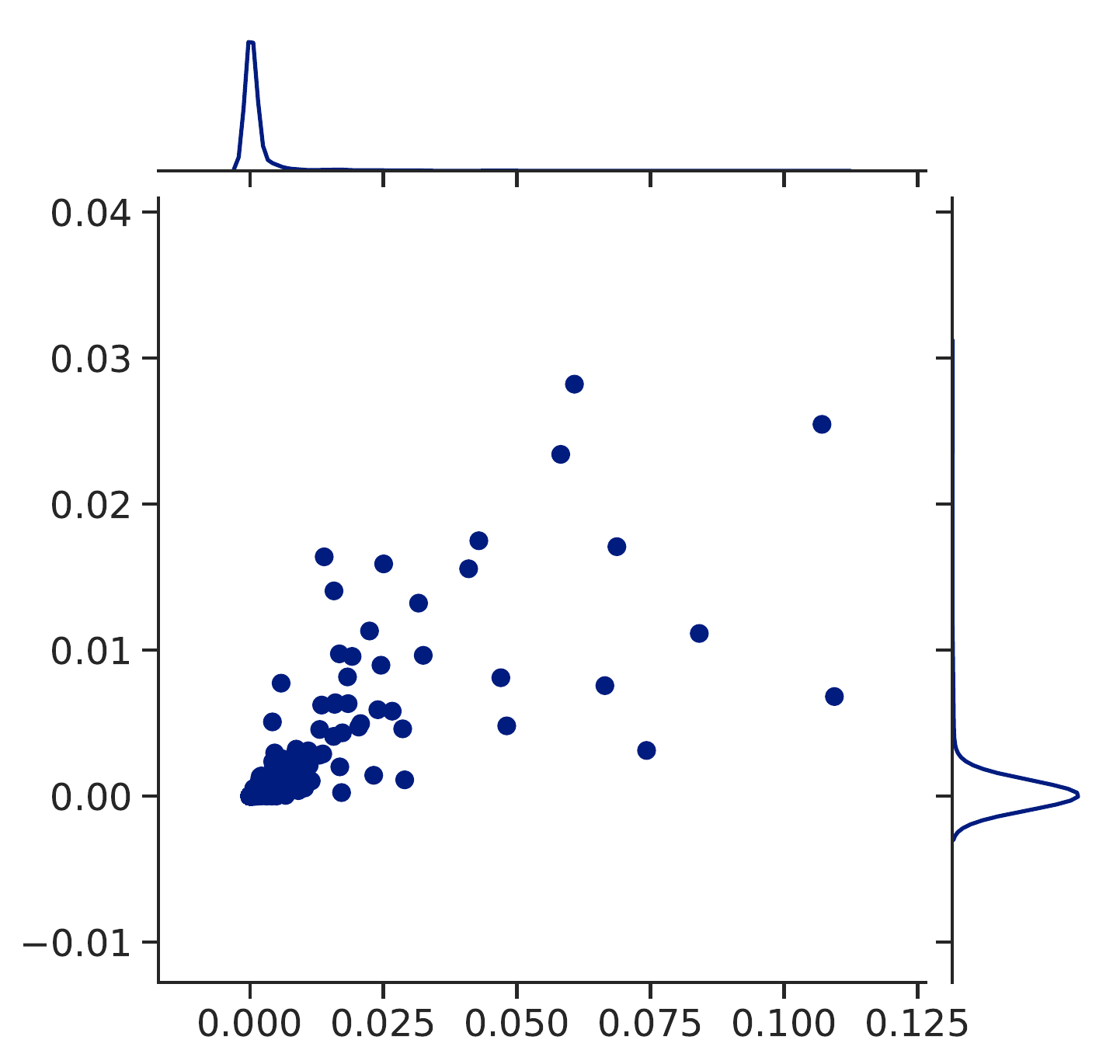}%
  }%
  \caption{Influence distribution on initial training set (x-axis)
  vs.\ final training set (y-axis) for translate augmentations.
  Points that are uninfluential typically remain uninfluential.
  }%
\label{fig:inf_joint_translate}
\end{center}
\end{minipage}
\end{figure}

\subsection{Refinements: Sample Reweighting and Score Updating}%
\label{sec:improvements}

\textbf{Reweighting.}
To motivate reweighting individual samples, consider an augmentation which is the identity map: $f_T: z_i \to \{z_i\}$.
Since we add augmentations back to the training set, our augmentation
policy will duplicate selected samples, resulting in a net
effect which reweights samples with twice the original weight.
Using transformations that result in larger augmentation sets will result in
larger weights.
One approach is post-processing; \citet{fithian2014local}, for example, show that the case of class imbalanced sampling can be
corrected with a shift of the logistic regressor's bias.
To normalize for the effect of this implicit reweighting, we divide the weights
of the original samples and its augmented samples by the size of that set,
$|f_T(z_i)|$.
Under this scheme, we guarantee that we conserve the weight originally assigned
to a point (and conserve the ratios of labels).
More sophisticated policies, such as reweighting samples by a measure of how
trustworthy they are (e.g., perhaps some bounds can be derived on the
label-preserving properties of an augmentation), remain to be investigated as
future work.

We find that in many cases, the performance of reweighting is similar in expectation to the
base case.
However, in some cases, reweighting can have a negative impact, as we discuss in Section~\ref{sec:refinementexps}.
We expect this policy to be more useful in the case of class
imbalance, where the duplication of minority class samples may significantly
alter the distribution over classes.

\textbf{Updating scores.} Once we decide to augment a data point, we can either continue to use the same
influence information which we derived for the un-augmented dataset, or we can
choose to update it.
The reason for doing this is to account for the drifting model behavior as
points are added to the training set and the model is
retrained.
However, if having a single estimate of influence for the whole lifetime of the
model is sufficient, then avoiding repeated influence calculations will reduce
the amount of computation required while also enabling an increased level of
parallelism (e.g., minibatching, distributed computations).
We find that this modification results in similar behavior to that of
reweightings, where expected performance of a policy remains similar.
Overall, we have not observed a significant enough effect to suggest that this
technique is justified given the extra cost it requires.
The benefit of this is that it implies that many applications may need to only compute selection metadata
one time throughout the augmentation process.

\section{Experiments}%
\label{sec:exps}

In this section, we provide detailed results on the performance of our proposed
policies for data subsampling. For all experiments, we use a Convolutional
Neural Network (CNN) to create bottleneck
features, which we then use as input into a linear logistic regression model.
This is equivalent to freezing the weights of the CNN, or using a set of basis
functions, $\phi_i(\cdot)$, to transform the inputs~\citep{bishop2006pattern}, and
allows us to quickly calculate training loss and model influence. We explore the
results of our augmentation policies on three datasets: binary classification
variants of MNIST, CIFAR10, and NORB\@.
For MNIST features, we use a LeNet
architecture \citep{lecun1998gradient} with ReLu activations, and for CIFAR10
and NORB, we use ResNet50v2  \citep{he2016identity}. While for CIFAR10 and NORB we generate the bottleneck features once due to cost, for MNIST, we additionally study the effect of re-generating these features as new points are selected and augmented (i.e., training both the features and model from scratch throughout the experiments).

In terms of augmentations, we consider three examples: translation, rotation, and crop. To control for sources of variance in model performance, all augmentations under
consideration are applied exhaustively in a deterministic fashion to any selected
samples, and the resulting augmented points are then added back to the training set. Formally, given a data point, $z = (x, y) \in \mathcal{X} \times \mathcal{Y}$, our augmentation is a map from
a data point to a finite set of data points: $f_T: z \to \{z_1, \dots, z_n : z_i
\in \mathcal{X} \times \mathcal{Y}\}$. We controlled for augmentation-induced regularization by performing a simple
 cross validation sweep for the regularization parameter $\lambda$ each time the
 model was re-trained, and we found regularization to have negligible impact in
 the trends we observed. For all datasets and augmentations, we make the effect
 of augmentation more apparent by adding augmented test points to the test set.
 For example, in the case of translation, we test the performance of applying
 translation augmentations to the original training set, and then determine the
 accuracy using an augmented variant of the test data that has been appended
 with translated test examples.
All augmentations are performed using
 Imgaug~\citep{imgaug}, and our code is written
 in Python using Keras CNN implementations.
Full implementation details are provided in Appendix~\ref{sec:implementation},
and our code is publicly available
 online\footnote{\url{https://github.com/mkuchnik/Efficient_Augmentation}}.

\newcommand{\trimfig}[1]{\includegraphics[trim= 20 0 10 0, clip, width=0.32\linewidth]{#1}}
\newcommand{\trimfigleft}[1]{\includegraphics[trim= 0 0 10 0, clip, width=0.336\linewidth]{#1}}

\begin{figure}[ht]
\begin{center}
  \subfloat[MNIST-translate]{%
  \trimfigleft{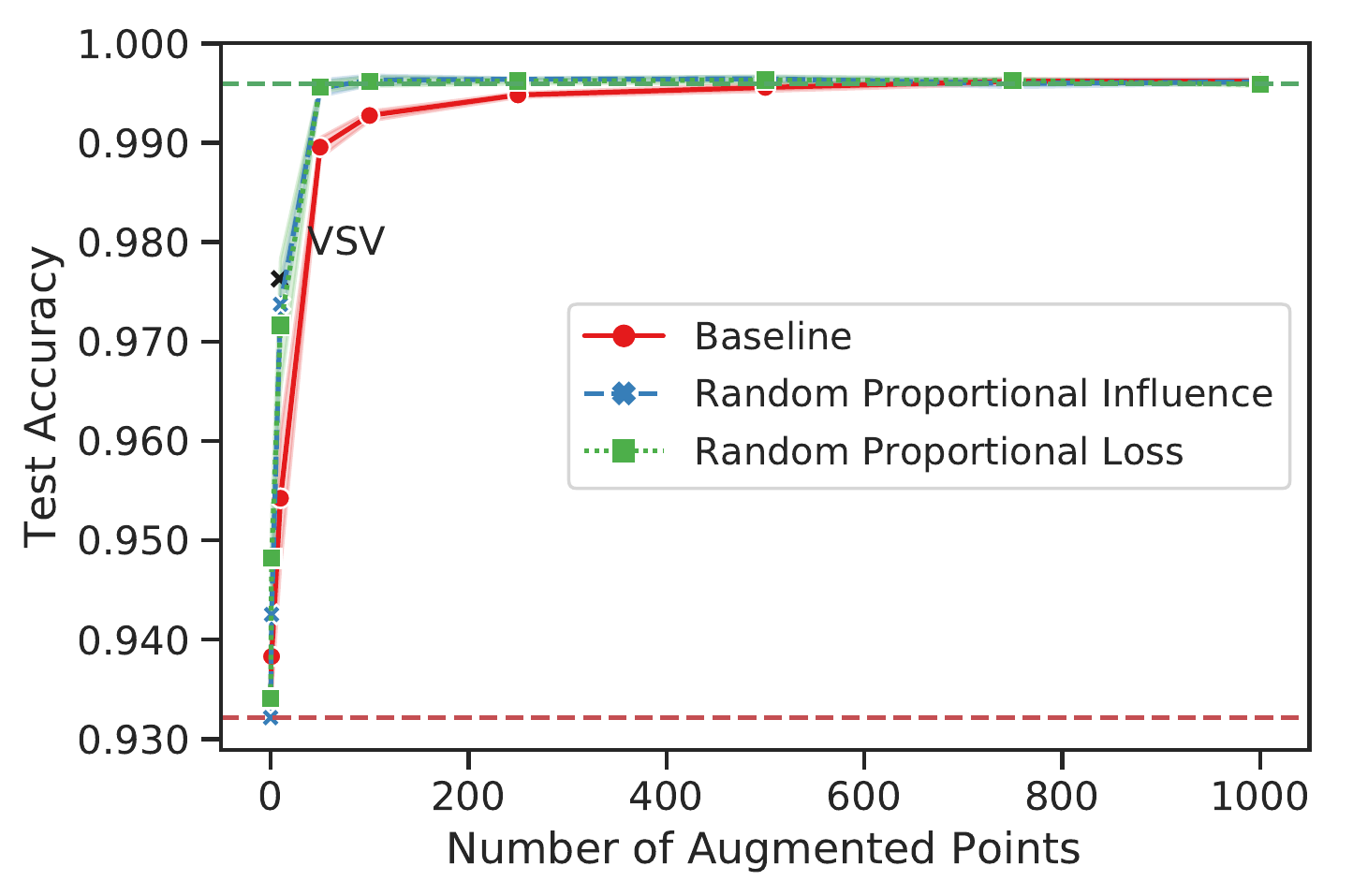}
  }%
  \subfloat[CIFAR10-translate]{%
  \trimfig{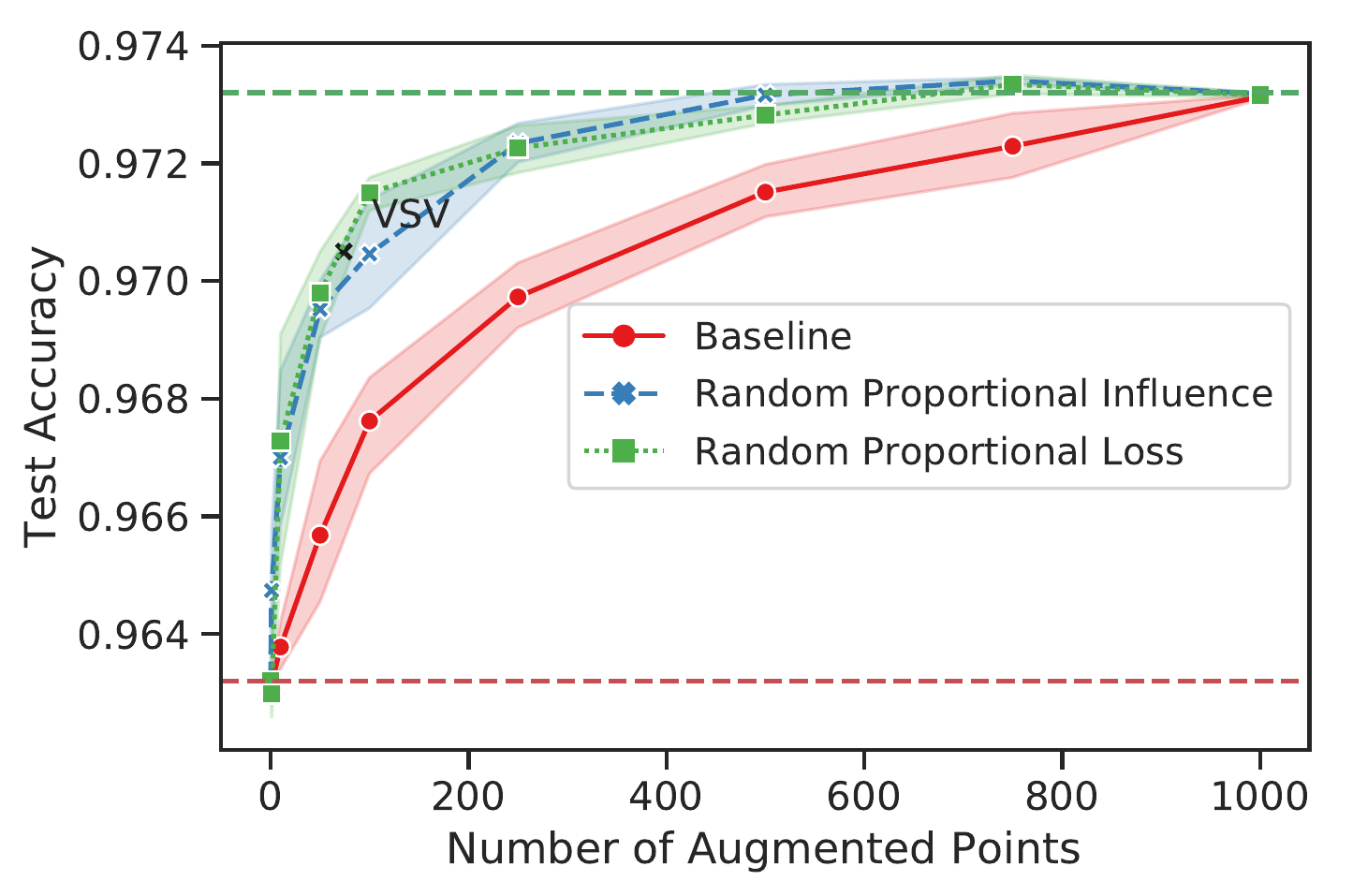}
  }%
  \subfloat[NORB-translate]{%
  \trimfig{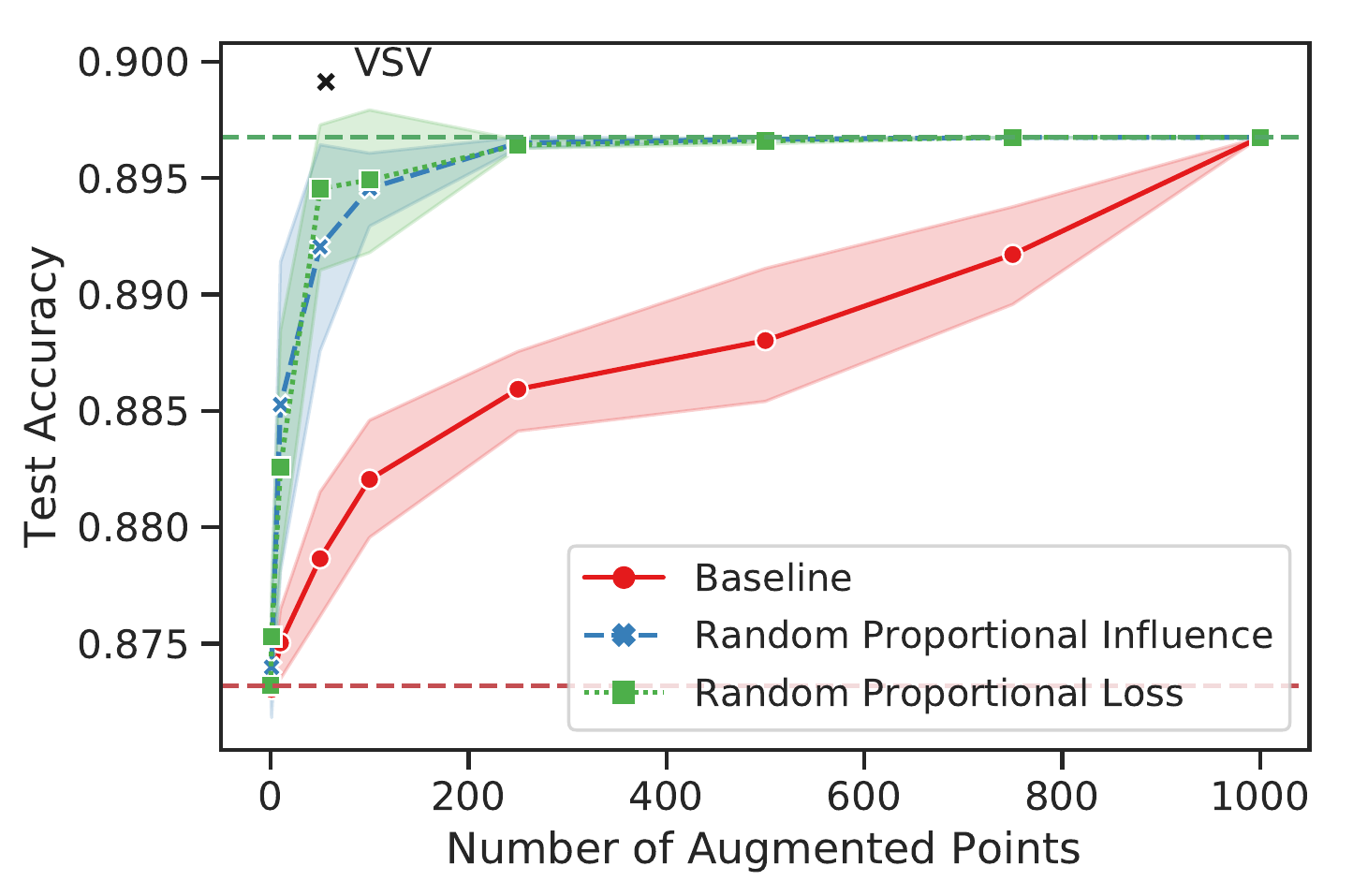}
  }%
  \newline \vspace{-1.35em}
  \subfloat[MNIST-rotate]{%
  \trimfigleft{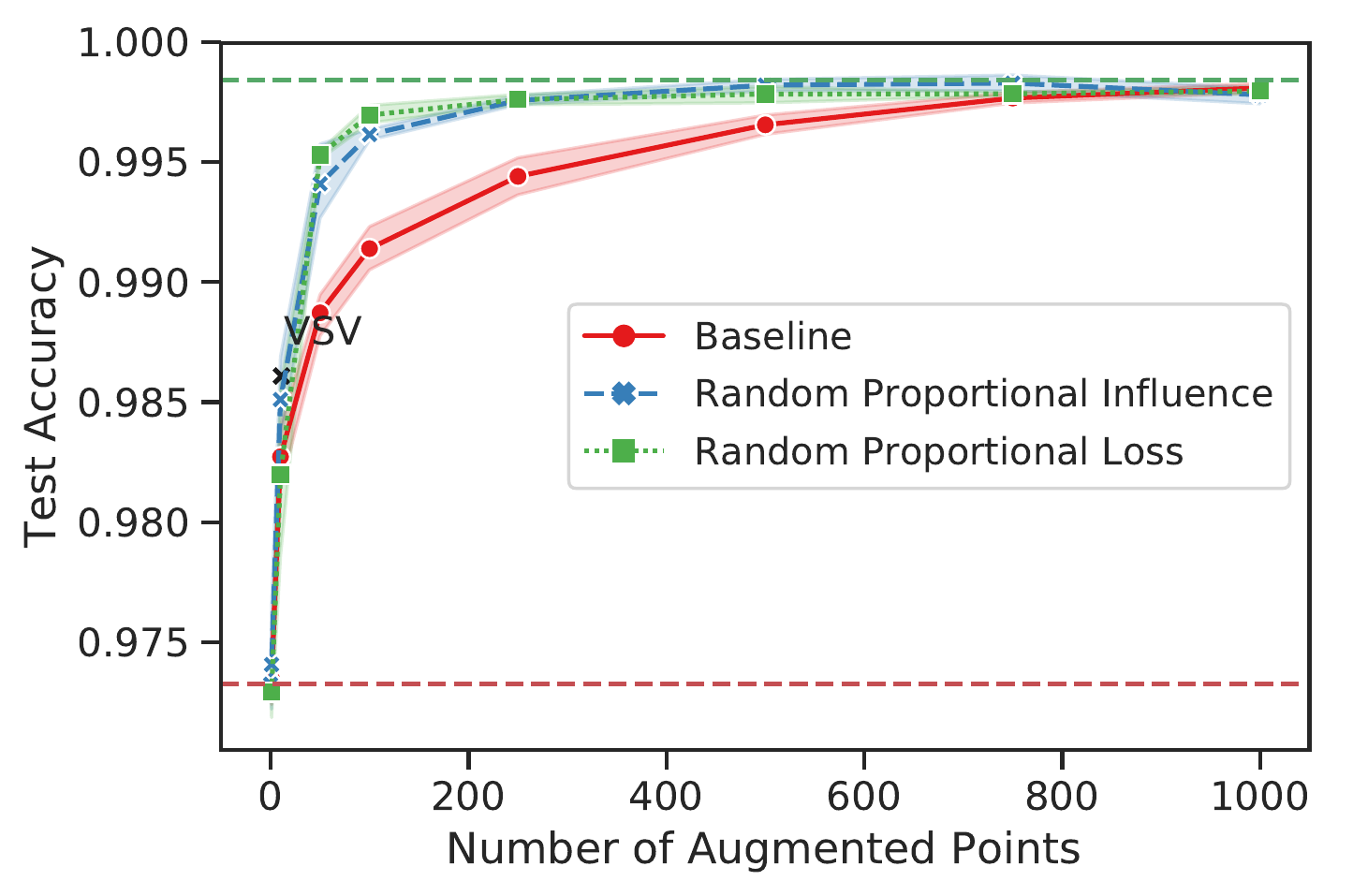}
  }%
  \subfloat[CIFAR10-rotate]{%
  \trimfig{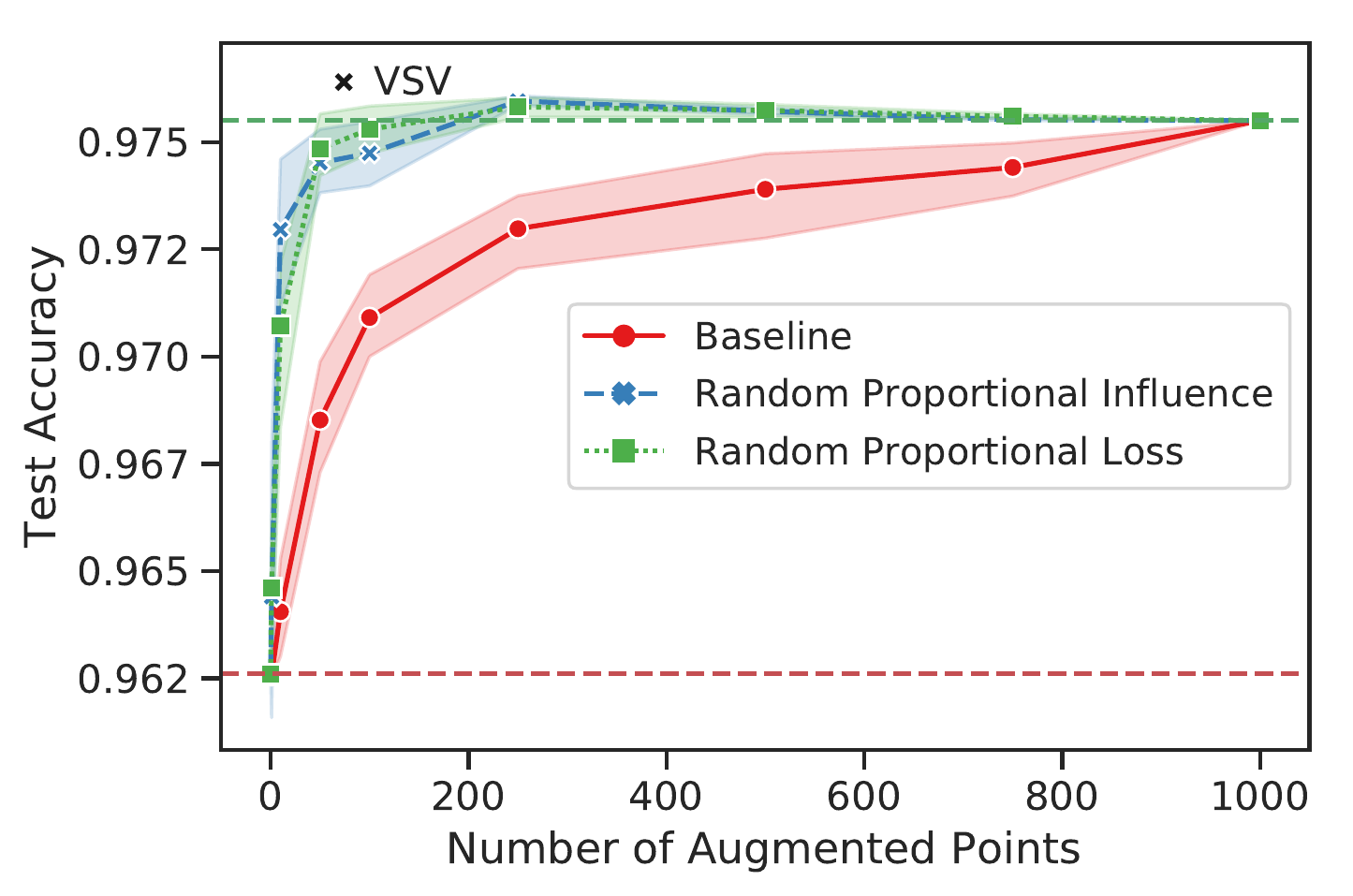}
  }%
  \subfloat[NORB-rotate]{%
  \trimfig{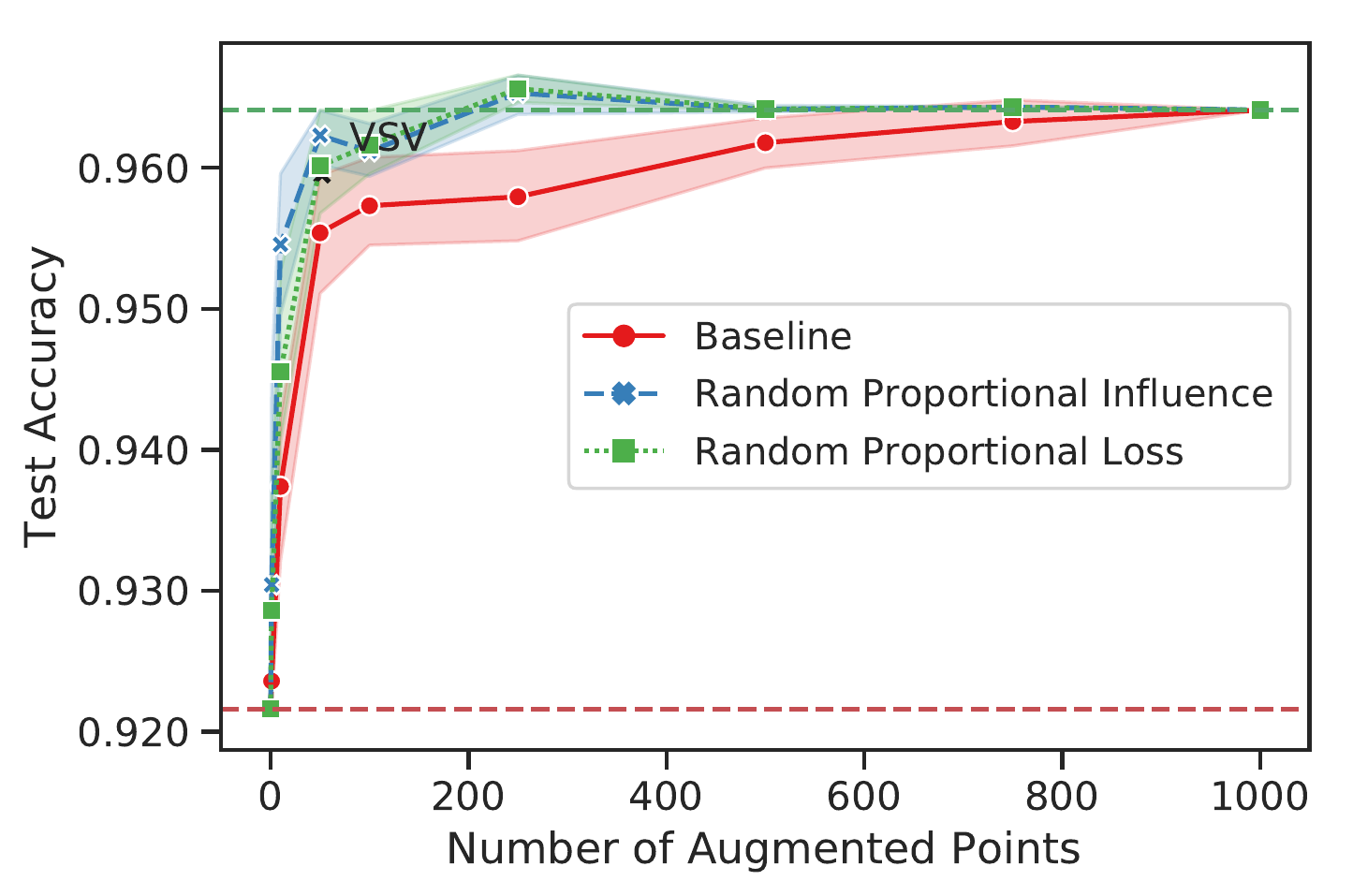}
  }%
  \newline  \vspace{-1.35em}
  \subfloat[MNIST-crop]{%
  \trimfigleft{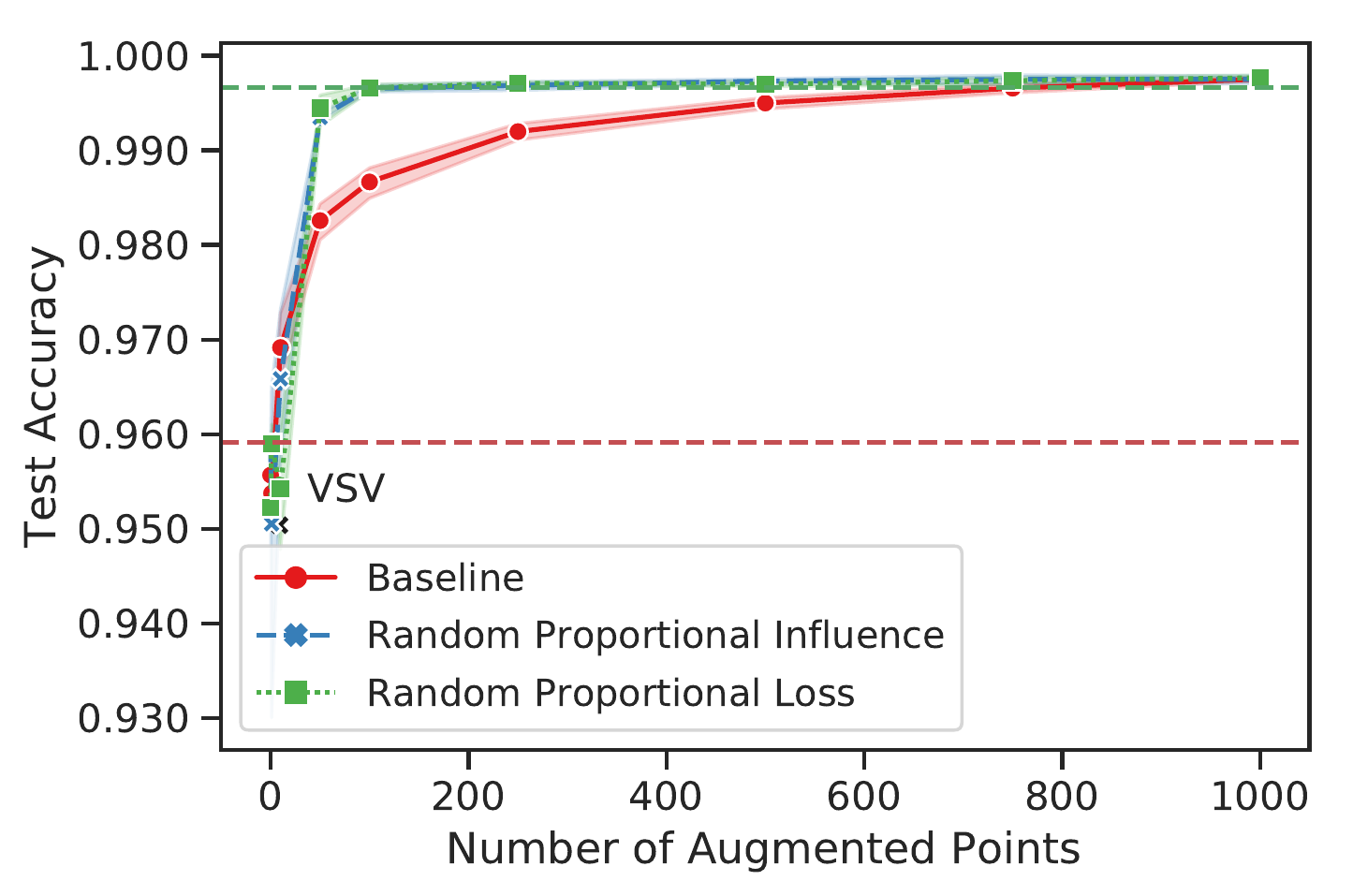}
  }%
  \subfloat[CIFAR10-crop]{%
  \trimfig{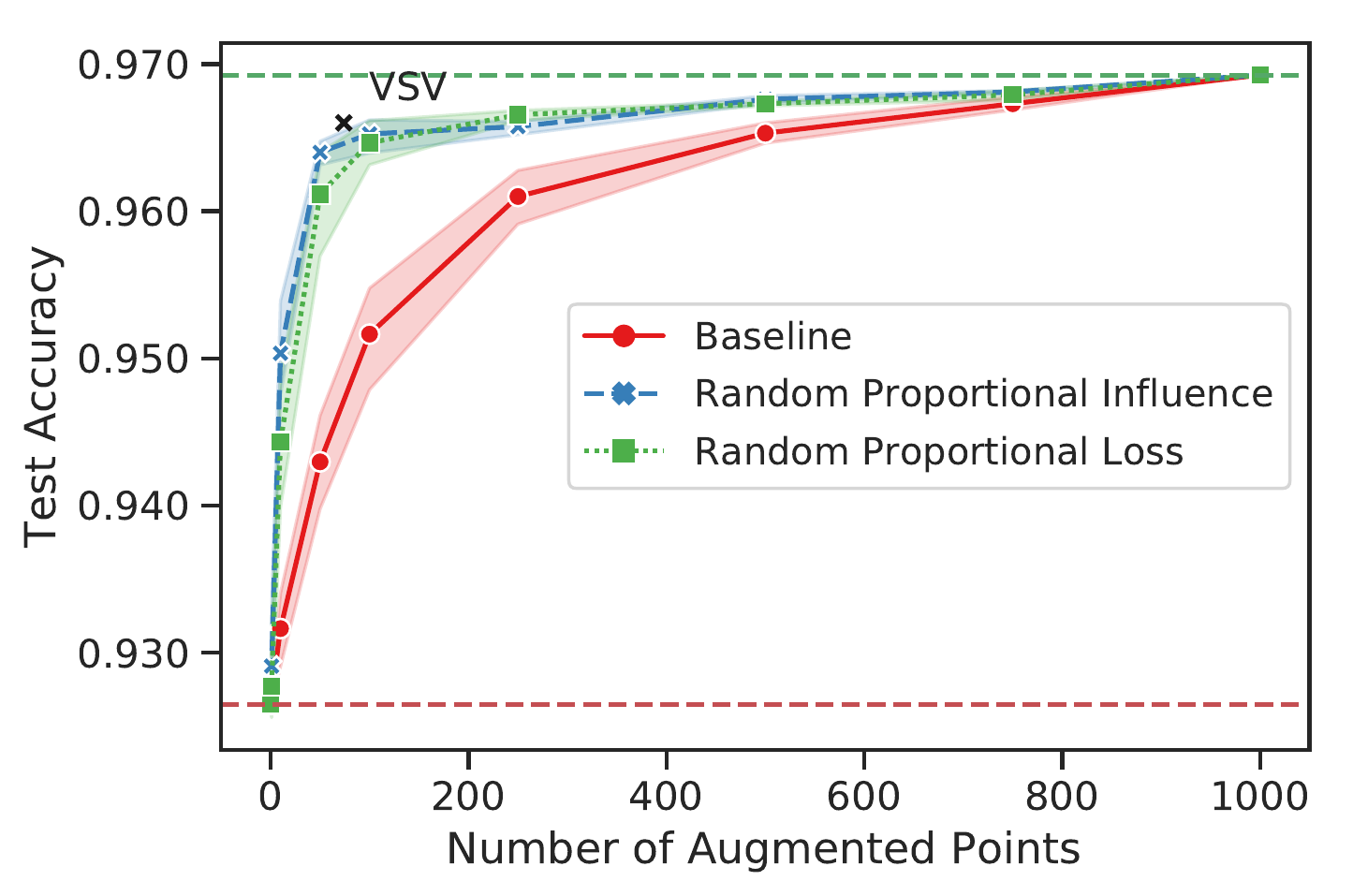}
  }%
  \subfloat[NORB-crop]{%
  \trimfig{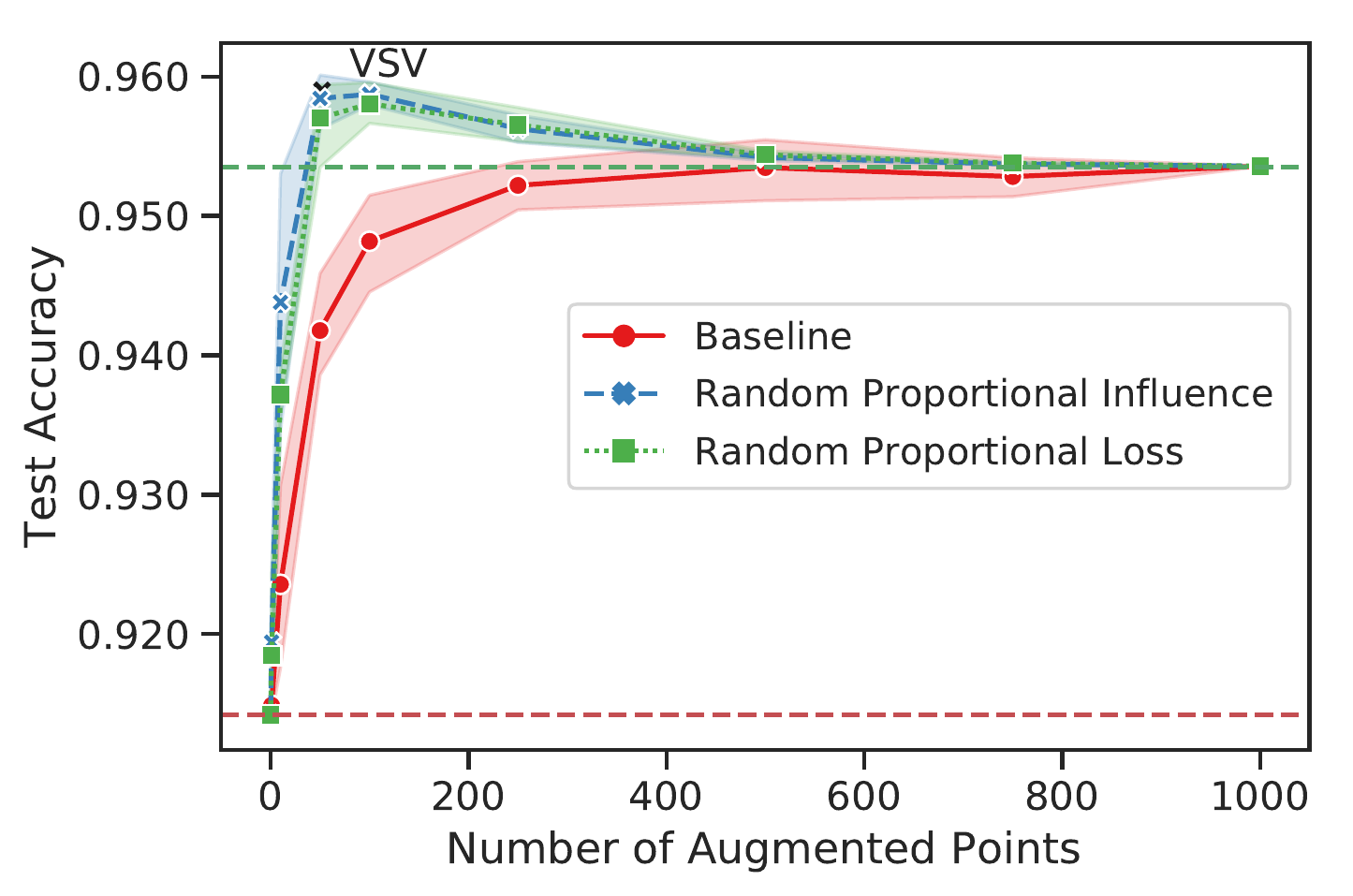}
  }%
  \vspace{.5em}
\end{center}%
\caption{The performance of random policies using influence and loss vs.\
  the baseline (simple random sampling). Random sampling based on loss/influence
  consistently outperforms the baseline.}%
\label{fig:general}
\end{figure}

\subsection{General Policies: Influence and Loss} In Figure~\ref{fig:general},
we explore a first set of policies in which we randomly sample points for
augmentation proportional either to their loss (green) or influence value
(blue). To calculate the loss and influence, we incur a one-time cost of
training the model on the original dataset. As a baseline (red), we compare
these methods to a simple strategy in which data points for augmentation are
drawn entirely at random (irrespective of loss or influence). The red-dotted
horizontal line indicates the test accuracy with no augmentation, and the green-dotted
line indicates the test accuracy after augmenting the entire training set.
Note that all policies have the same accuracy when the number of points is $0$ or $k$, where $k$ is the number of points in the
original training set, which correspond to the un-augmented training set and the fully augmented
training set, respectively\footnote{In practice, the non-convexity of CNNs results in accuracies
which may vary slightly.}.
We observe similar behavior in terms of the deterministic policy, which is provided in Appendix~\ref{sec:fullplots}.

Across the datasets and transformation types, we notice several trends. First,
the policies based on loss and influence consistently outperform the random
baseline. This is particularly true for the rotation augmentation for all three
datasets, where the random-influence and random-loss policies achieve the full
augmentation accuracy with only 5--10\% of the data, compared to 90--100\% of the
data for random sampling. Second, we note that the policies based on influence
vs.\ loss behave very similarly. While influence has slightly better performance
(particularly on the NORB dataset), the policies are, for the most part,
equivalent. A benefit of this is that the loss calculation is slightly simpler
than influence to calculate, as it does not require calculating the inverse
Hessian component, $H_{\hat{\theta}}^{-1}$, as described in~\ref{eq:infl}.
Third, we note that it is possible to achieve \textit{higher} accuracy than full
augmentation using only a reduced set of points for augmentation, as observed in
several of the plots (most notably on NORB).
We believe that this higher performance may be due to a stronger bias towards
harder examples in the dataset as compared to full augmentation.

Finally, we explore the effect of using support vectors for augmentation, which was proposed in the Virtual Support Vector literature~\citep{burges1997improving, decoste2002training}. In particular, we find VSV points by tuning a linear SVM on the bottleneck features of
the original training set, and then using these points as the set of
augmentation points for the logistic regression model with bottleneck features.
We use search over $C \in \{0.01, 0.1, 1, 10, 100\}$ via cross-validation, and
the best resulting model is used to obtain support vectors. Interestingly, we
note that, though this transfer approach was not originally proposed in the VSV
literature, it results in strong performance on a few of our tests (e.g.,
NORB-translate, NORB-crop, CIFAR10-rotate). However, the approach is not as
reliable as the proposed policies in terms of finding the optimal subset of
points for transformation (performing significantly below optimal, e.g., for
MNIST and CIFAR10-translate), and the major limitation is that the augmentation
set size is fixed to the number of support vectors rather than being able to vary depending on a desired data budget.

\subsection{Refinements: Sample Reweighting and Score Updating}%
\label{sec:refinementexps}

\begin{figure}[H]
\begin{center}
  \subfloat[MNIST-rotate]{%
  \trimfigleft{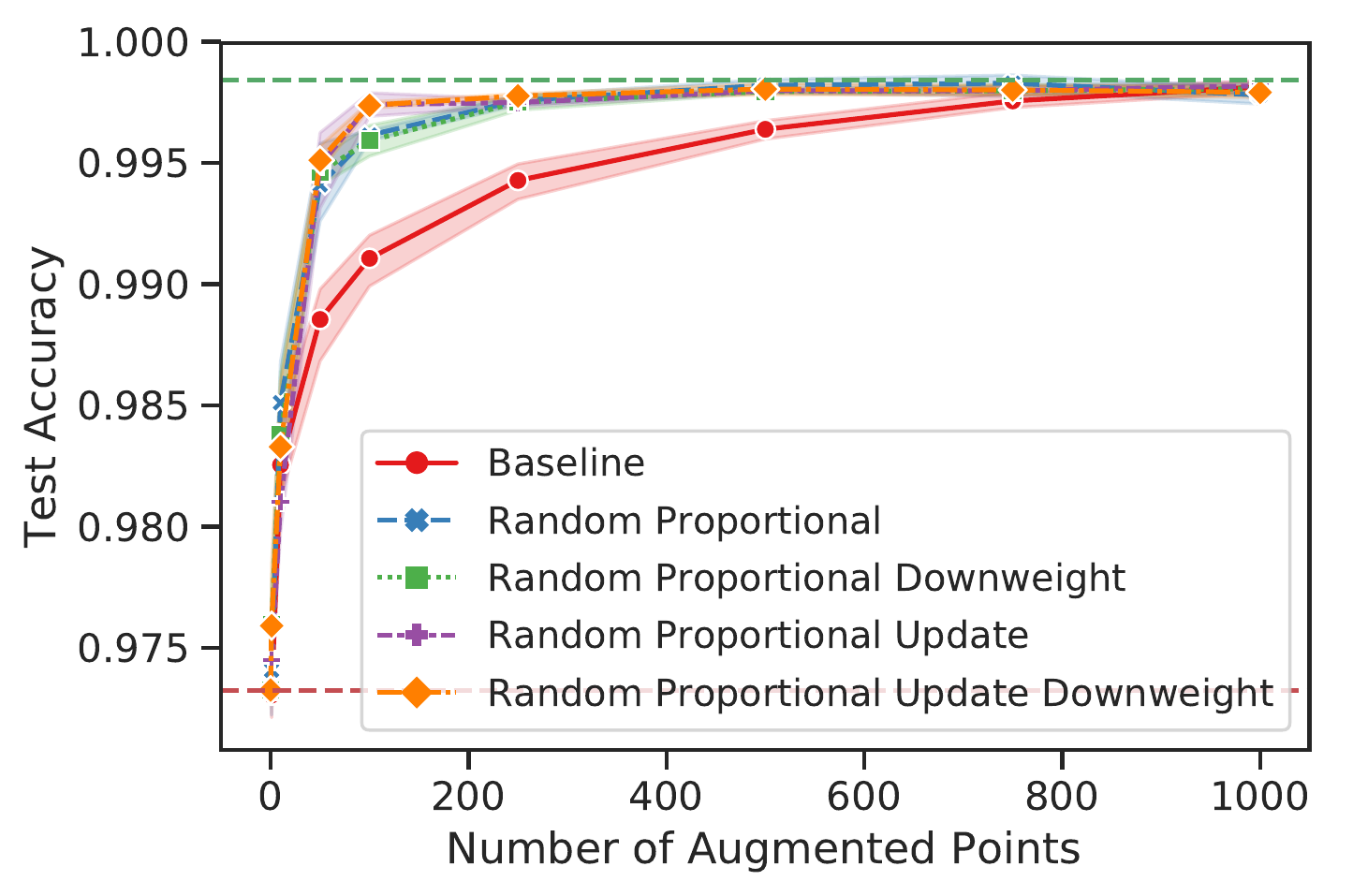}
  }%
  \subfloat[CIFAR10-rotate]{%
  \trimfig{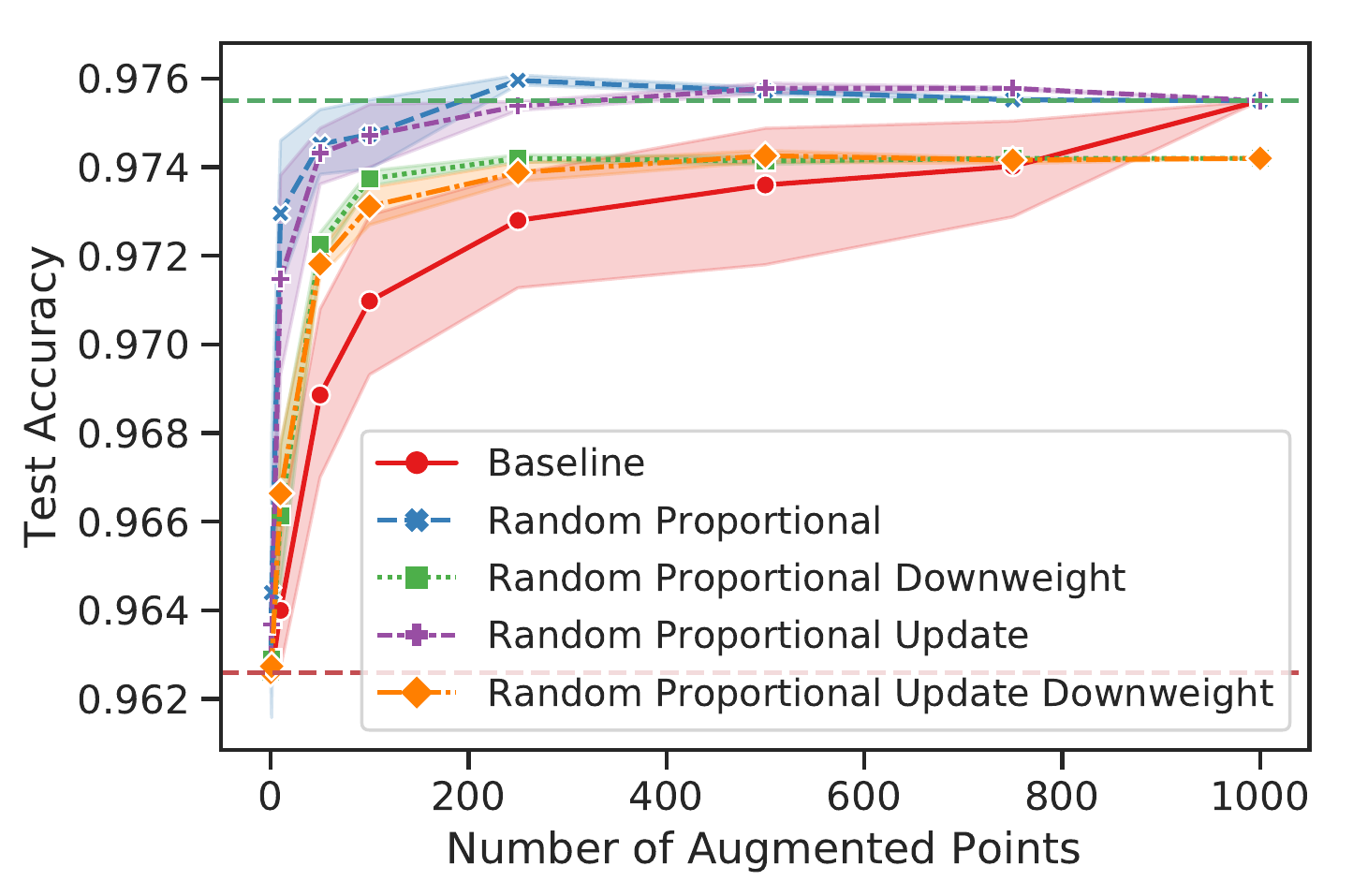}
  }%
  \subfloat[NORB-rotate]{%
  \trimfig{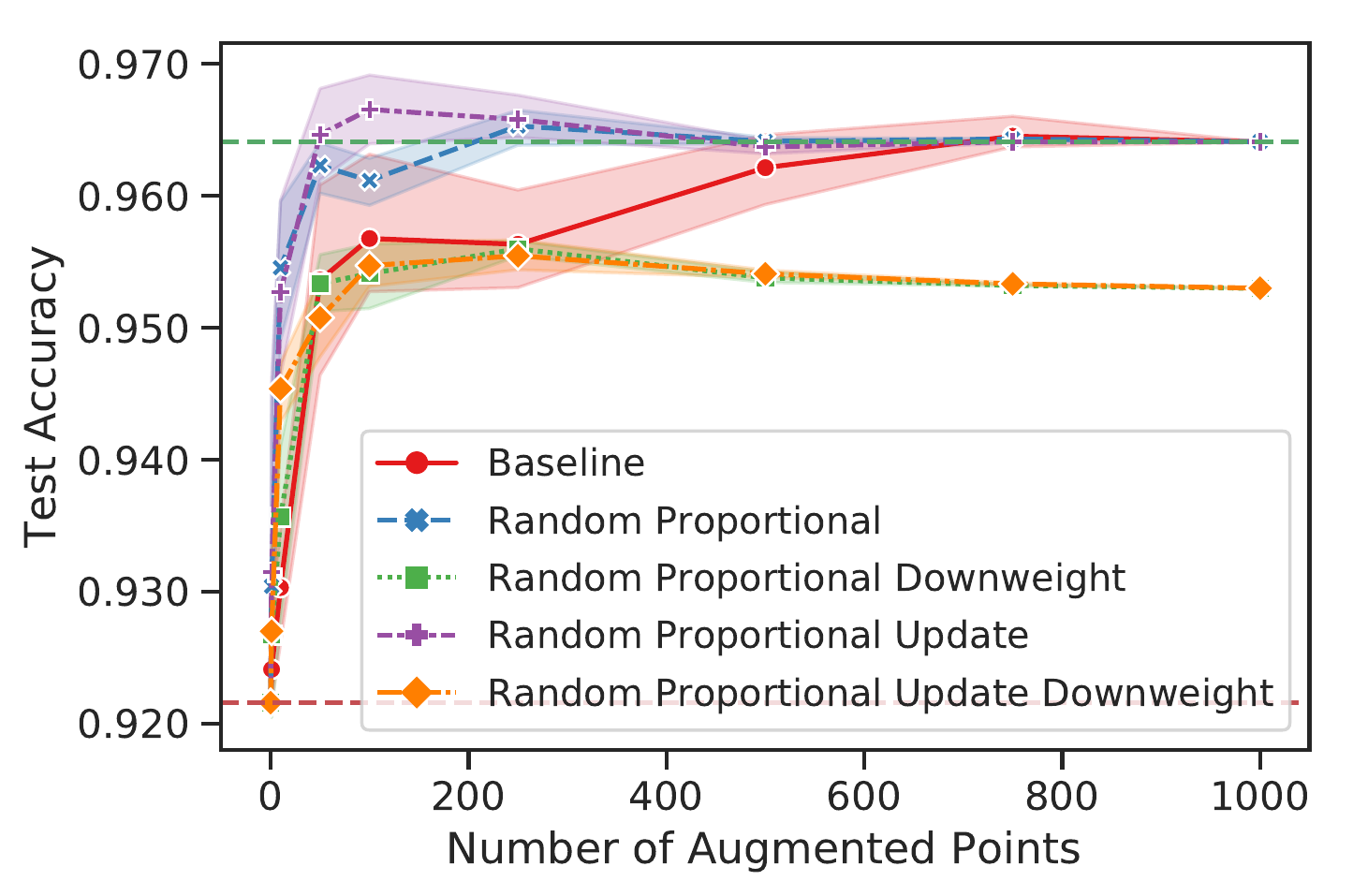}
  }%
  \caption{The performance of policies when point downweighting is used or
  augmentation scores are updated.}%
\label{fig:selected_mod}
\end{center}
\end{figure}

We additionally investigate the effect of two refinements on the initial policies:
(i) reweighting the samples as they are added back to the training set and
(ii) updating the scores as the augmentation proceeds, as described in
Section~\ref{sec:improvements}.
The latter policy assumes that the method is
run in an online fashion, in contrast to the policies described thus far.
This
adds extra expense to the total run time, because the model must be continually
updated as new points are augmented. In Figure~\ref{fig:selected_mod}, we
observe the effect of these modifications for all datasets using the rotation
augmentation with model influence as the score.
Full results are provided
in Appendix~\ref{sec:fullplots}. Interestingly, while reweighting points seems
to have a positive (if negligible) effect for MNIST, we see that it can actually
hurt performance in CIFAR10 and NORB\@.
This may indicate that the amplifying effect of augmentation may in fact be
beneficial when training the model, and that reweighting may increase
the role of regularization in a detrimental manner.
In terms of the score updating, we see that, although updating the score can have a slight positive impact (e.g., for NORB-rotate), the performance appears to roughly match that of the original policy.
Given the extra expense required in model updating, we therefore conclude that the simpler policies are preferable.

\vspace{-1em}

\begin{wrapfigure}{R}{0.48\textwidth}
\vspace{-1em}
\begin{center}
  \includegraphics[trim={1em 0 0 0}, clip, width=\linewidth]{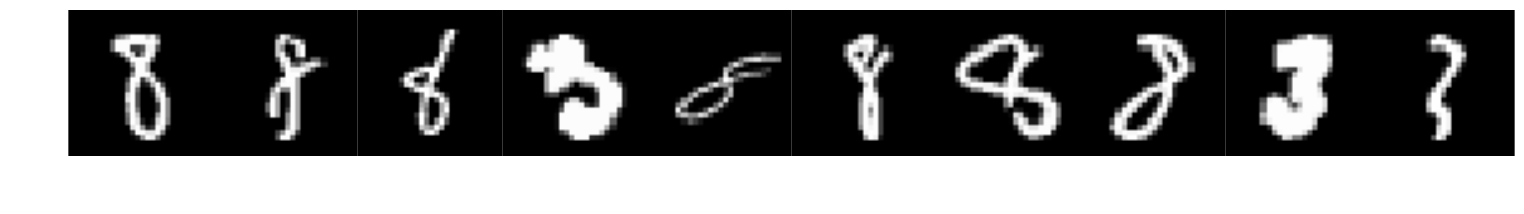}\\
  \vspace{-2mm}
  \includegraphics[trim={1em 0 0 0}, clip, width=\linewidth]{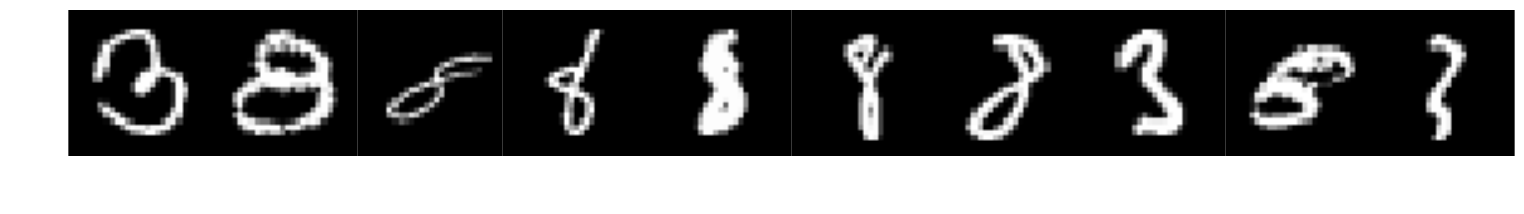} \\
  \vspace{-2mm}
  \includegraphics[trim={1em 0 0 0}, clip, width=\linewidth]{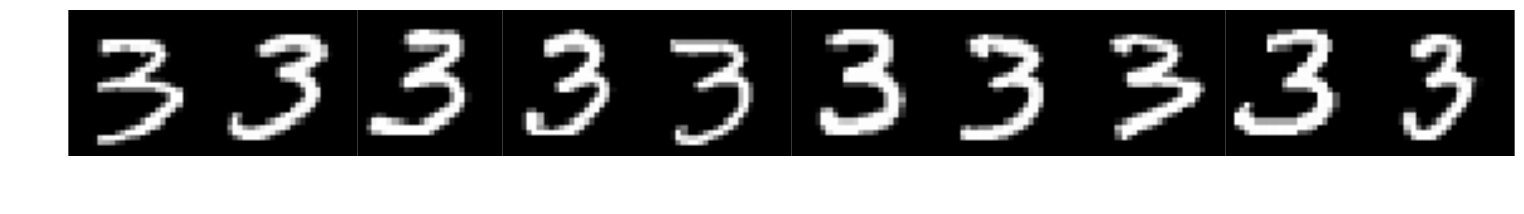} \\
  \vspace{-2mm}
  \includegraphics[trim={1em 0 0 0}, clip, width=\linewidth]{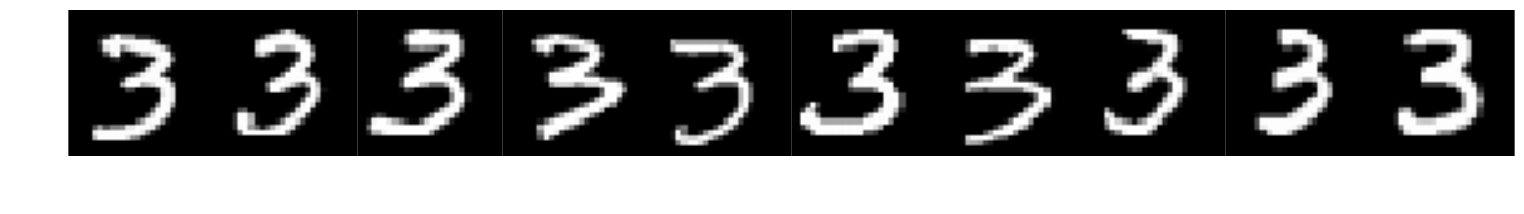}
\end{center}
\caption{Points with highest influence / loss (top) and lowest influence / loss (bottom).}
\end{wrapfigure}

\subsection{Understanding Policies} Finally, to give insight into the behavior
of the proposed polices, we examine the 10 points with highest influence/loss
vs.\ least influence/loss for MNIST\@.
We observe similar results  for the other
datasets (CIFAR10, NORB); additional results are provided in
Appendix~\ref{sec:diagnosing}. These examples help visualize the benefits of
downsampling, as it is clear that the bottom set of points are all quite
similar. The top points, in contrast, appear more diverse---both in terms of
class label as well as features (thin lines, thick lines, slanted, straight,
etc.). We postulate that promoting this diversity and removing redundancy is key in learning invariances through augmentation more efficiently.

\section{Discussion}
In this paper, we demonstrate that not all training points are equally useful for augmentation, and
we propose simple policies that can select the most viable subset of points. Our
policies, based on notions of training loss and model influence, are widely applicable to general machine learning models.
Obtaining access to an augmentation score vector can be obtained in only one training
cycle on the original data (e.g., a fixed cost), yet the potential improvements
in augmented training can scale superlinearly with respect to the original
dataset size.
With many fewer data points to augment, the augmentations themselves can be
applied in a more efficient manner in terms of compute and expert oversight.
At an extreme, they can be specialized on a per-example basis.

A natural area of future work is to explore subset selection policies that take the entire subset into account, rather than the greedy policies described.
For example, even if two samples may independently have large leave-one-out influence, it
may be the case that these points influence each other and leave-one-out influence
may be an overestimate (e.g., consider the case of two identical samples). 
Including second-order information or encouraging subset diversity%
\footnote{%
See Appendix~\ref{sec:sample_clustering} and~\ref{sec:dpp}
for preliminary work
investigating subset diversity.}
may therefore help to improve performance even further.

\ificlrfinal%
\subsubsection*{Acknowledgments}
We thank Tri Dao and Pang Wei Koh for their valuable discussions and feedback. This
material is based upon work supported by the National Defense Science and Engineering Graduate Fellowship.
\fi

\bibliography{ms}
\bibliographystyle{iclr2019_conference}

\newpage
\appendix

\section{Additional Plots: Metrics}%
\label{sec:histograms}

Here we provide histogram plots for loss and influence for all datasets. The key take-away from these results is that the distribution of these metrics indicate that most points have low loss and influence, and thus (according to our policies) can be augmented with low probability.

\begin{figure}[ht]
\begin{center}
  \subfloat[MNIST]{%
  \includegraphics[width=0.33\linewidth]{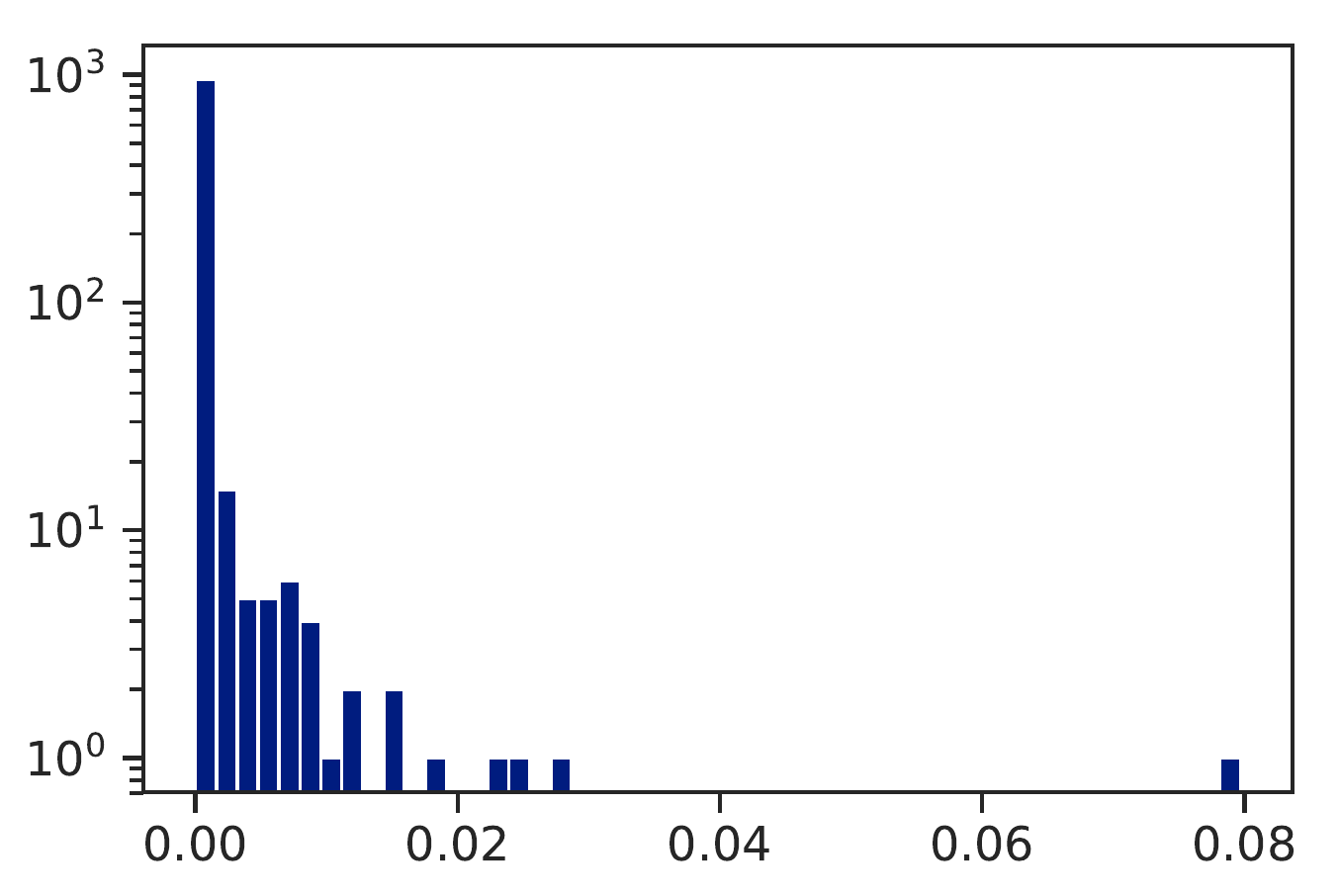}%
  }%
  \subfloat[CIFAR10]{%
  \includegraphics[width=0.33\linewidth]{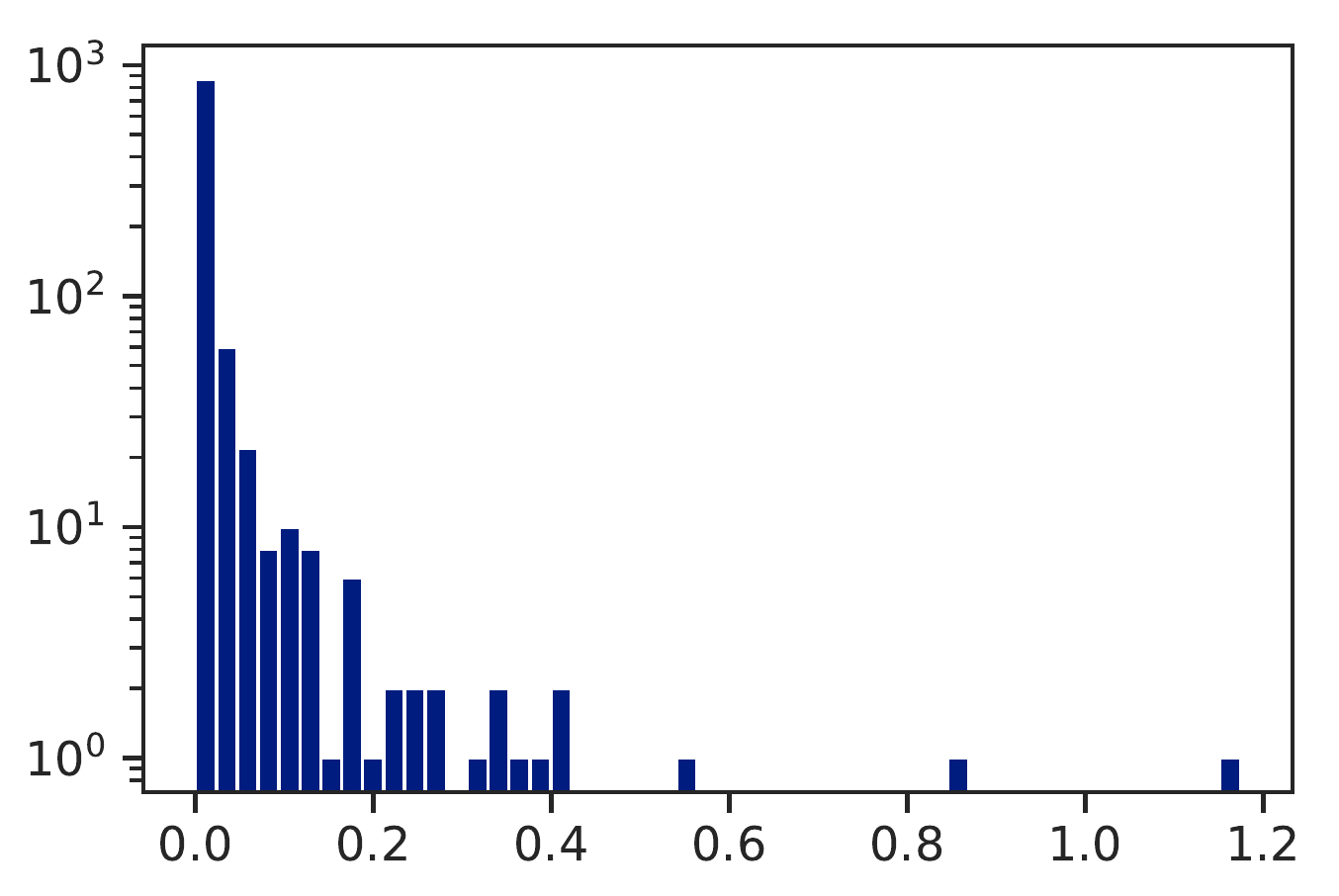}%
  }%
  \subfloat[NORB]{%
  \includegraphics[width=0.33\linewidth]{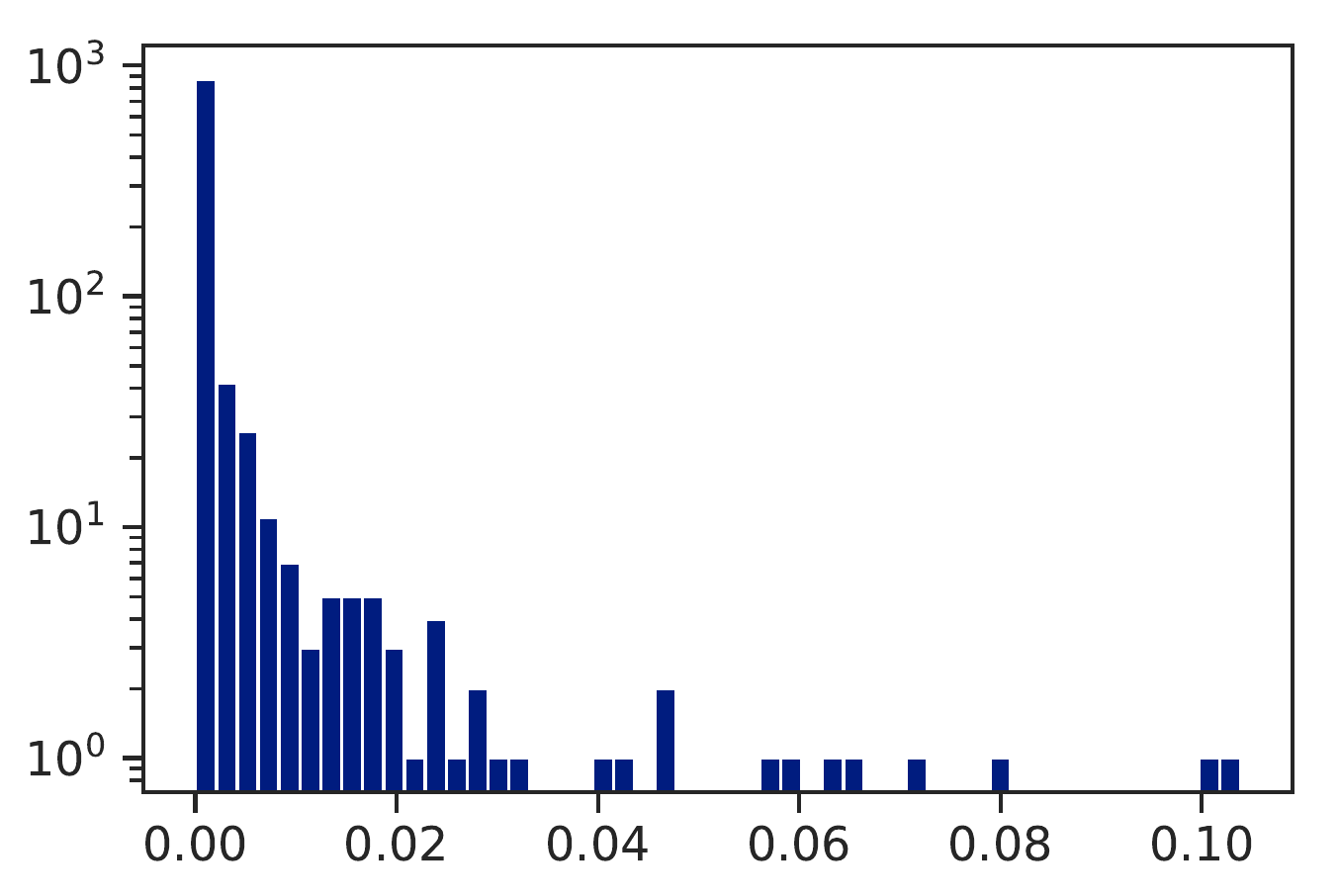}%
  }%
\end{center}
\caption{Distribution of log loss values on initial training set for translate
  augmentations. The distributions seem to have similar shape, but with
  different scales.
  Most values are not influential and can be augmented with low priority.}%
\label{fig:loss_hist_translate}%
\end{figure}

\begin{figure}[ht]
\begin{center}
  \subfloat[MNIST]{%
  \includegraphics[width=0.33\linewidth]{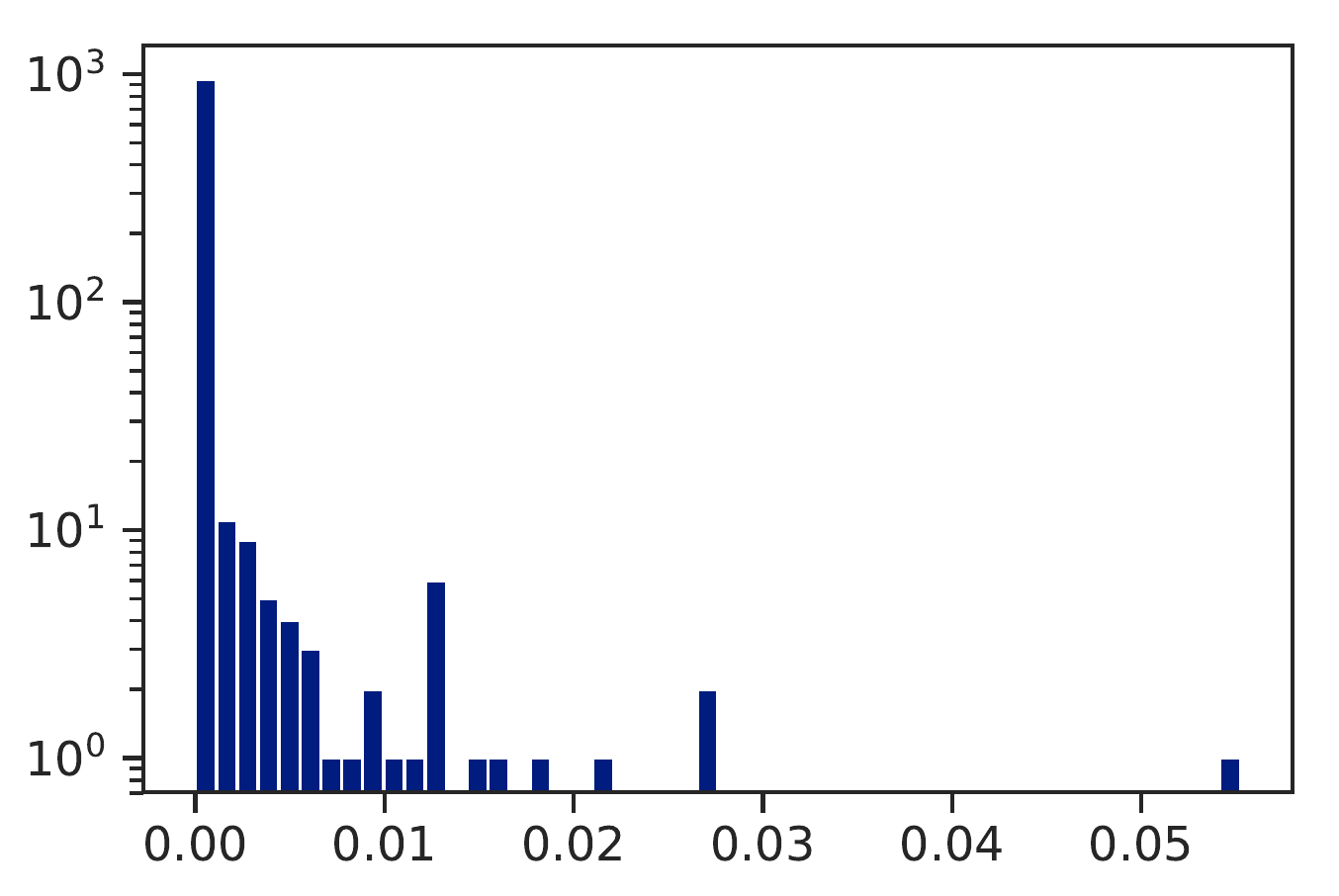}%
  }%
  \subfloat[CIFAR10]{%
  \includegraphics[width=0.33\linewidth]{aug_results_CIFAR10_0_vs_1_translate_10_abs_initial_aug_scores_histogram.pdf}%
  }%
  \subfloat[NORB]{%
  \includegraphics[width=0.33\linewidth]{aug_results_NORB_0_vs_1_translate_10_abs_initial_aug_scores_histogram.pdf}%
  }%
\end{center}
\caption{Distribution of influence values on initial training set for translate
  augmentations. The distributions seem to have similar shape, but with
  different scales.
  Most values are not influential and can be augmented with low priority.}%
\label{fig:full_inf_hist_translate}%
\end{figure}

\begin{figure}[ht]
\begin{center}
  \subfloat[MNIST]{%
  \includegraphics[width=0.34\linewidth]{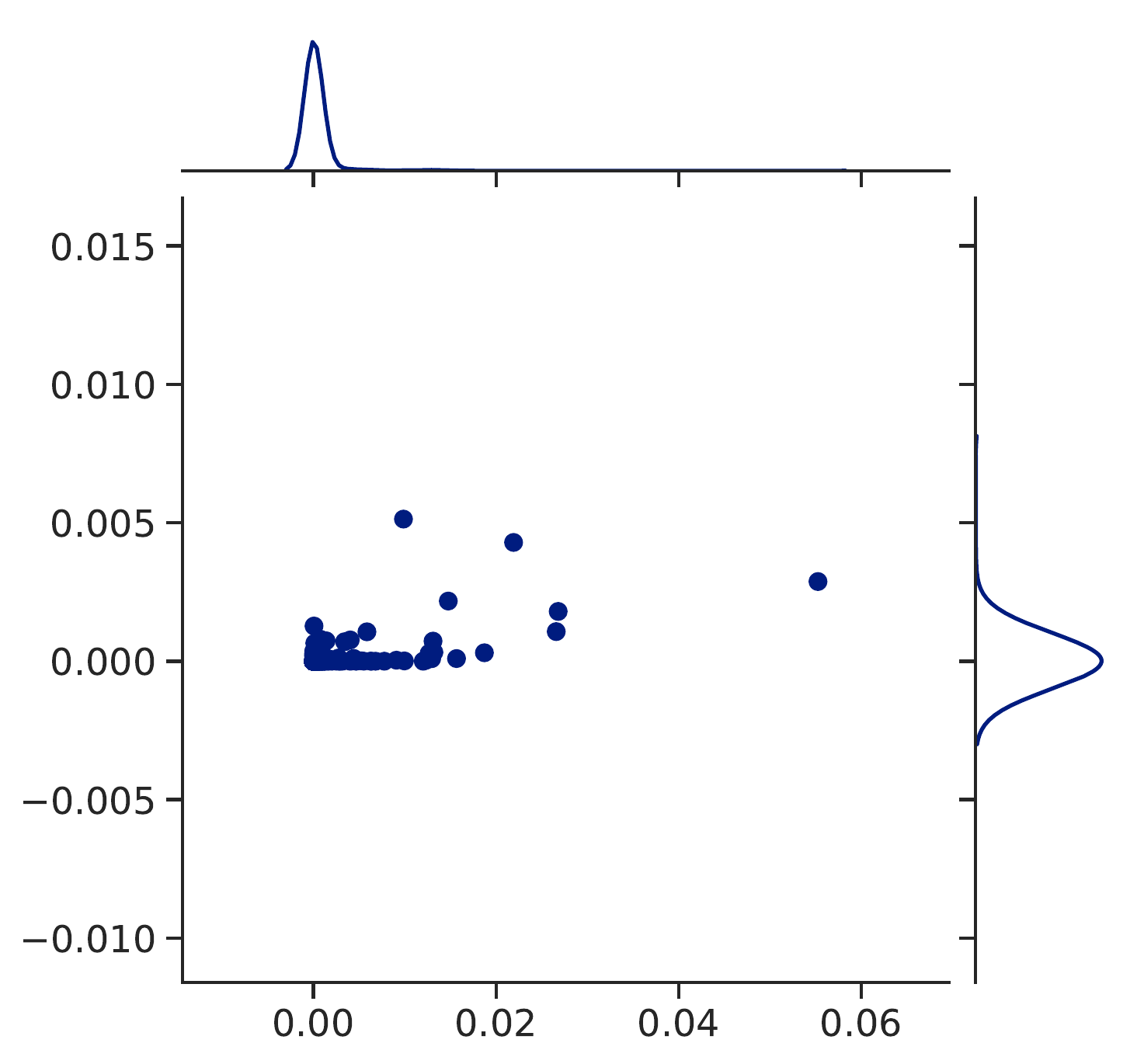}%
  }%
  \subfloat[CIFAR10]{%
  \includegraphics[width=0.31\linewidth]{aug_results_CIFAR10_0_vs_1_translate_10_init_after_joint_histogram.pdf}%
  }%
  \subfloat[NORB]{%
  \includegraphics[width=0.33\linewidth]{aug_results_NORB_0_vs_1_translate_10_init_after_joint_histogram.pdf}%
  }%
\end{center}
  \caption{Distribution of influence values on initial training set (x-axis)
  vs.\ final training set (y-axis) for translate augmentations.
  The distributions seem to have similar shape, but with
  different scales.
  Most values are not influential and can be augmented with low priority.
  Points that are not influential usually stay uninfluential.
  }%
\label{fig:full_inf_joint_translate}%
\end{figure}

\newpage
\section{Experiment Details}%
\label{sec:implementation}
Here we provide full implementation details on our experiments throughout the paper.

\textbf{Setup.} There are a few key architectural ideas in our tests: the data, the
augmentations, the selection policy, a featurization preprocessing component, and a logistic regression model. Our implementation is in Python.
The dataset is loaded (possibly via third party libraries) into a NumPy array.
We then run this dataset through a trained CNN model, such as LeNet \citep{lecun1998gradient} or ResNet50v2 \citep{he2016identity}, to obtain
a feature vector.
The logistic regression model is then trained on this resulting ``featurized''
dataset and tested on a ``featurized'' test set.
Once training is complete, both loss and influence can then be measured for each
training point, and can therefore be used as scores. Augmentations are then applied exhaustively to the test set.
We refer to this test set as ``poisoned''.
The test distribution has changed, and therefore a gap has formed between the
original test performance and the ``poisoned'' test performance.
We attempt to close this gap by applying augmentations to the training set. We
proceed by initializing a set with the un-augmented training set.
We augment points in rounds, and the un-augmented training set corresponds to round $0$.
Every round, our policy is given a vector of scores, and it selects a point to
augment.
This point is featurized and added to the set.
If score updates are enabled and the current round is sampled for testing, the
model scores are re-calculated for the original data points.
The CNN can be optionally retrained, but the logistic regression model must be
retrained to obtain the current test accuracy.
Each stochastic policy is tested $5$ times.
Plots show $95\%$ confidence intervals and fix $C=10$ for the logistic
regression hyperparameter.

\textbf{Implementation.} We perform experiments in Python, using Keras
\citep{chollet2015keras}, Tensorflow \citep{tensorflow}, Scikit-Learn
\citep{scikit-learn,sklearn_api}, AutoGrad \citep{maclaurin2015autograd}, and
Imgaug~\citep{imgaug}.
We wrap Keras implementations of the CNNs in Scikit-Learn transformers, and we
create new classes utilizing Scikit-Learn classifiers and their corresponding
influence functions calculated with the autograd system.
This allows us to decouple input data, bottleneck features, and the final
classifier that calculates influence.
It also allows us to perform any additional (i.e., cross validation) tuning
rather easily.
Augmentations are performed by Imgaug~\citep{imgaug}.
Our code is publicly available
 online\footnote{\url{https://github.com/mkuchnik/Efficient_Augmentation}}.

\textbf{Models.} %
For all experiments, we use a CNN to create bottleneck
features, which we then use as inputs into a linear logistic regression model.
This is equivalent to freezing the weights of the CNN, or using a set of basis
functions, $\phi_i(\cdot)$, to transform the inputs~\citep{bishop2006pattern}.
A LeNet architecture \citep{lecun1998gradient} with ReLu activations was used for
MNIST\@; however, this model had issues performing well on the
augmented sets for CIFAR10 and NORB\@.
We had also tried a larger model from the Keras
examples\footnote{\url{https://github.com/keras-team/keras/blob/master/examples/mnist_cnn.py}}
on MNIST, which
resulted in similar performance to using LeNet.
Both LeNet and the Keras neural network were fast to train, so we retrained the models for
$40-50$ epochs with
Adam \citep{kingma2014adam}
and a minibatch size of $512$, which was enough to obtain convergence.
We used a ResNet50v2 model \citep{he2016identity} model trained on the CIFAR10 dataset for the CIFAR10 tests, and we obtained good performance without
using augmentations in the training process.
Using a pretrained ImageNet ResNet50 model resulted in poor performance (both
computationally and in accuracy).
For NORB, we were able to obtain good performance on the translate task without any
training-time data augmentations being applied on the NORB dataset.
However, the other augmentations
resulted in high prediction degradation, so the ResNet model was retrained with
random rotations, shifts, and flips applied to images.
All ResNet models were frozen after the initial training.

\textbf{Datasets.} We convert each of the datasets into a binary classification
task. MNIST is $3$ vs.\ $8$, CIFAR10 is airplane vs.\ automobile, and NORB is
animal vs.\ human.
$1000$ training examples are sampled from the resulting binary classification
problem.
The MNIST train class split is 517/483, and its test class split is 1010/974.
The CIFAR10 train class split is 523/477, and its test class split is 1000/1000.
THE NORB train class split is 500/500, and its test class split is 4860/4860.

\textbf{Augmentations.} Our tests use translate, rotate, and crop. %
Each of these augmentations is applied over a range of parameters, which results
in multiple augmented images.
Translate is applied for $2$ pixels in all cardinal directions (e.g., up, down,
left, and right) on MNIST, $3$ pixels for CIFAR10, and $6$ pixels for
NORB (note: this pixel difference is to account for NORB images being $3$ times larger than CIFAR10).
Rotate is applied for the $15$ ($14$ after removing identity transform) rotations evenly spaced
between $\pm30\degree$ for MNIST\@.
CIFAR10 and NORB use $\pm5\degree,\pm2.5\degree$.
For MNIST, crop is applied excluding the outer $[1,2,\dots,6]$ pixels on all $4$ image sides,
and zoom is applied to rescale the resulting image back to its original
dimensions.
CIFAR10 and NORB exclude the outer 2 pixels.
Usually, augmentations are constructed to preserve labels, but it is possible in
principle to construct augmentations that utilize label information for the
augmentation itself or perhaps induce a change in label (e.g., an image dataset with
segmentation information can segment out all non-background classes to change
the label of an image to background).
Such augmentations are expensive, require domain expertise, and are hard to
validate, but they may be viable if the number of total augmentations can be
controlled.

\section{Additional Plots: Policies}%
\label{sec:fullplots}

\begin{figure}[h!]
\begin{center}
  \subfloat[MNIST-translate]{%
  \trimfigleft{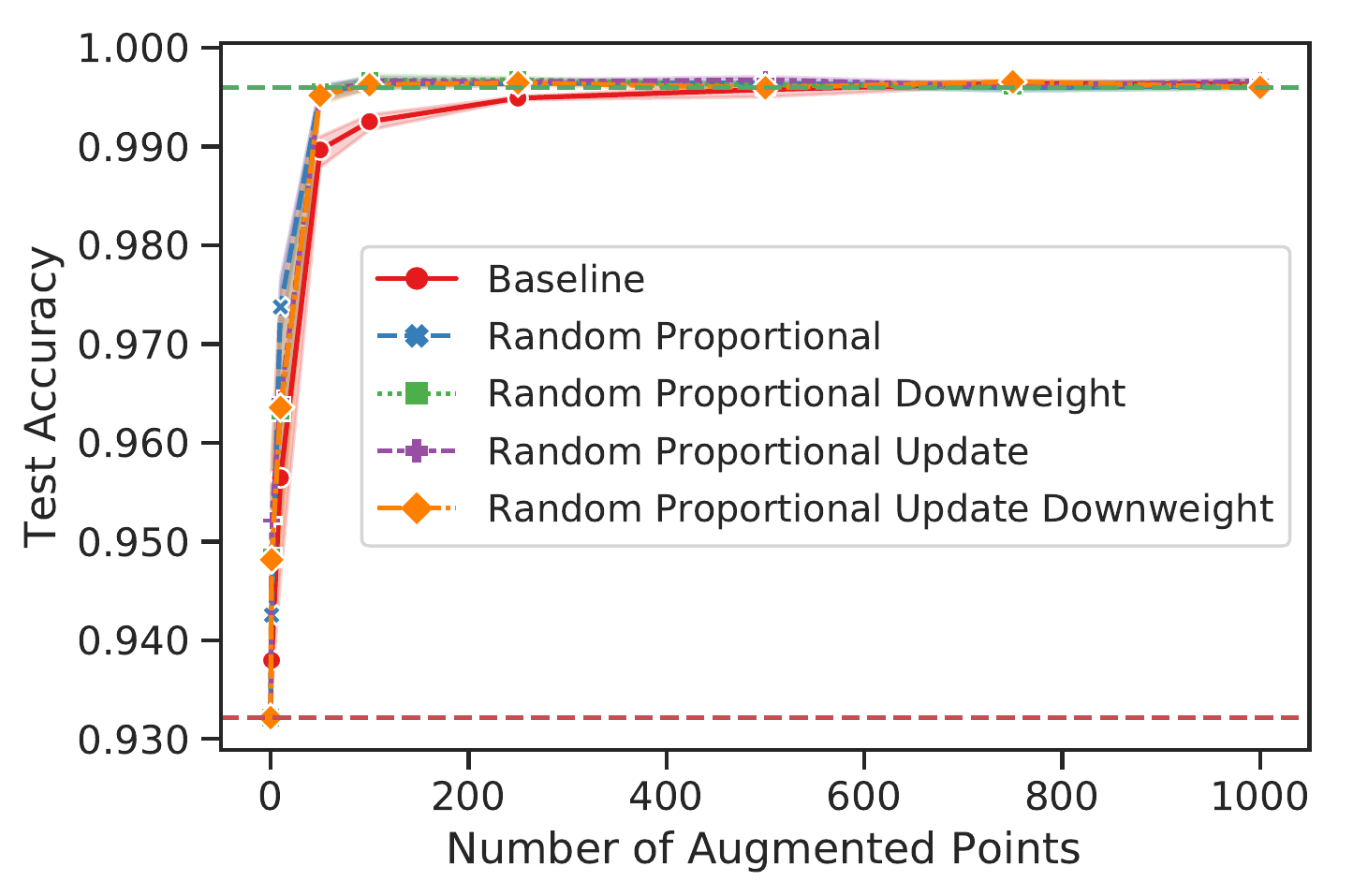}
  }%
   \subfloat[CIFAR10-translate]{%
  \trimfig{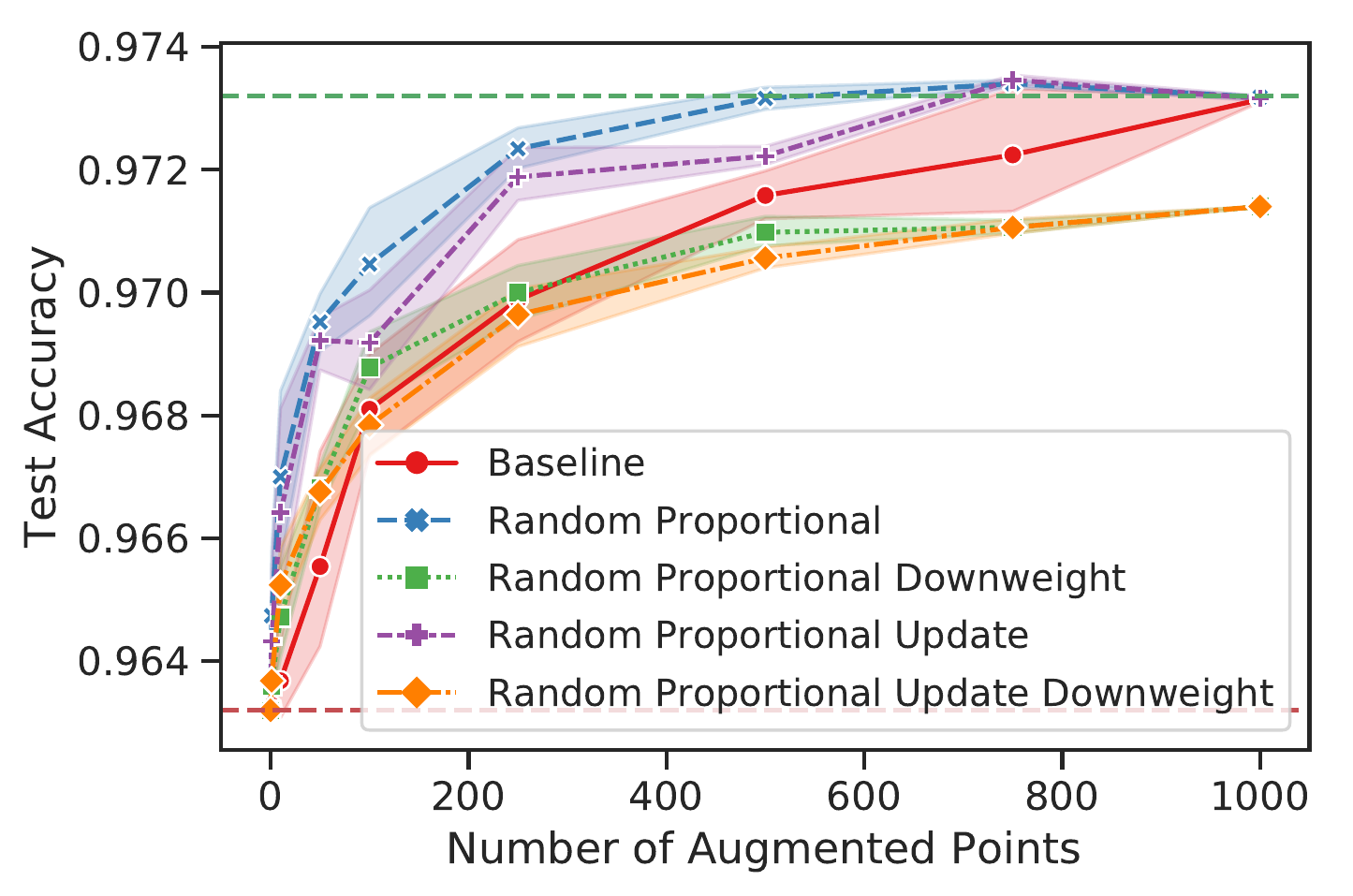}
  }%
    \subfloat[NORB-translate]{%
  \trimfig{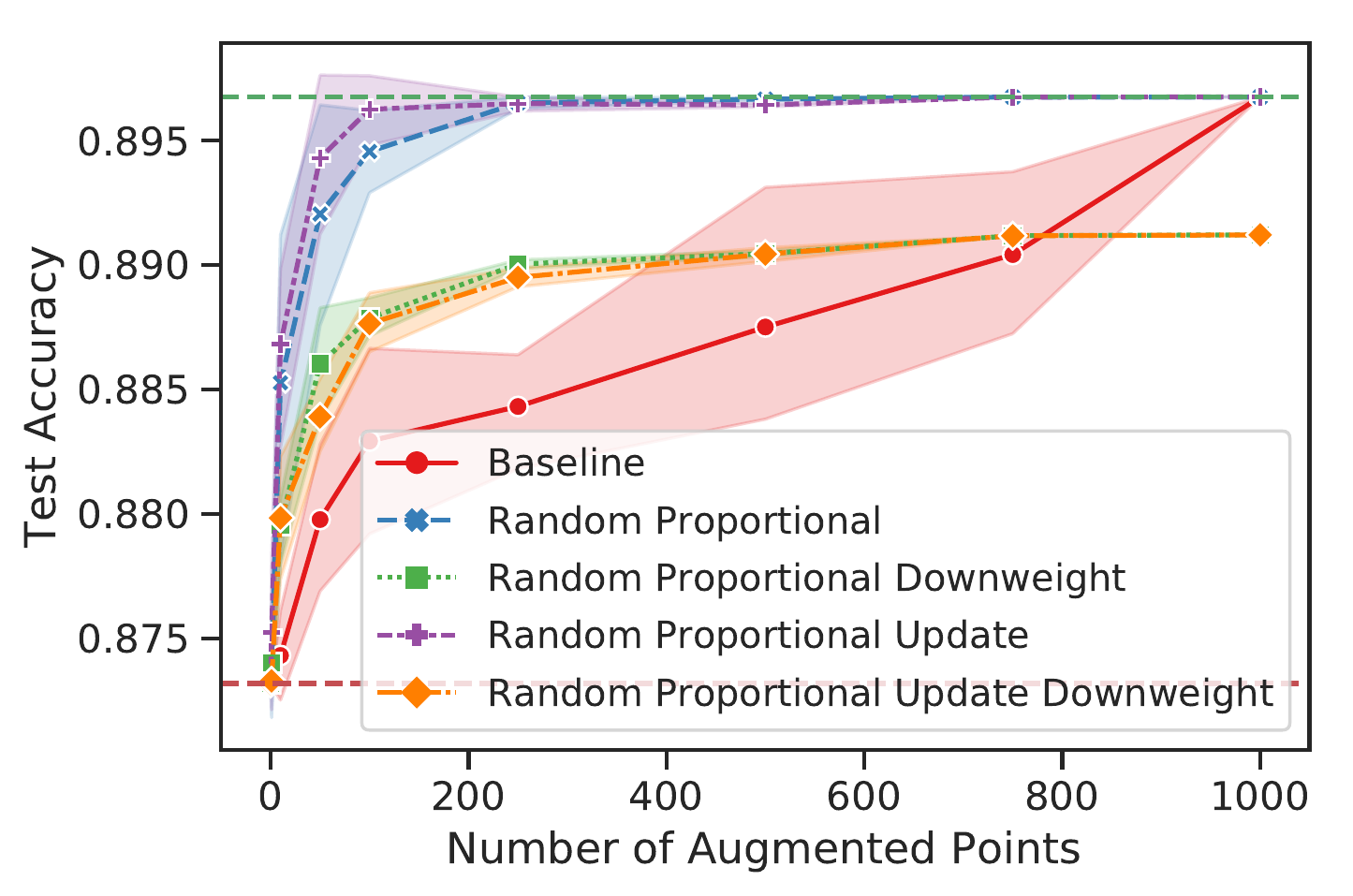}
  }%
  \newline
  \subfloat[MNIST-rotate]{%
  \trimfigleft{aug_results_MNIST_3_vs_8_rotate_10_modifications_accuracy.pdf}
  }%
    \subfloat[CIFAR10-rotate]{%
  \trimfig{aug_results_CIFAR10_0_vs_1_rotate_10_modifications_accuracy.pdf}
  }%
    \subfloat[NORB-rotate]{%
  \trimfig{aug_results_NORB_0_vs_1_rotate_10_modifications_accuracy.pdf}
  }%
  \newline
  \subfloat[MNIST-crop]{%
  \trimfigleft{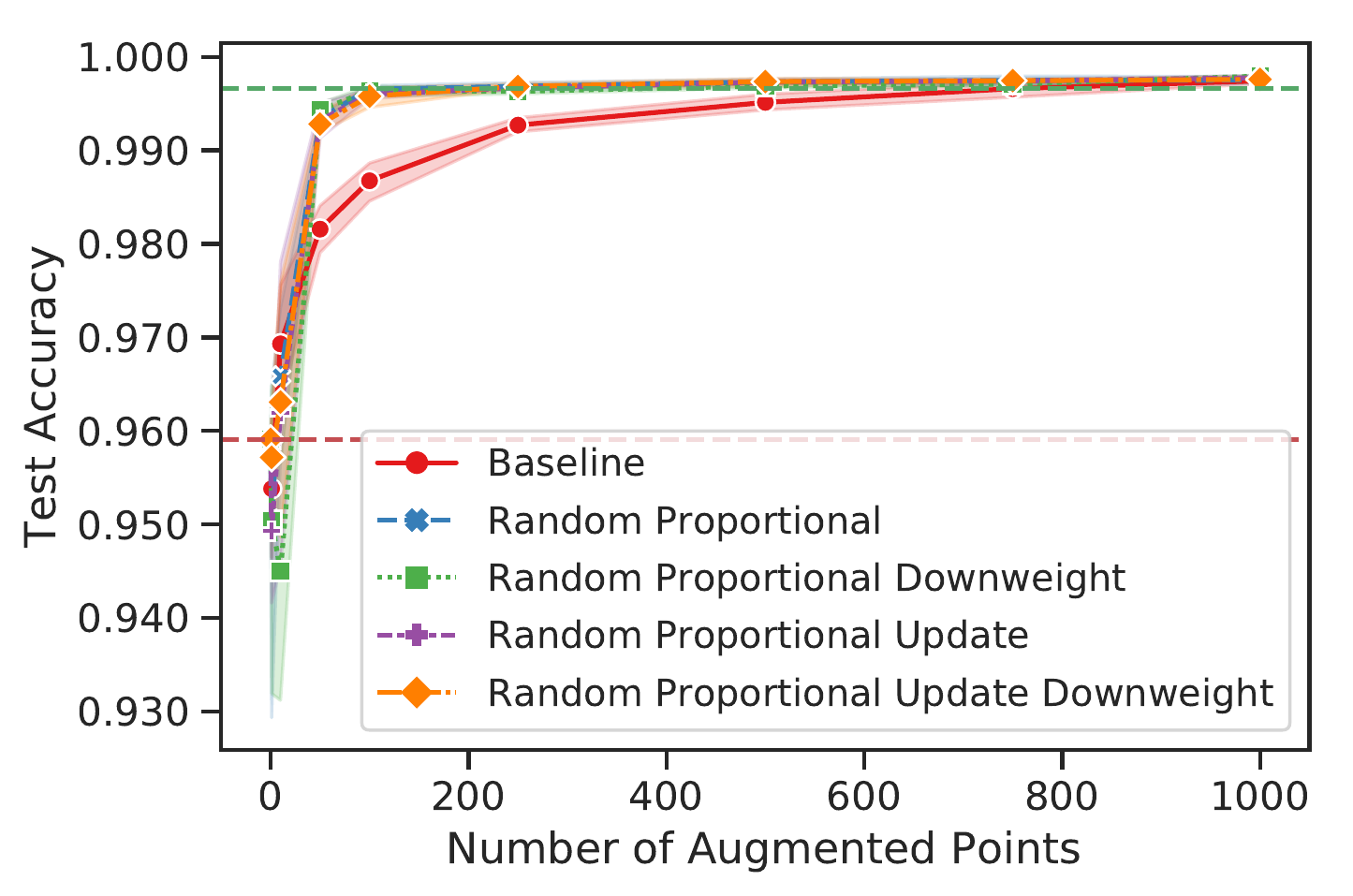}
  }%
  \subfloat[CIFAR10-crop]{%
  \trimfig{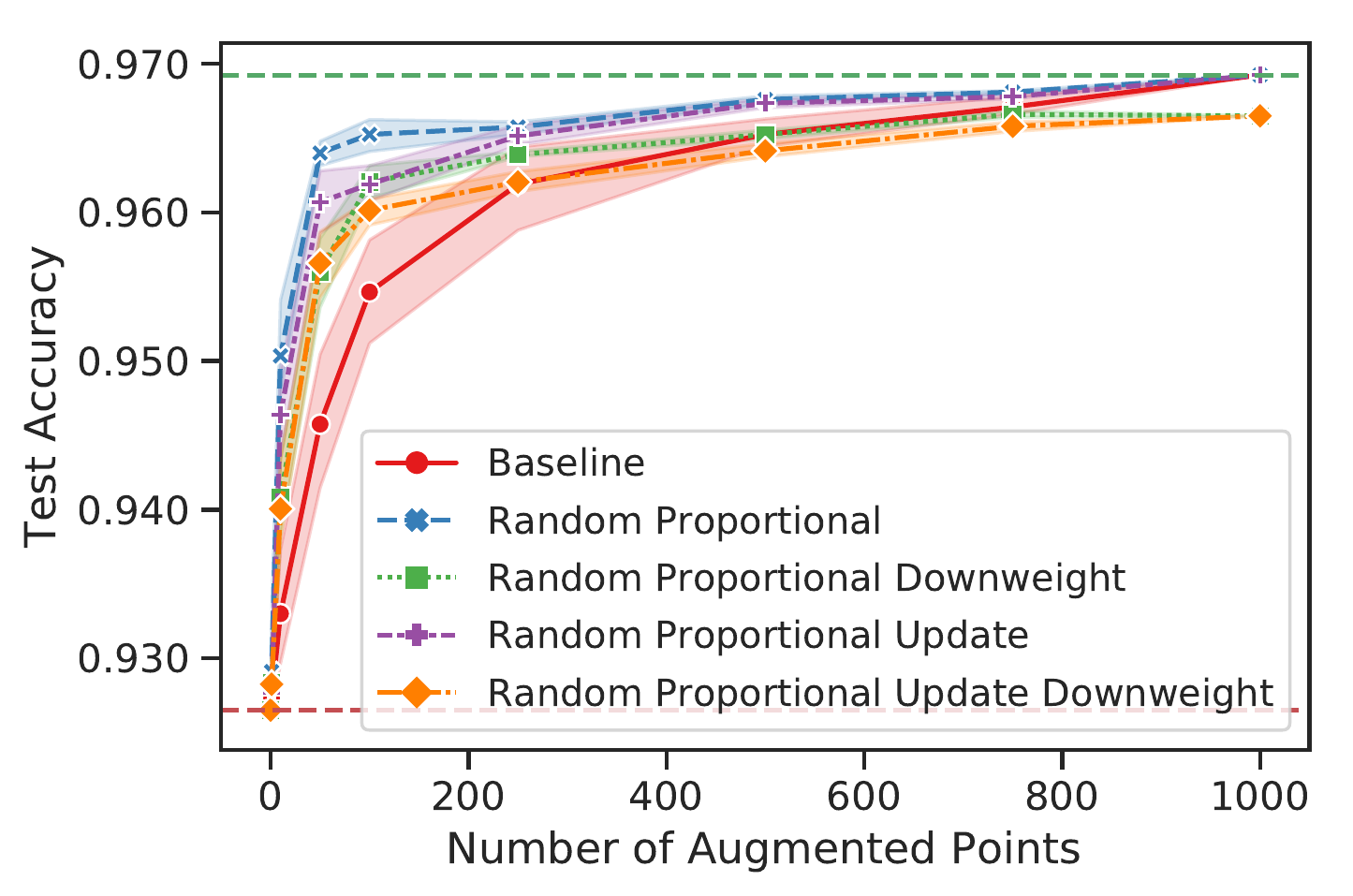}
  }%
  \subfloat[NORB-crop]{%
  \trimfig{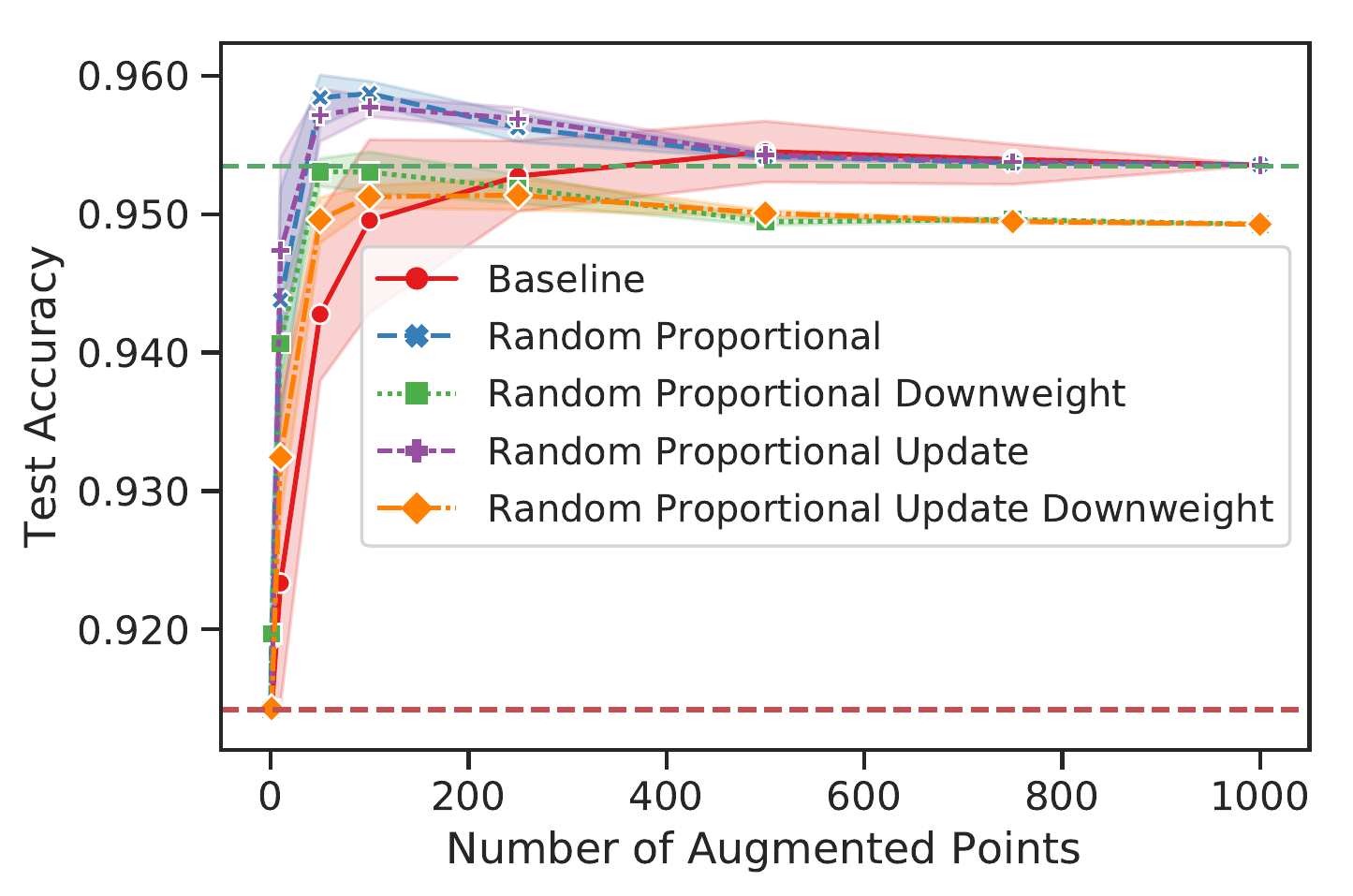}
  }%
  \caption{The performance of randomized policies using influence.}%
\label{fig:all_mod}%
\end{center}
\end{figure}

\begin{figure}[H]
\begin{center}
  \subfloat[MNIST-translate]{%
  \trimfigleft{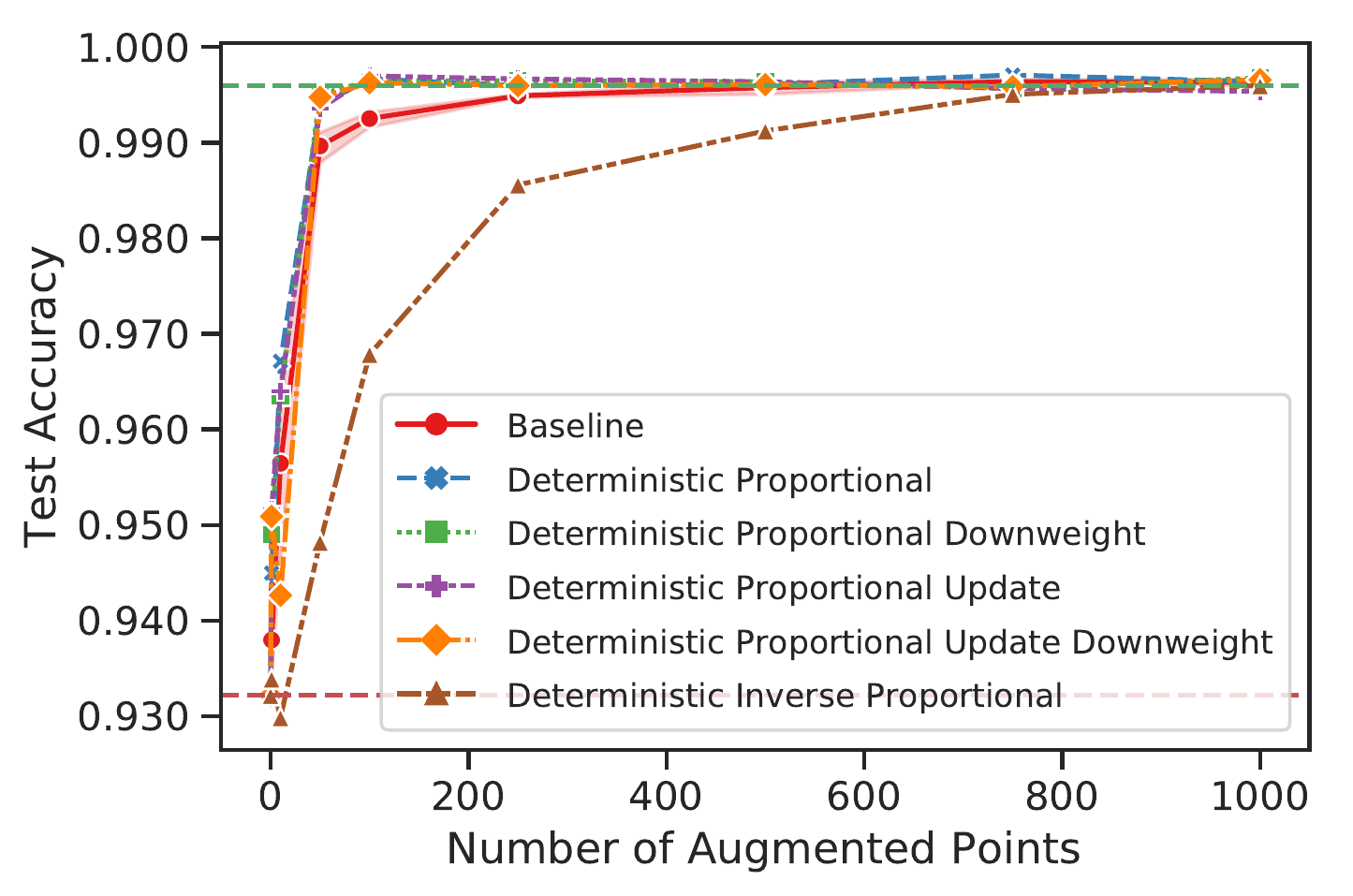}
  }%
    \subfloat[CIFAR10-translate]{%
  \trimfig{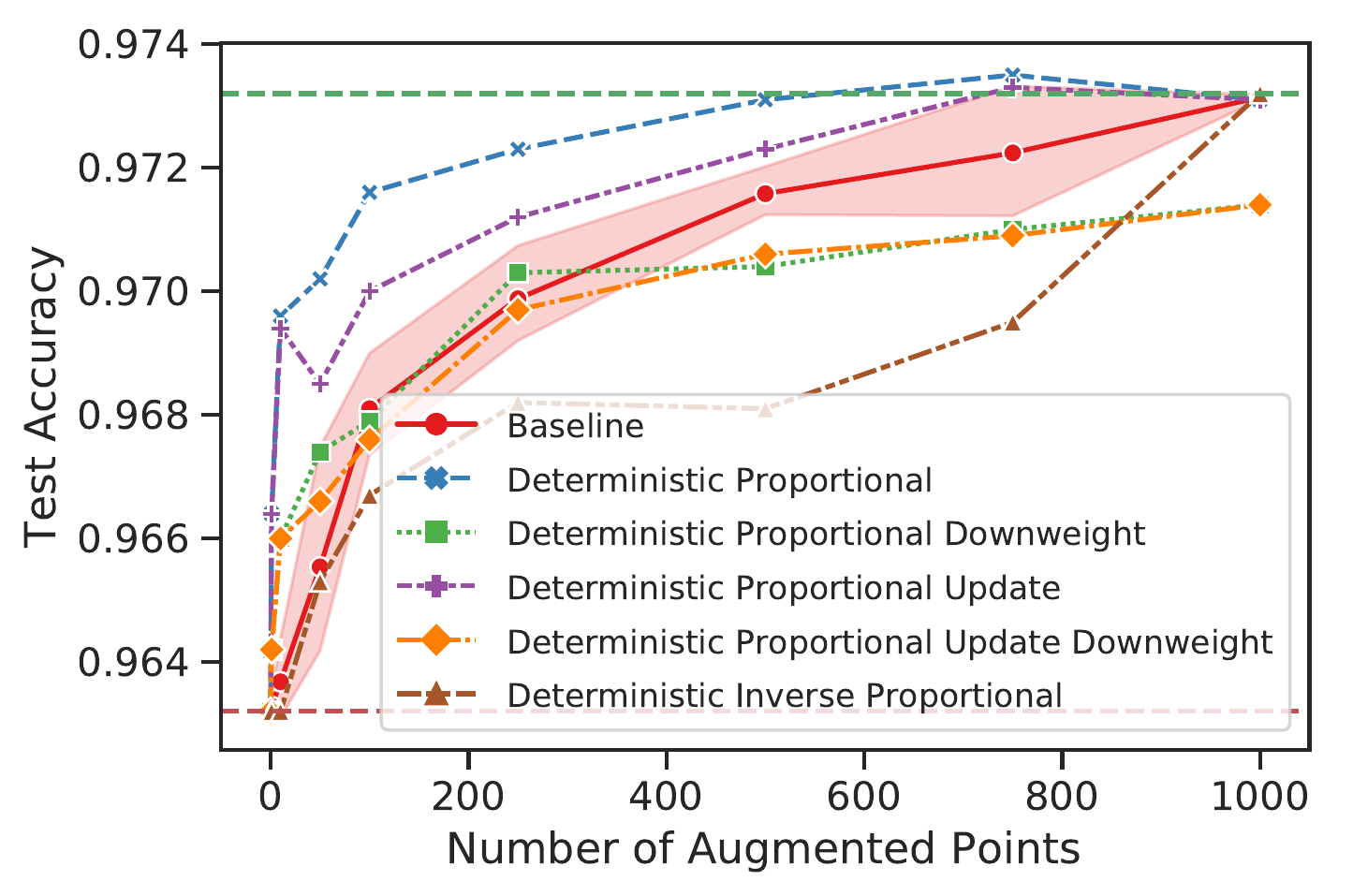}
  }%
    \subfloat[NORB-translate]{%
  \trimfig{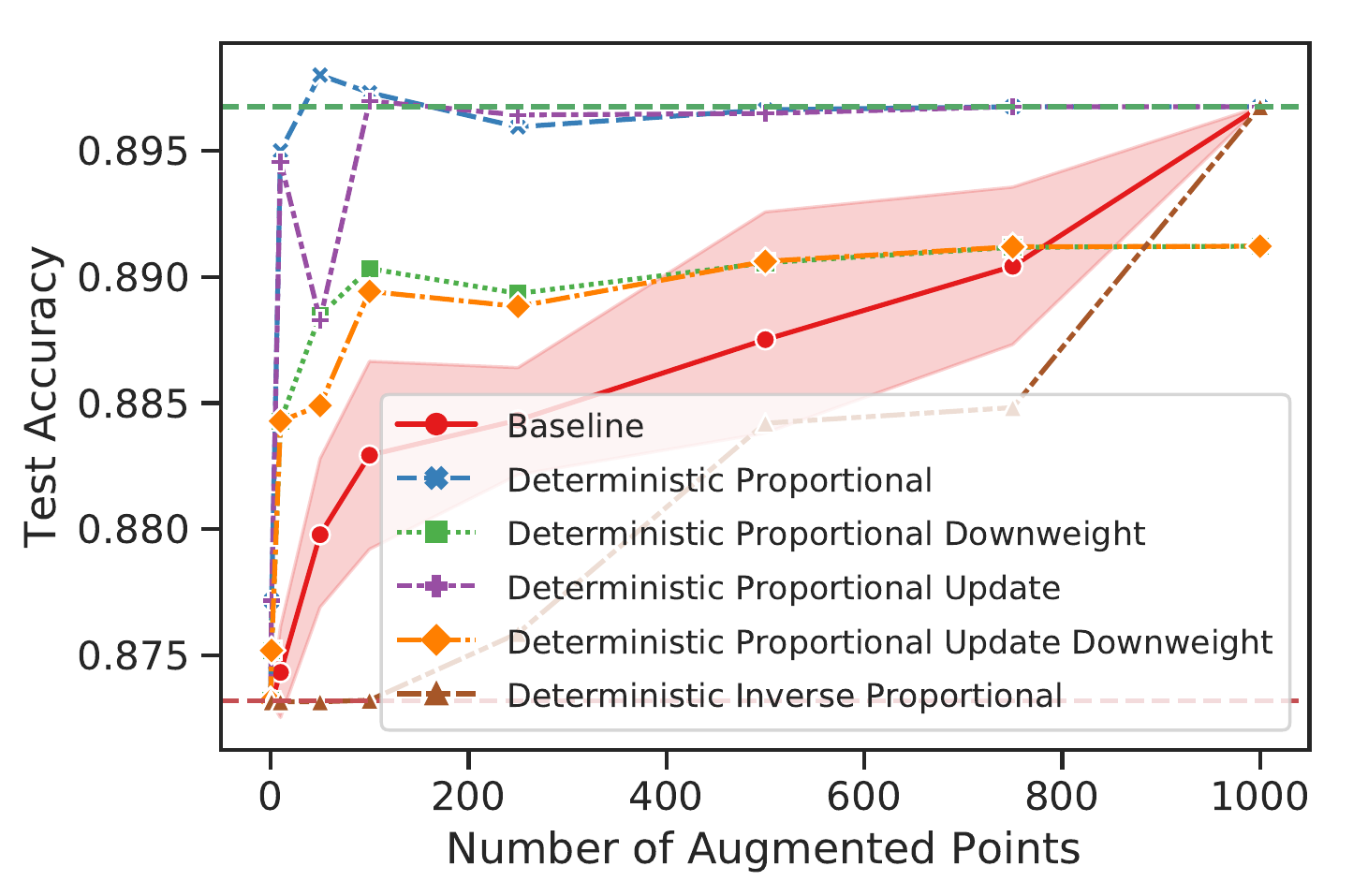}
  }%
  \newline
  \subfloat[MNIST-rotate]{%
  \trimfigleft{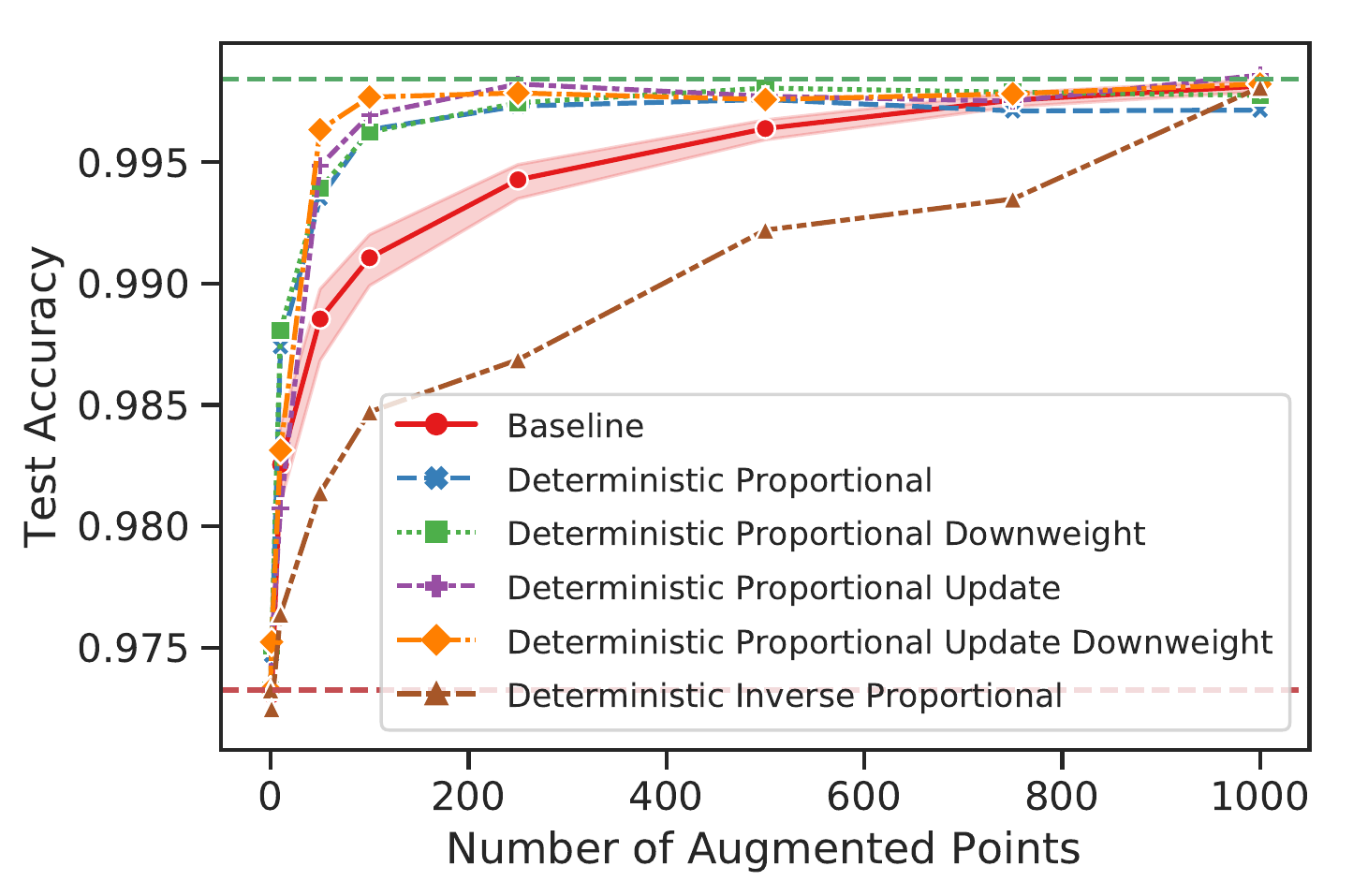}
  }%
    \subfloat[CIFAR10-rotate]{%
  \trimfig{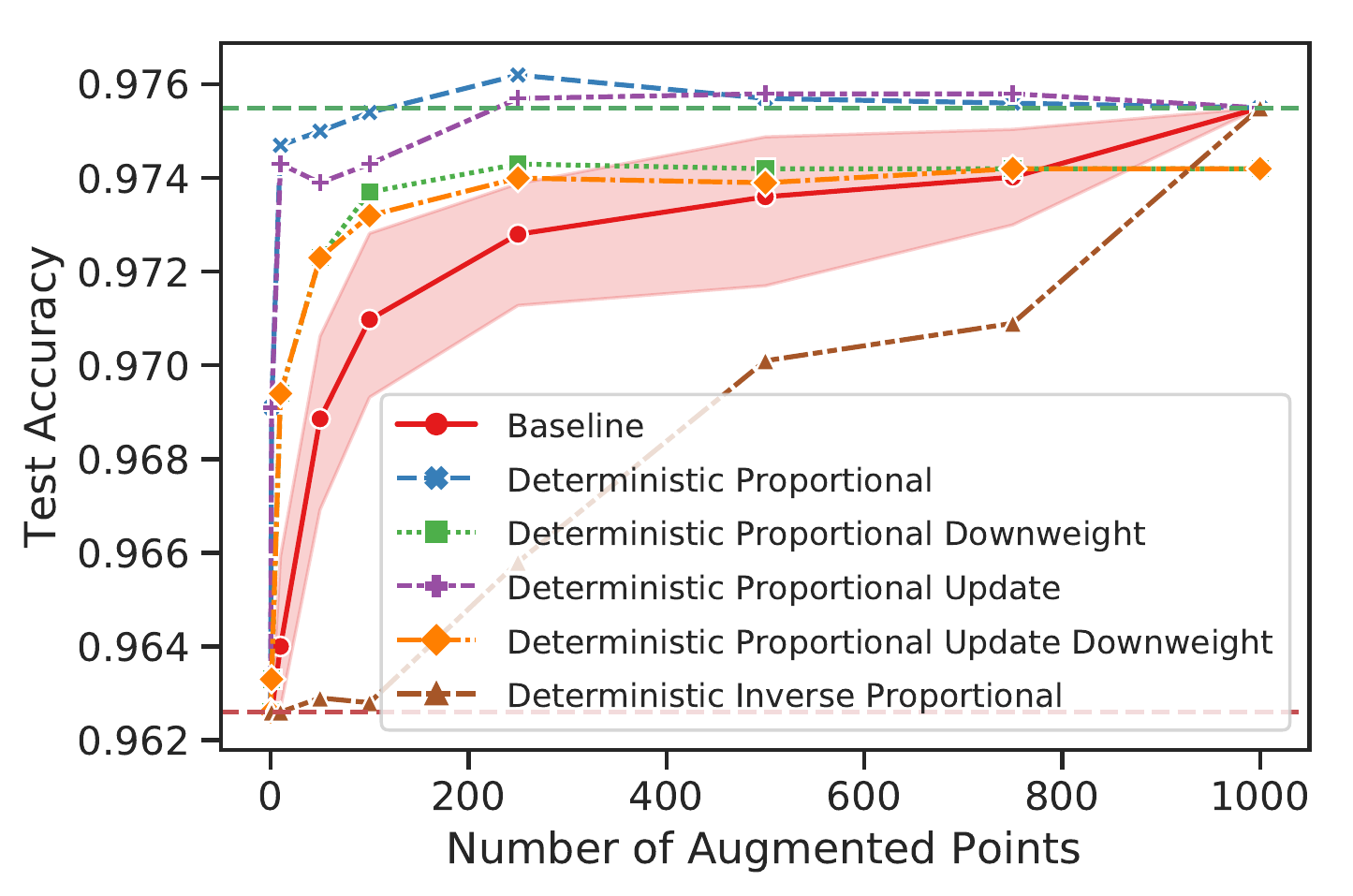}
  }%
  \subfloat[NORB-rotate]{%
  \trimfig{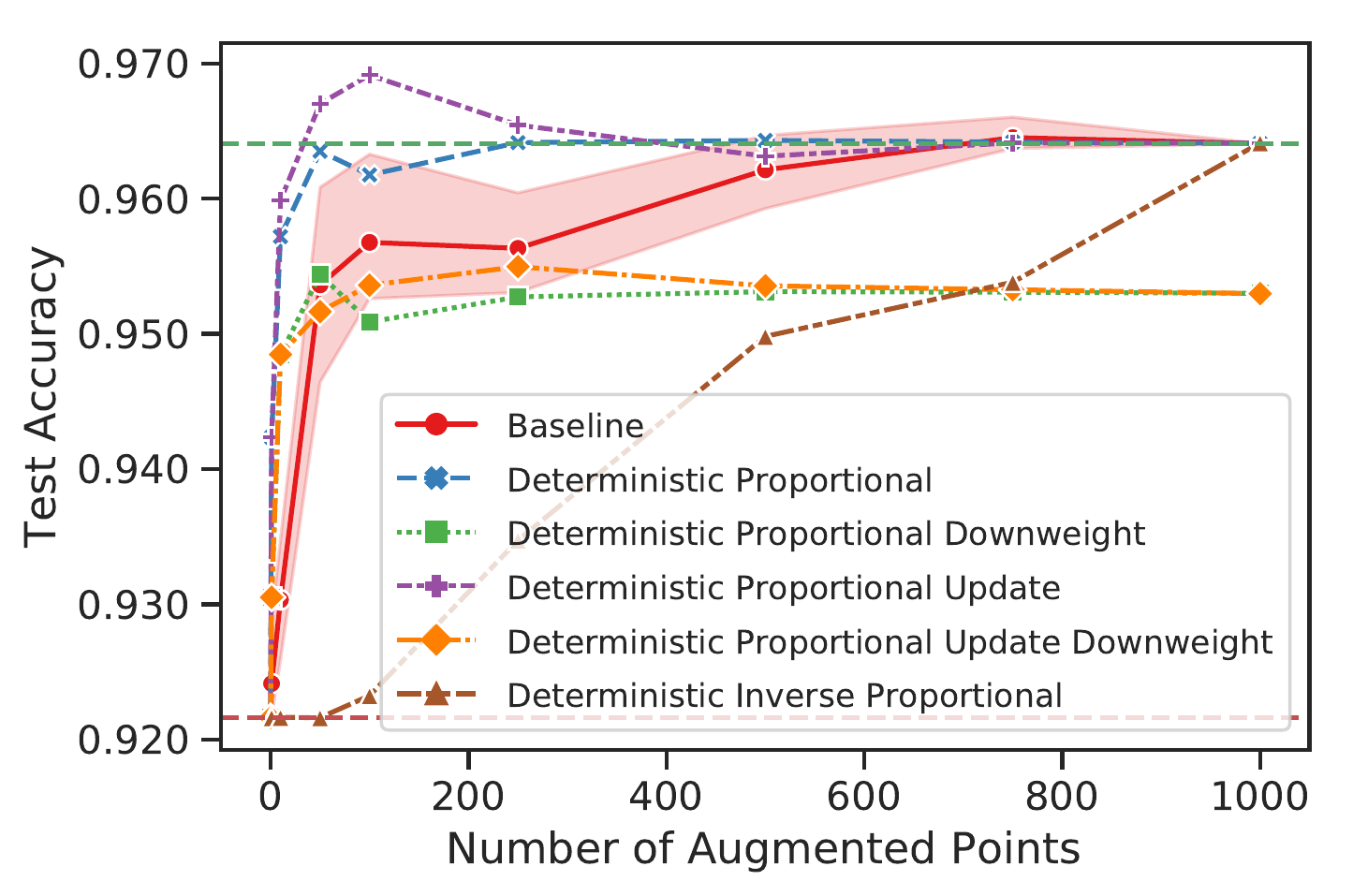}
  }%
  \newline
  \subfloat[MNIST-crop]{%
  \trimfigleft{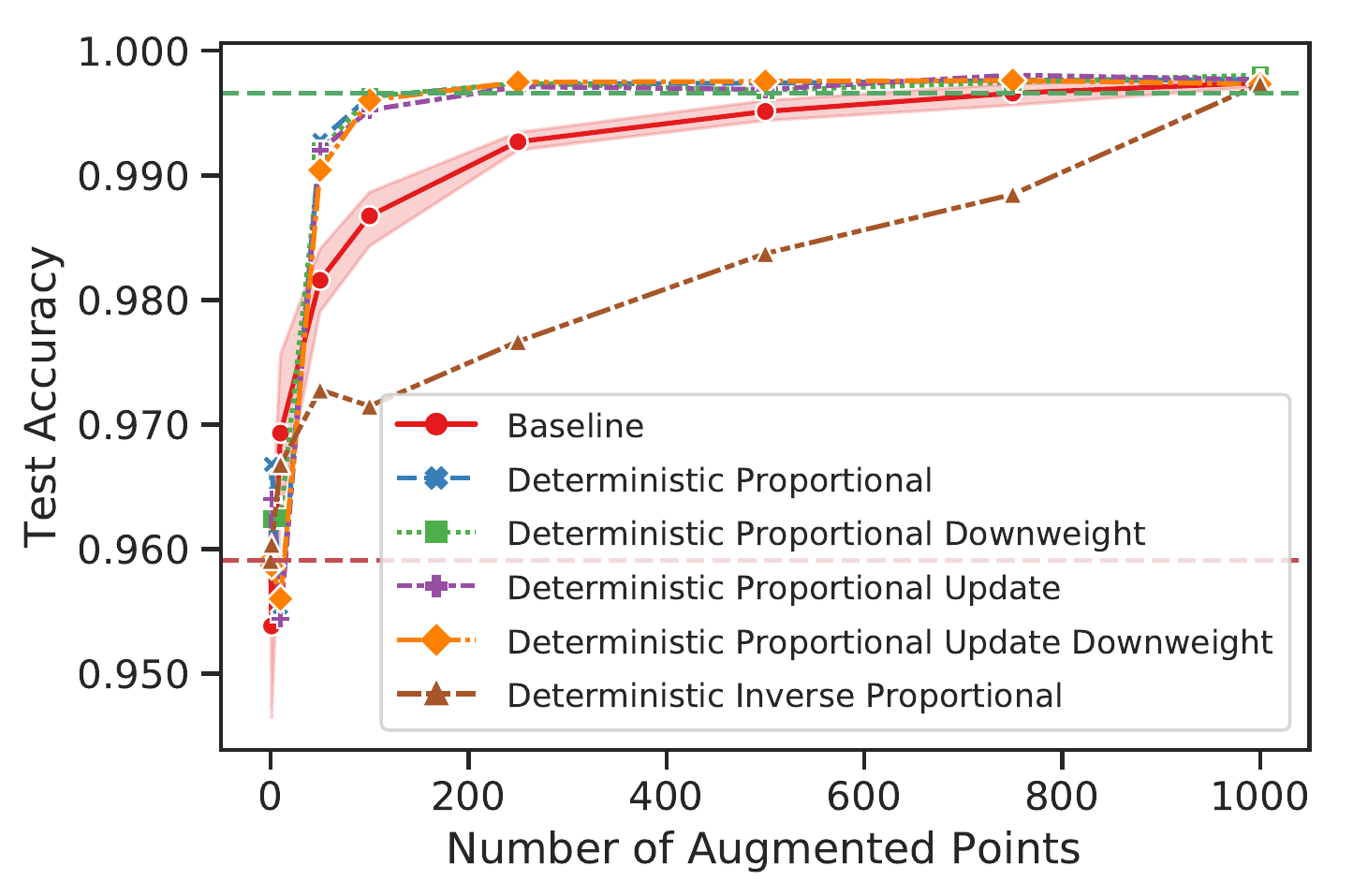}
  }%
  \subfloat[CIFAR10-crop]{%
  \trimfig{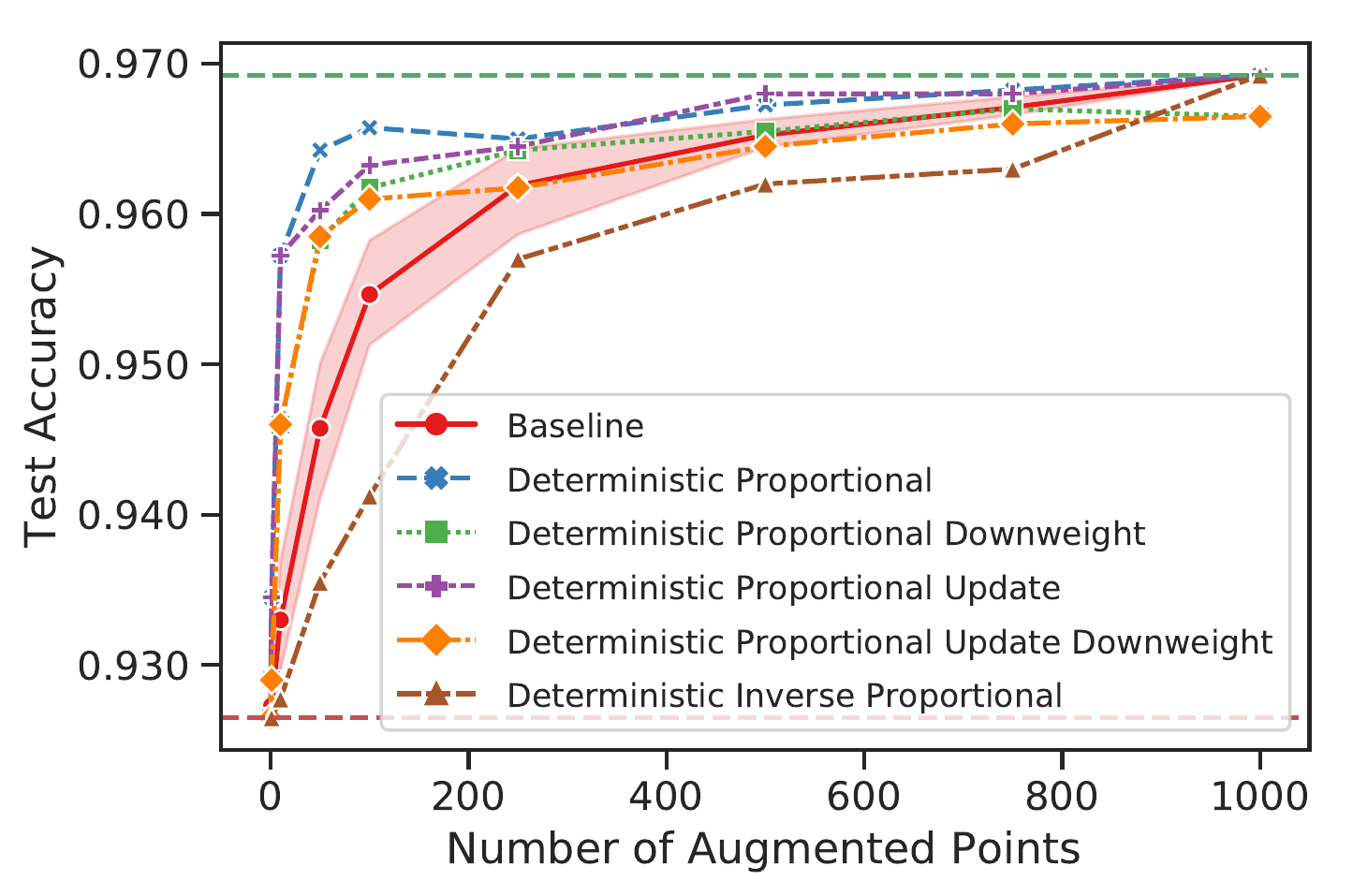}
  }%
  \subfloat[NORB-crop]{%
  \trimfig{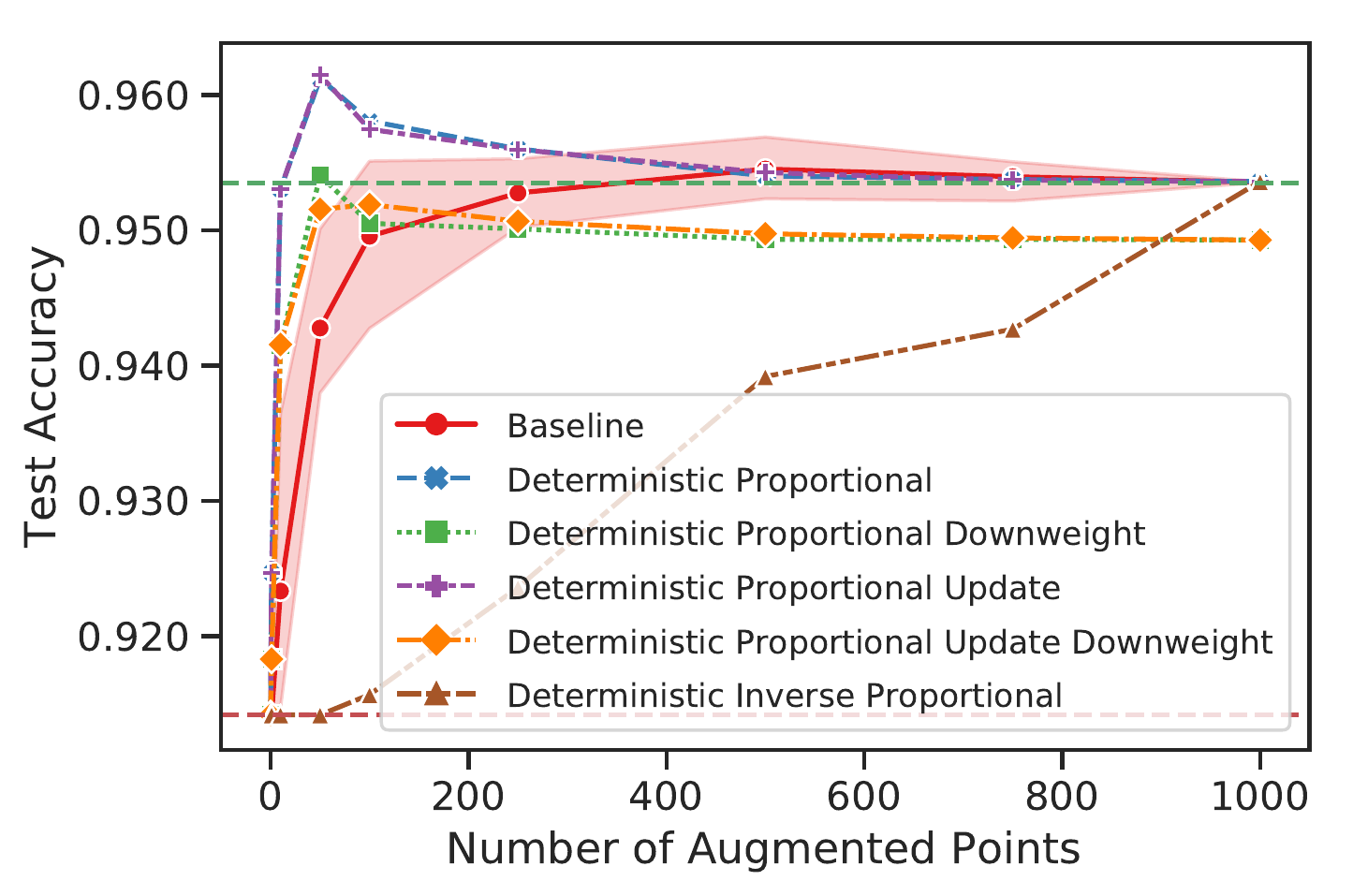}
  }%
  \caption{The performance of deterministic policies using influence.}%
\label{fig:all_det}%
\end{center}
\end{figure}

Below we provide full experiments for the randomized (Figure~\ref{fig:all_mod}) and deterministic (Figure~\ref{fig:all_det}) policies using model influence as the scoring metric across all datasets and transformations.
Please see Appendix~\ref{sec:fulltables} for tables listing Area Under the
Curve (AUC) statistics.

\newpage
\section{Full Results: Policies}%
\label{sec:fulltables}

Below we provide Area Under the Curve (AUC) results from the plots provided in
Appendix~\ref{sec:fullplots} as well as the corresponding results for loss.
Compared to the plots, the AUC provides a single metric by which to
determine the best performing policy. We separate tables across augmentation
(translate/rotate/crop) and score function (loss/influence).

In the tables below, we make the following abbreviations corresponding to the policies described in Section~\ref{sec:policies}:
Det (Deterministic), Prop (Proportional), Down (Downweight), and Rand (Random).

\newcommand{\tablesize}{0.33}
\newcommand{\hspacesize}{\hspace{0em}}
\subsection{Area Under Curve (AUC) Results using Loss}
\begin{table}[ht]
\makebox[\linewidth]{%
\hspacesize%
\begin{minipage}{\tablesize\textwidth}
\captionsetup{font=scriptsize}
\begin{center}
\resizebox{\textwidth}{!}{%
\begin{tabular}{l*{2}{S}}
\toprule
                                     {Policy} &  {AUC Mean} &     {AUC Std.} \\
\midrule
            Det Prop Update &  995.442440 &  {\textemdash} \\
 Det Prop Update Down &  995.375302 &  {\textemdash} \\
        Rand Prop Update Down &  995.372933 &       0.350032 \\
                          Rand Prop &  995.304607 &       0.241679 \\
                   Det Prop &  995.239113 &  {\textemdash} \\
                   Rand Prop Update &  995.230565 &       0.267833 \\
        Det Prop Down &  995.166633 &  {\textemdash} \\
               Rand Prop Down &  995.153468 &       0.154085 \\
                                     Baseline &  993.598992 &       0.354964 \\
                  Rand Inverse Prop &  985.775242 &       0.655752 \\
           Det Inverse Prop &  984.249143 &  {\textemdash} \\
\bottomrule
\end{tabular}
   }
\caption{AUC statistics: MNIST Translate.}%
\label{table:auc_mnist_translate_loss}%
\end{center}
\end{minipage}
\begin{minipage}{\tablesize\textwidth}
\captionsetup{font=scriptsize}
\begin{center}
\resizebox{\textwidth}{!}{%
\begin{tabular}{l*{2}{S}}
\toprule
                                     {Policy} & {AUC Mean} &     {AUC Std.} \\
\midrule
                   Det Prop &  972.55530 &  {\textemdash} \\
                          Rand Prop &  972.42286 &       0.118743 \\
            Det Prop Update &  971.97890 &  {\textemdash} \\
                   Rand Prop Update &  971.87607 &       0.112116 \\
                                     Baseline &  970.58856 &       0.670877 \\
        Det Prop Down &  970.16260 &  {\textemdash} \\
               Rand Prop Down &  969.97975 &       0.131681 \\
 Det Prop Update Down &  969.93160 &  {\textemdash} \\
        Rand Prop Update Down &  969.85123 &        0.24464 \\
                  Rand Inverse Prop &  969.29076 &        0.33845 \\
           Det Inverse Prop &  968.78950 &  {\textemdash} \\
\bottomrule
\end{tabular}
   }
\caption{AUC statistics:  CIFAR10 Translate.}%
\label{table:auc_cifar10_translate_loss}
\end{center}
\end{minipage}
\begin{minipage}{\tablesize\textwidth}
\captionsetup{font=scriptsize}
\begin{center}
\resizebox{\textwidth}{!}{%
\begin{tabular}{l*{2}{S}}
\toprule
                                     {Policy} &  {AUC Mean} &     {AUC Std.} \\
\midrule
                   Det Prop &  896.517078 &  {\textemdash} \\
            Det Prop Update &  896.095175 &  {\textemdash} \\
                          Rand Prop &  895.900965 &       0.503207 \\
                   Rand Prop Update &  895.792543 &       0.685641 \\
        Det Prop Down &  890.209887 &  {\textemdash} \\
               Rand Prop Down &  890.043979 &       0.299415 \\
 Det Prop Update Down &  889.867706 &  {\textemdash} \\
        Rand Prop Update Down &  889.492327 &       0.151238 \\
                                     Baseline &  888.851792 &        1.87646 \\
           Det Inverse Prop &  882.326872 &  {\textemdash} \\
                  Rand Inverse Prop &  881.991451 &       0.572188 \\
\bottomrule
\end{tabular}
}
\caption{AUC statistics:  NORB Translate.}%
\label{table:auc_norb_translate_loss}
\end{center}
\end{minipage}
}
\end{table}

\begin{table}[ht]
\makebox[\linewidth]{%
\hspacesize%
\begin{minipage}{\tablesize\textwidth}
\captionsetup{font=scriptsize}
\begin{center}
\resizebox{\textwidth}{!}{%
\begin{tabular}{l*{2}{S}}
\toprule
                                     {Policy} &  {AUC Mean} &     {AUC Std.} \\
\midrule
                   Det Prop &  997.285047 &  {\textemdash} \\
        Rand Prop Update Down &  997.266237 &       0.151984 \\
                   Rand Prop Update &  997.245585 &       0.167294 \\
               Rand Prop Down &  997.215262 &        0.19317 \\
                          Rand Prop &  997.076478 &       0.212783 \\
            Det Prop Update &  996.966179 &  {\textemdash} \\
        Det Prop Down &  996.913995 &  {\textemdash} \\
 Det Prop Update Down &  996.846892 &  {\textemdash} \\
                                     Baseline &  995.393290 &       0.636857 \\
                  Rand Inverse Prop &  990.824483 &       0.466477 \\
           Det Inverse Prop &  989.940827 &  {\textemdash} \\
\bottomrule
\end{tabular}
}
\caption{AUC statistics:  MNIST Rotate.}%
\label{table:auc_mnist_rotate_loss}
\end{center}
\end{minipage}
\begin{minipage}{\tablesize\textwidth}
\captionsetup{font=scriptsize}
\begin{center}
\resizebox{\textwidth}{!}{%
\begin{tabular}{l*{2}{S}}
\toprule
                                     {Policy} & {AUC Mean} &     {AUC Std.} \\
\midrule
                   Det Prop &  975.65445 &  {\textemdash} \\
                          Rand Prop &  975.42124 &      0.0581282 \\
            Det Prop Update &  975.36765 &  {\textemdash} \\
                   Rand Prop Update &  975.29324 &       0.127473 \\
        Det Prop Down &  973.81660 &  {\textemdash} \\
               Rand Prop Down &  973.71172 &      0.0849028 \\
 Det Prop Update Down &  973.70660 &  {\textemdash} \\
        Rand Prop Update Down &  973.50381 &        0.16896 \\
                                     Baseline &  973.38605 &       0.461298 \\
                  Rand Inverse Prop &  969.17370 &        0.27215 \\
           Det Inverse Prop &  969.04100 &  {\textemdash} \\
\bottomrule
\end{tabular}
}
\caption{AUC statistics:  CIFAR10 Rotate.}%
\label{table:auc_cifar10_rotate_loss}
\end{center}
\end{minipage}
\begin{minipage}{\tablesize\textwidth}
\captionsetup{font=scriptsize}
\begin{center}
\resizebox{\textwidth}{!}{%
\begin{tabular}{l*{2}{S}}
\toprule
                                     {Policy} &  {AUC Mean} &     {AUC Std.} \\
\midrule
            Det Prop Update &  964.547737 &  {\textemdash} \\
                   Rand Prop Update &  963.728331 &       0.799993 \\
                   Det Prop &  963.621708 &  {\textemdash} \\
                          Rand Prop &  963.385632 &       0.548808 \\
                                     Baseline &  960.356881 &        2.67395 \\
        Rand Prop Update Down &  953.545434 &       0.321935 \\
 Det Prop Update Down &  953.328436 &  {\textemdash} \\
               Rand Prop Down &  953.083712 &       0.257922 \\
        Det Prop Down &  952.623313 &  {\textemdash} \\
                  Rand Inverse Prop &  947.480150 &        1.34337 \\
           Det Inverse Prop &  944.813796 &  {\textemdash} \\
\bottomrule
\end{tabular}
}
\caption{AUC statistics:  NORB Rotate.}%
\label{table:auc_norb_rotate_loss}
\end{center}
\end{minipage}
}
\end{table}

\begin{table}[ht]
\makebox[\linewidth]{%
\hspacesize%
\begin{minipage}{\tablesize\textwidth}
\captionsetup{font=scriptsize}
\begin{center}
\resizebox{\textwidth}{!}{%
\begin{tabular}{l*{2}{S}}
\toprule
                                     {Policy} &  {AUC Mean} &     {AUC Std.} \\
\midrule
        Det Prop Down &  995.839250 &  {\textemdash} \\
               Rand Prop Down &  995.806372 &        0.47547 \\
                          Rand Prop &  995.767893 &       0.306641 \\
        Rand Prop Update Down &  995.649690 &        0.30183 \\
                   Rand Prop Update &  995.533021 &       0.460844 \\
                   Det Prop &  995.472278 &  {\textemdash} \\
 Det Prop Update Down &  995.260333 &  {\textemdash} \\
            Det Prop Update &  994.900634 &  {\textemdash} \\
                                     Baseline &  992.704111 &       0.565716 \\
           Det Inverse Prop &  984.145917 &  {\textemdash} \\
                  Rand Inverse Prop &  983.968887 &       0.383831 \\
\bottomrule
\end{tabular}
}
\caption{AUC statistics:  MNIST Crop.}%
\label{table:auc_mnist_crop_loss}
\end{center}
\end{minipage}
\begin{minipage}{\tablesize\textwidth}
\captionsetup{font=scriptsize}
\begin{center}
\resizebox{\textwidth}{!}{%
\begin{tabular}{l*{2}{S}}
\toprule
                                     {Policy} &  {AUC Mean} &     {AUC Std.} \\
\midrule
                   Det Prop &  966.573375 &  {\textemdash} \\
                          Rand Prop &  966.221575 &       0.478134 \\
            Det Prop Update &  966.083375 &  {\textemdash} \\
                   Rand Prop Update &  965.468450 &       0.509909 \\
        Det Prop Down &  964.256500 &  {\textemdash} \\
               Rand Prop Down &  963.145650 &       0.275241 \\
 Det Prop Update Down &  963.056500 &  {\textemdash} \\
        Rand Prop Update Down &  962.776900 &       0.381258 \\
                                     Baseline &  961.452600 &         1.0522 \\
                  Rand Inverse Prop &  958.989500 &       0.338857 \\
           Det Inverse Prop &  958.131875 &  {\textemdash} \\
\bottomrule
\end{tabular}
}
\caption{AUC statistics:  CIFAR10 Crop.}%
\label{table:auc_cifar10_crop_loss}
\end{center}
\end{minipage}
\begin{minipage}{\tablesize\textwidth}
\captionsetup{font=scriptsize}
\begin{center}
\resizebox{\textwidth}{!}{%
\begin{tabular}{l*{2}{S}}
\toprule
                                     {Policy} &  {AUC Mean} &     {AUC Std.} \\
\midrule
            Det Prop Update &  954.828832 &  {\textemdash} \\
                   Det Prop &  954.812114 &  {\textemdash} \\
                          Rand Prop &  954.420201 &       0.464086 \\
                   Rand Prop Update &  954.417243 &       0.241737 \\
                                     Baseline &  950.219537 &        2.27557 \\
        Rand Prop Update Down &  949.879362 &       0.388978 \\
 Det Prop Update Down &  949.714712 &  {\textemdash} \\
               Rand Prop Down &  949.647088 &       0.699074 \\
        Det Prop Down &  949.446219 &  {\textemdash} \\
                  Rand Inverse Prop &  936.743966 &        1.30082 \\
           Det Inverse Prop &  934.525206 &  {\textemdash} \\
\bottomrule
\end{tabular}
}
\caption{AUC statistics:  NORB Crop.}%
\label{table:auc_norb_crop_loss}
\end{center}
\end{minipage}
}
\end{table}

\newpage
\subsection{Area Under Curve (AUC) Results using Influence}
\begin{table}[ht]
\makebox[\linewidth]{%
\hspacesize%
\begin{minipage}{\tablesize\textwidth}
\captionsetup{font=scriptsize}
\begin{center}
\resizebox{\textwidth}{!}{%
\begin{tabular}{l*{2}{S}}
\toprule
                                     {Policy} &  {AUC Mean} &     {AUC Std.} \\
\midrule
                   Rand Prop Update &  995.415675 &        0.40924 \\
                          Rand Prop &  995.357631 &        0.14778 \\
                   Det Prop &  995.307107 &  {\textemdash} \\
               Rand Prop Down &  995.229748 &       0.404106 \\
        Det Prop Down &  995.142087 &  {\textemdash} \\
        Rand Prop Update Down &  995.136210 &        0.37608 \\
            Det Prop Update &  995.013810 &  {\textemdash} \\
 Det Prop Update Down &  994.455444 &  {\textemdash} \\
                                     Baseline &  993.933518 &       0.590857 \\
           Det Inverse Prop &  985.554587 &  {\textemdash} \\
                  Rand Inverse Prop &  985.164052 &        1.48664 \\
\bottomrule
\end{tabular}
   }
\caption{AUC statistics:  MNIST Translate.}%
\label{table:auc_mnist_translate}%
\end{center}%
\end{minipage}
\begin{minipage}{\tablesize\textwidth}
\captionsetup{font=scriptsize}
\begin{center}
\resizebox{\textwidth}{!}{%
\begin{tabular}{l*{2}{S}}
\toprule
                                     {Policy} & {AUC Mean} &     {AUC Std.} \\
\midrule
                   Det Prop &  972.63530 &  {\textemdash} \\
                          Rand Prop &  972.42670 &       0.219211 \\
                   Rand Prop Update &  971.95439 &      0.0676162 \\
            Det Prop Update &  971.92390 &  {\textemdash} \\
                                     Baseline &  970.74096 &       0.583576 \\
               Rand Prop Down &  970.25514 &       0.165278 \\
        Det Prop Down &  970.12760 &  {\textemdash} \\
        Rand Prop Update Down &  969.99458 &       0.144871 \\
 Det Prop Update Down &  969.96660 &  {\textemdash} \\
           Det Inverse Prop &  968.69450 &  {\textemdash} \\
                  Rand Inverse Prop &  968.61675 &       0.467616 \\
\bottomrule
\end{tabular}
   }
\caption{AUC statistics:  CIFAR10 Translate.}%
\label{table:auc_cifar10_translate}
\end{center}
\end{minipage}
\begin{minipage}{\tablesize\textwidth}
\captionsetup{font=scriptsize}
\begin{center}
\resizebox{\textwidth}{!}{%
\begin{tabular}{l*{2}{S}}
\toprule
                                     {Policy} &  {AUC Mean} &     {AUC Std.} \\
\midrule
                   Det Prop &  896.517078 &  {\textemdash} \\
                   Rand Prop Update &  896.096000 &       0.254452 \\
            Det Prop Update &  896.095175 &  {\textemdash} \\
                          Rand Prop &  895.844146 &       0.661057 \\
        Det Prop Down &  890.193940 &  {\textemdash} \\
 Det Prop Update Down &  889.859990 &  {\textemdash} \\
               Rand Prop Down &  889.832379 &       0.189458 \\
        Rand Prop Update Down &  889.605667 &       0.139056 \\
                                     Baseline &  887.544912 &        2.93673 \\
           Det Inverse Prop &  882.326667 &  {\textemdash} \\
                  Rand Inverse Prop &  882.154498 &       0.514772 \\
\bottomrule
\end{tabular}
  }
\caption{AUC statistics:  NORB Translate.}%
\label{table:auc_norb_translate}
\end{center}
\end{minipage}
}
\end{table}

\begin{table}[ht]
\makebox[\linewidth]{%
\hspacesize%
\begin{minipage}{\tablesize\textwidth}
\captionsetup{font=scriptsize}
\begin{center}
\resizebox{\textwidth}{!}{%
\begin{tabular}{l*{2}{S}}
\toprule
                                     {Policy} &  {AUC Mean} &     {AUC Std.} \\
\midrule
        Rand Prop Update Down &  997.277006 &      0.0705017 \\
 Det Prop Update Down &  997.249261 &  {\textemdash} \\
                          Rand Prop &  997.199751 &       0.080611 \\
                   Rand Prop Update &  997.161838 &       0.166519 \\
            Det Prop Update &  997.127974 &  {\textemdash} \\
        Det Prop Down &  997.111038 &  {\textemdash} \\
               Rand Prop Down &  997.054338 &        0.19958 \\
                   Det Prop &  996.671354 &  {\textemdash} \\
                                     Baseline &  995.120491 &       0.329292 \\
                  Rand Inverse Prop &  990.700954 &       0.417743 \\
           Det Inverse Prop &  990.457947 &  {\textemdash} \\
\bottomrule
\end{tabular}
}
\caption{AUC statistics:  MNIST Rotate.}%
\label{table:auc_mnist_rotate}
\end{center}
\end{minipage}
\begin{minipage}{\tablesize\textwidth}
\captionsetup{font=scriptsize}
\begin{center}
\resizebox{\textwidth}{!}{%
\begin{tabular}{l*{2}{S}}
\toprule
                                     {Policy} & {AUC Mean} &     {AUC Std.} \\
\midrule
                   Det Prop &  975.62445 &  {\textemdash} \\
            Det Prop Update &  975.43015 &  {\textemdash} \\
                          Rand Prop &  975.40772 &       0.183442 \\
                   Rand Prop Update &  975.32086 &       0.161372 \\
        Det Prop Down &  973.90660 &  {\textemdash} \\
               Rand Prop Down &  973.79193 &      0.0806332 \\
 Det Prop Update Down &  973.72160 &  {\textemdash} \\
        Rand Prop Update Down &  973.67758 &       0.149203 \\
                                     Baseline &  973.01140 &        1.40159 \\
                  Rand Inverse Prop &  969.08550 &       0.426833 \\
           Det Inverse Prop &  968.83600 &  {\textemdash} \\
\bottomrule
\end{tabular}
}
\caption{AUC statistics:  CIFAR10 Rotate.}%
\label{table:auc_cifar10_rotate}
\end{center}
\end{minipage}
\begin{minipage}{\tablesize\textwidth}
\captionsetup{font=scriptsize}
\begin{center}
\resizebox{\textwidth}{!}{%
\begin{tabular}{l*{2}{S}}
\toprule
                                     {Policy} &  {AUC Mean} &     {AUC Std.} \\
\midrule
            Det Prop Update &  964.547737 &  {\textemdash} \\
                   Rand Prop Update &  964.150897 &       0.677828 \\
                   Det Prop &  963.631996 &  {\textemdash} \\
                          Rand Prop &  963.605189 &       0.358473 \\
                                     Baseline &  959.909469 &        2.16396 \\
        Rand Prop Update Down &  953.593107 &       0.552382 \\
               Rand Prop Down &  953.405461 &       0.563808 \\
 Det Prop Update Down &  953.361872 &  {\textemdash} \\
        Det Prop Down &  952.613025 &  {\textemdash} \\
                  Rand Inverse Prop &  947.514942 &        1.95937 \\
           Det Inverse Prop &  944.817891 &  {\textemdash} \\
\bottomrule
\end{tabular}
}
\caption{AUC statistics:  NORB Rotate.}%
\label{table:auc_norb_rotate}
\end{center}
\end{minipage}
}
\end{table}

\begin{table}[ht]
\makebox[\linewidth]{%
\hspacesize%
\begin{minipage}{\tablesize\textwidth}
\captionsetup{font=scriptsize}
\begin{center}
\resizebox{\textwidth}{!}{%
\begin{tabular}{l*{2}{S}}
\toprule
                                     {Policy} &  {AUC Mean} &     {AUC Std.} \\
\midrule
                          Rand Prop &  996.018224 &       0.424856 \\
                   Det Prop &  995.965654 &  {\textemdash} \\
        Det Prop Down &  995.958849 &  {\textemdash} \\
        Rand Prop Update Down &  995.883374 &       0.391658 \\
                   Rand Prop Update &  995.828089 &       0.533589 \\
 Det Prop Update Down &  995.824921 &  {\textemdash} \\
            Det Prop Update &  995.723826 &  {\textemdash} \\
               Rand Prop Down &  995.265618 &       0.574943 \\
                                     Baseline &  992.995903 &       0.692921 \\
                  Rand Inverse Prop &  984.038069 &         1.0967 \\
           Det Inverse Prop &  982.944736 &  {\textemdash} \\
\bottomrule
\end{tabular}
}
\caption{AUC statistics:  MNIST Crop.}%
\label{table:auc_mnist_crop}
\end{center}
\end{minipage}
\begin{minipage}{\tablesize\textwidth}
\captionsetup{font=scriptsize}
\begin{center}
\resizebox{\textwidth}{!}{%
\begin{tabular}{l*{2}{S}}
\toprule
                                     {Policy} &  {AUC Mean} &     {AUC Std.} \\
\midrule
                   Det Prop &  966.585875 &  {\textemdash} \\
                          Rand Prop &  966.528575 &       0.322027 \\
            Det Prop Update &  966.180875 &  {\textemdash} \\
                   Rand Prop Update &  965.683800 &        0.25058 \\
        Det Prop Down &  964.369000 &  {\textemdash} \\
               Rand Prop Down &  963.935825 &       0.321492 \\
 Det Prop Update Down &  963.305250 &  {\textemdash} \\
        Rand Prop Update Down &  962.907725 &       0.218197 \\
                                     Baseline &  962.605750 &        1.30439 \\
                  Rand Inverse Prop &  958.602375 &       0.751481 \\
           Det Inverse Prop &  957.854375 &  {\textemdash} \\
\bottomrule
\end{tabular}
}
\caption{AUC statistics:  CIFAR10 Crop.}%
\label{table:auc_cifar10_crop}
\end{center}
\end{minipage}
\begin{minipage}{\tablesize\textwidth}
\captionsetup{font=scriptsize}
\begin{center}
\resizebox{\textwidth}{!}{%
\begin{tabular}{l*{2}{S}}
\toprule
                                     {Policy} &  {AUC Mean} &     {AUC Std.} \\
\midrule
                   Det Prop &  954.837834 &  {\textemdash} \\
            Det Prop Update &  954.828832 &  {\textemdash} \\
                   Rand Prop Update &  954.682016 &       0.503224 \\
                          Rand Prop &  954.603976 &       0.390722 \\
                                     Baseline &  951.898925 &         1.5072 \\
               Rand Prop Down &  950.104249 &        0.35844 \\
 Det Prop Update Down &  949.714712 &  {\textemdash} \\
        Rand Prop Update Down &  949.557803 &       0.470102 \\
        Det Prop Down &  949.446219 &  {\textemdash} \\
                  Rand Inverse Prop &  936.786667 &        2.05584 \\
           Det Inverse Prop &  934.529321 &  {\textemdash} \\
\bottomrule
\end{tabular}
}
\caption{AUC statistics:  NORB Crop.}%
\label{table:auc_norb_crop}
\end{center}
\end{minipage}
}
\end{table}

\newpage
\section{Diagnosing Influence and Loss}%
\label{sec:diagnosing}
\begin{figure}[ht]
\begin{center}
  \includegraphics[width=0.9\linewidth]{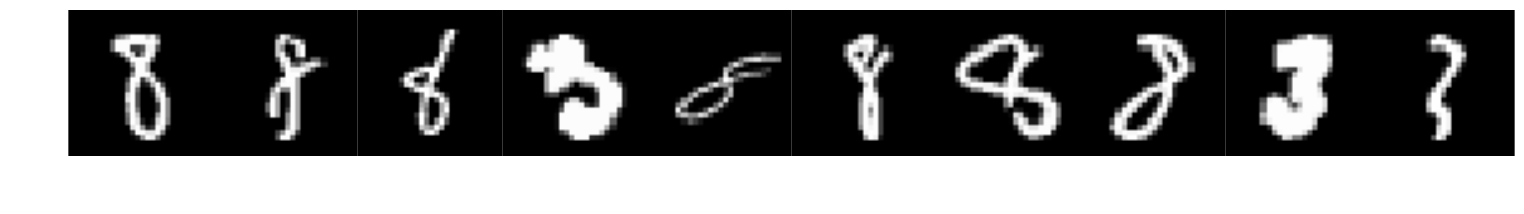}%
  \vspace{-6mm}
  \includegraphics[width=0.9\linewidth]{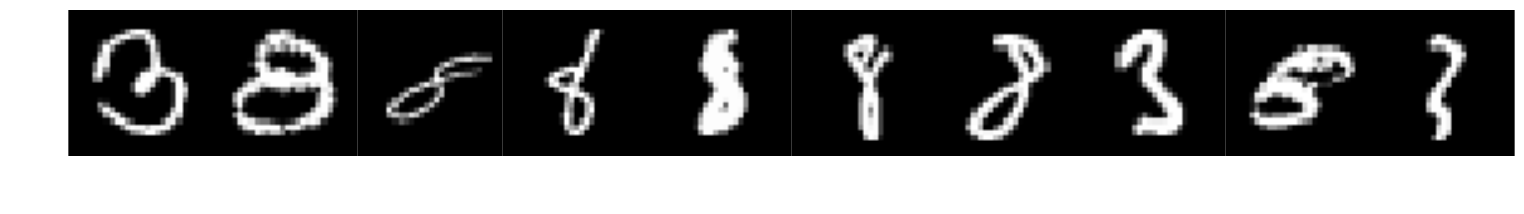}%
  \vspace{-6mm}
  \includegraphics[width=0.9\linewidth]{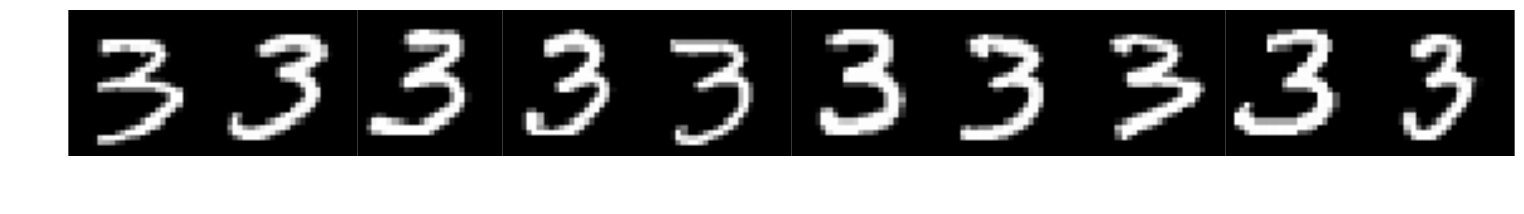}%
  \vspace{-6mm}
  \includegraphics[width=0.9\linewidth]{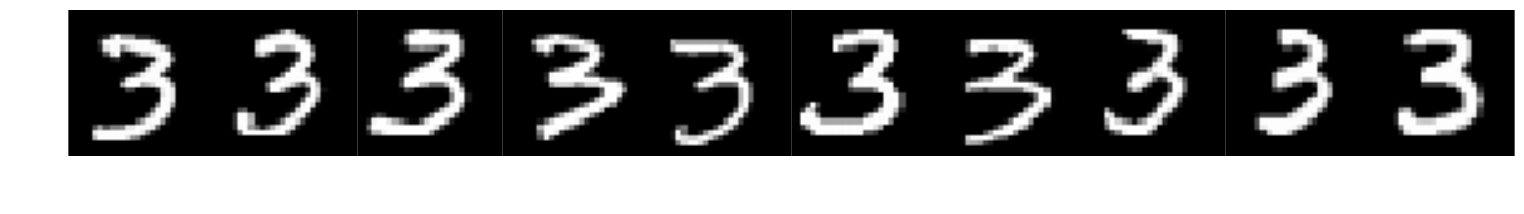}
\end{center}
\caption{From top to bottom: high influence, high loss, low influence, and
  low loss for MNIST.}
\end{figure}
\begin{figure}[ht]
\begin{center}
  \includegraphics[width=0.9\linewidth]{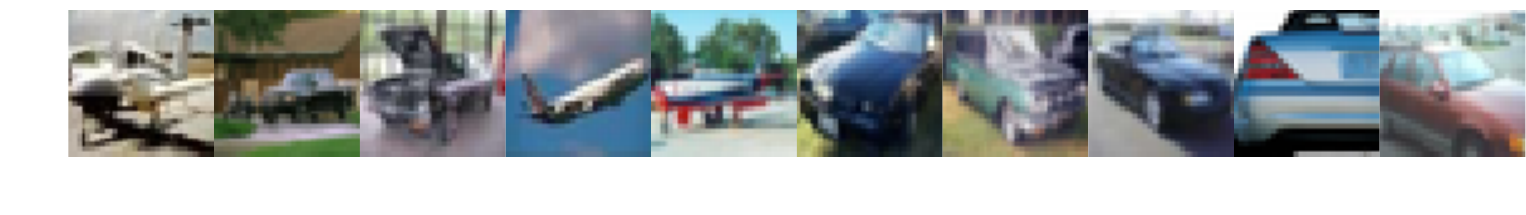}%
  \vspace{-6mm}
  \includegraphics[width=0.9\linewidth]{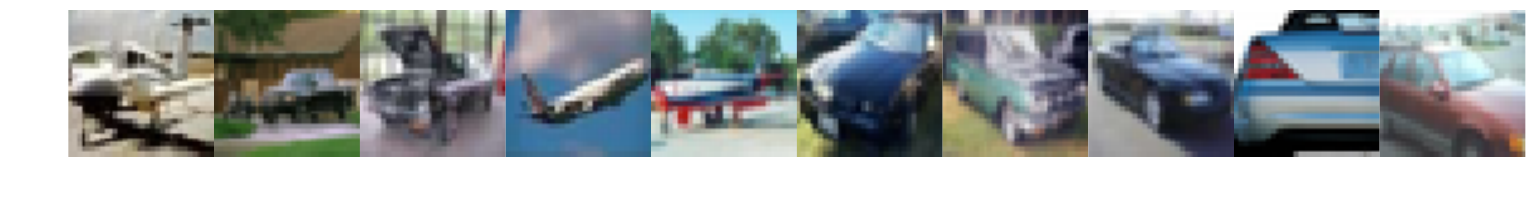}%
  \vspace{-6mm}
  \includegraphics[width=0.9\linewidth]{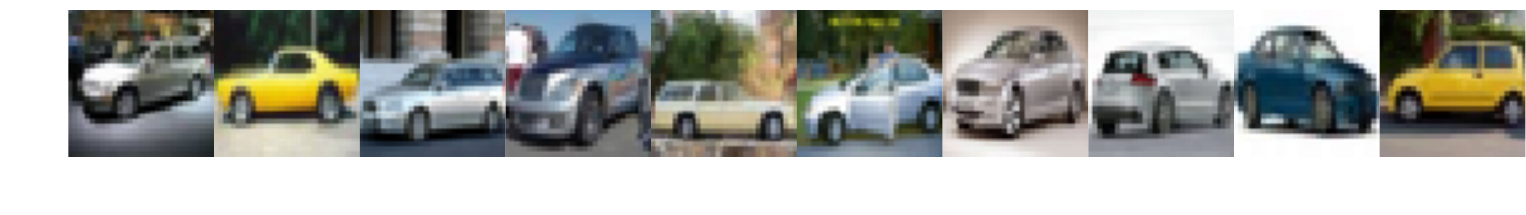}%
  \vspace{-6mm}
  \includegraphics[width=0.9\linewidth]{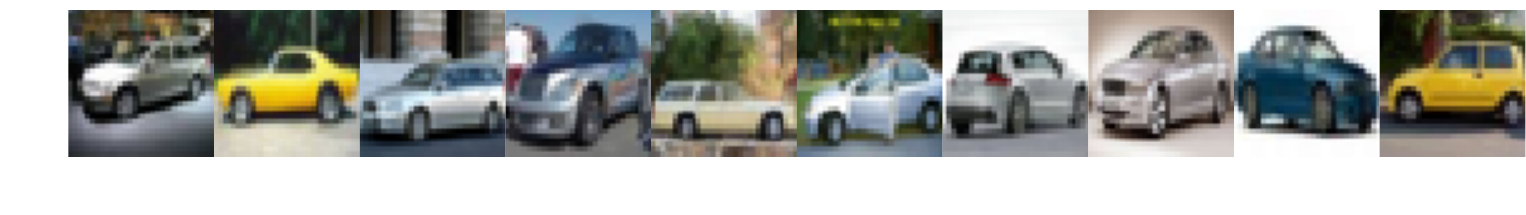}
\end{center}
\caption{From top to bottom: high influence, high loss, low influence, and
  low loss for CIFAR10.}
\end{figure}
\begin{figure}[ht]
\begin{center}
  \includegraphics[width=0.9\linewidth]{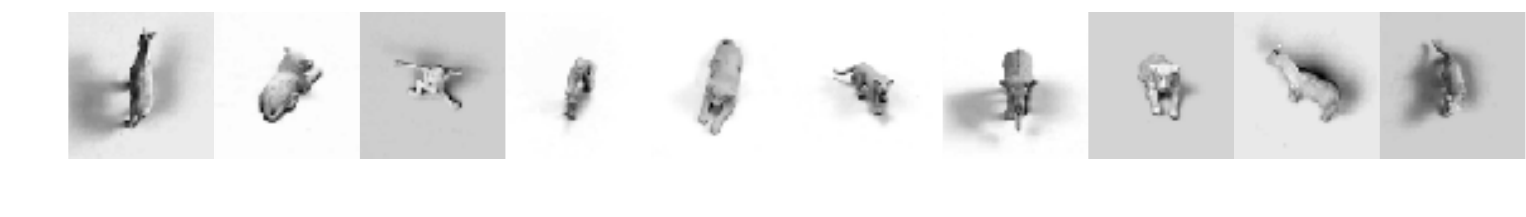}%
  \vspace{-6mm}
  \includegraphics[width=0.9\linewidth]{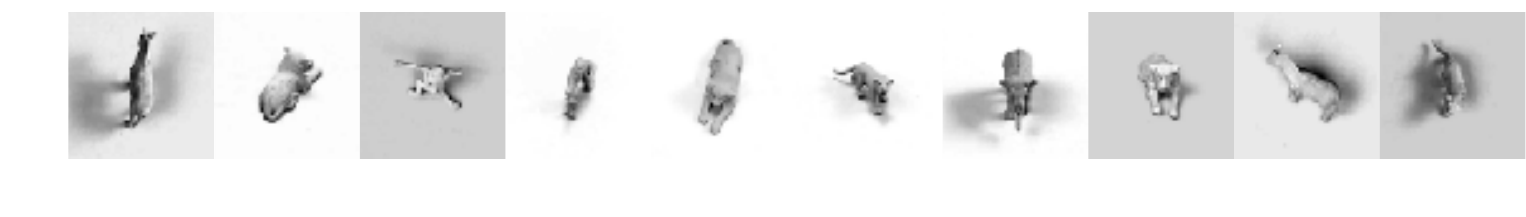}%
  \vspace{-6mm}
  \includegraphics[width=0.9\linewidth]{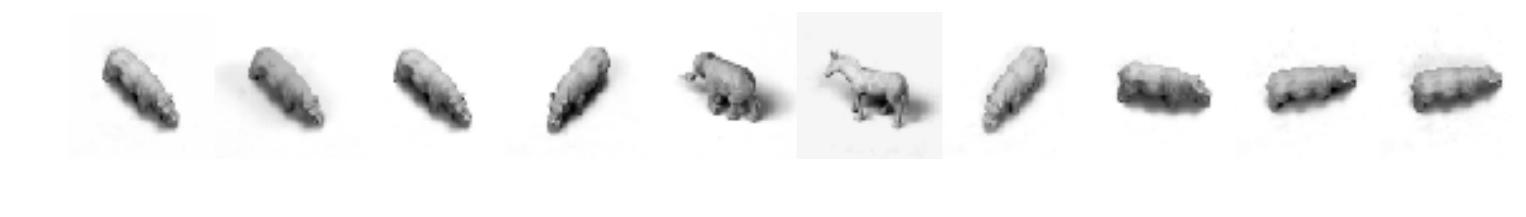}%
  \vspace{-6mm}
  \includegraphics[width=0.9\linewidth]{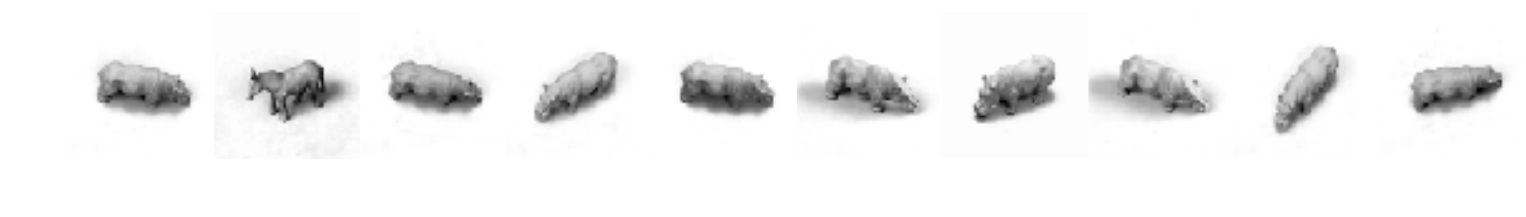}
\end{center}
\caption{From top to bottom: high influence, high loss, low influence, and
  low loss for NORB.}
\end{figure}

\newpage

\section{SVM Margin as a Score}%
\label{sec:margin}
Having used the VSV method, we can expand on this idea by using an SVM's margin
to score points.
The idea is that points farther away from the margin may also be less important
to the logistic regression model.
However, it is worth noting that the SVM and logistic regression decision surfaces may vary
significantly, and the solution of each model is dependent on a variety of
hyperparameters.
We include results utilizing the absolute value of the margin as well as the
inverse of that value.
We find that the model mismatch (i.e., SVM vs.\ logistic regression) combined
with a different score (i.e., margin vs.\ loss or influence) results in uniformly worse
performance than our proposed influence-based approach. 
However, it is worth noting that there is some transfer between models, which
poses potentially interesting questions for future investigation.

\begin{figure}[h!]
\begin{center}
  \subfloat[MNIST-translate]{%
  \trimfigleft{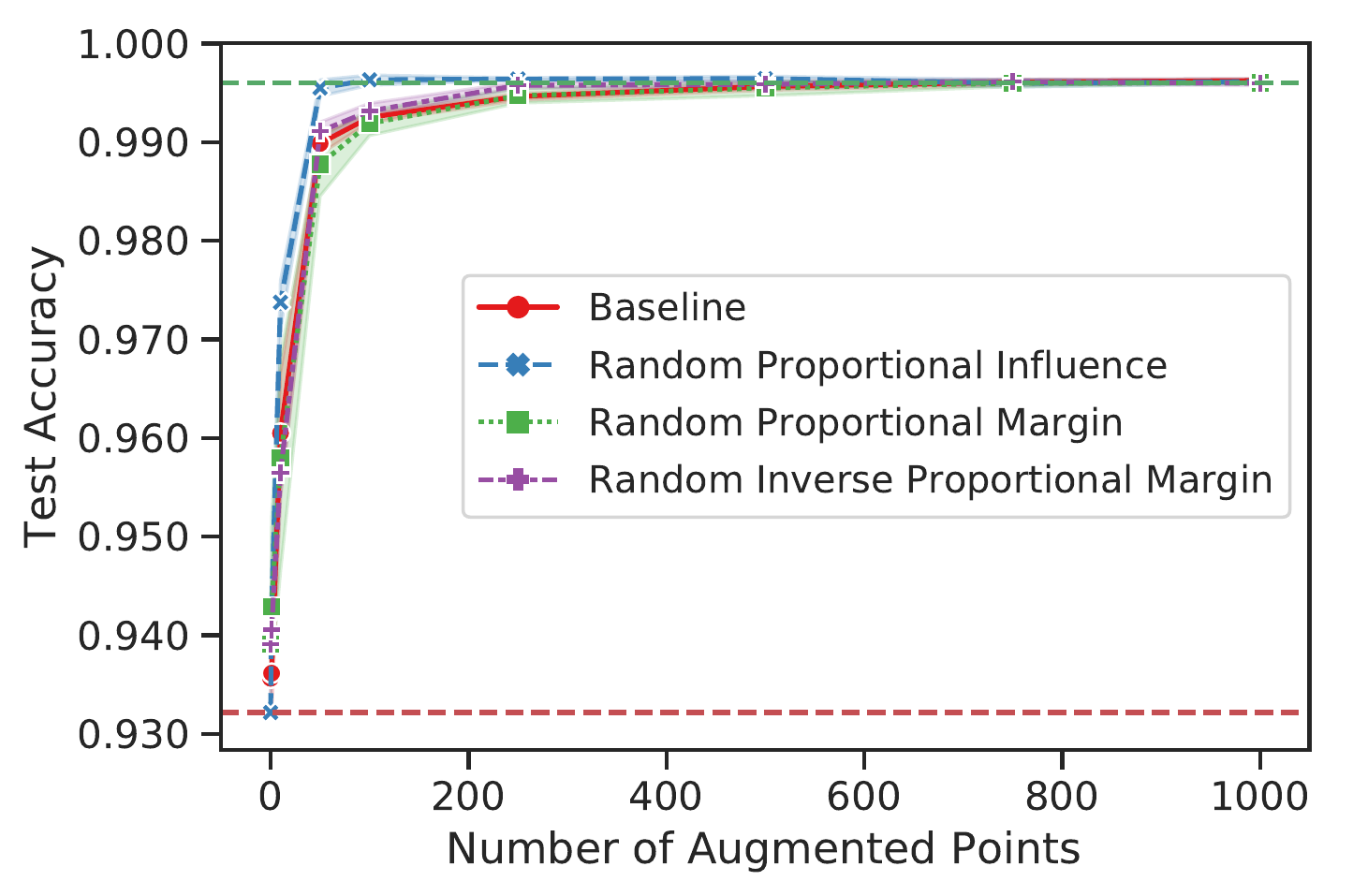}
  }%
   \subfloat[CIFAR10-translate]{%
  \trimfig{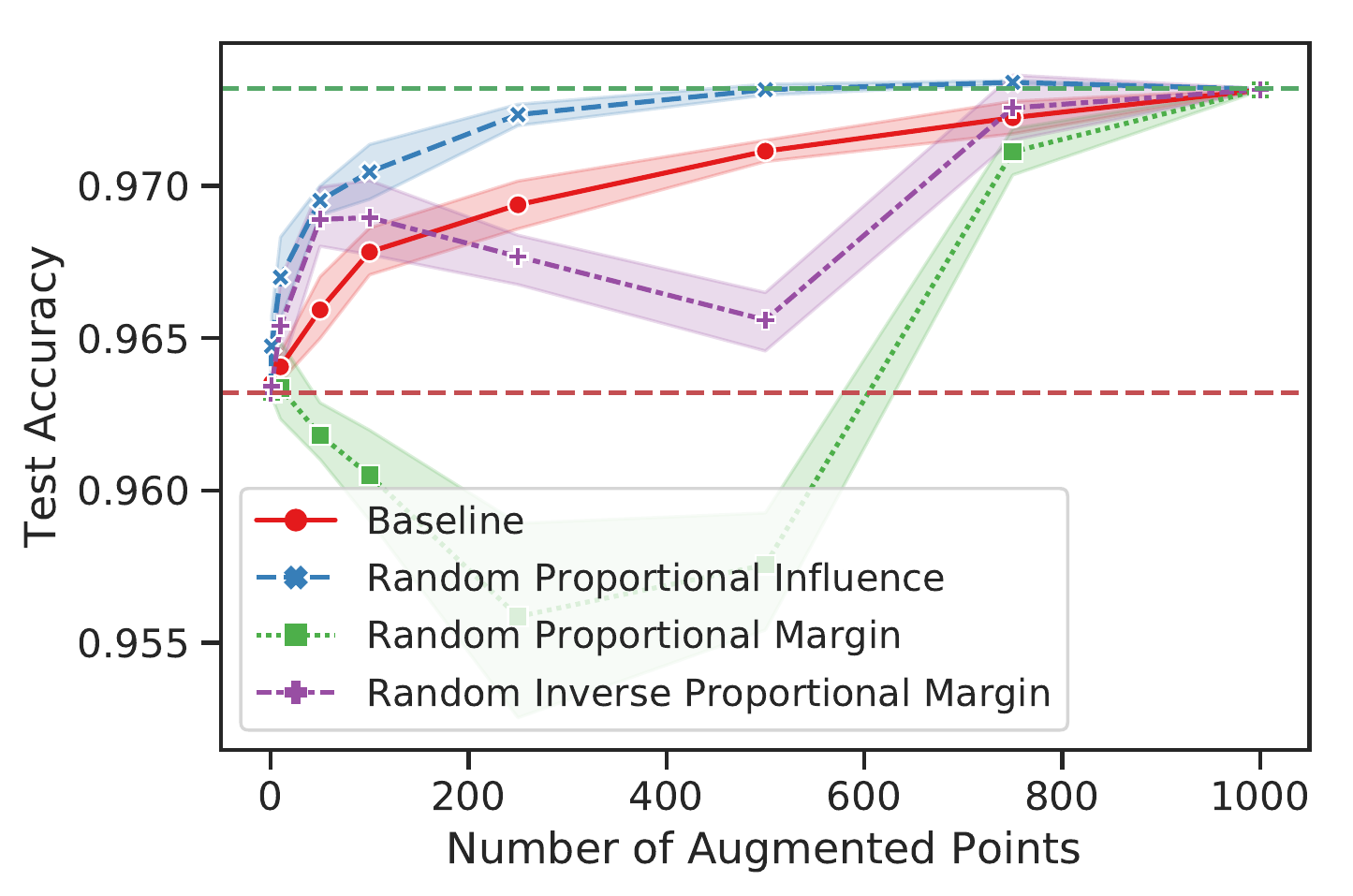}
  }%
    \subfloat[NORB-translate]{%
  \trimfig{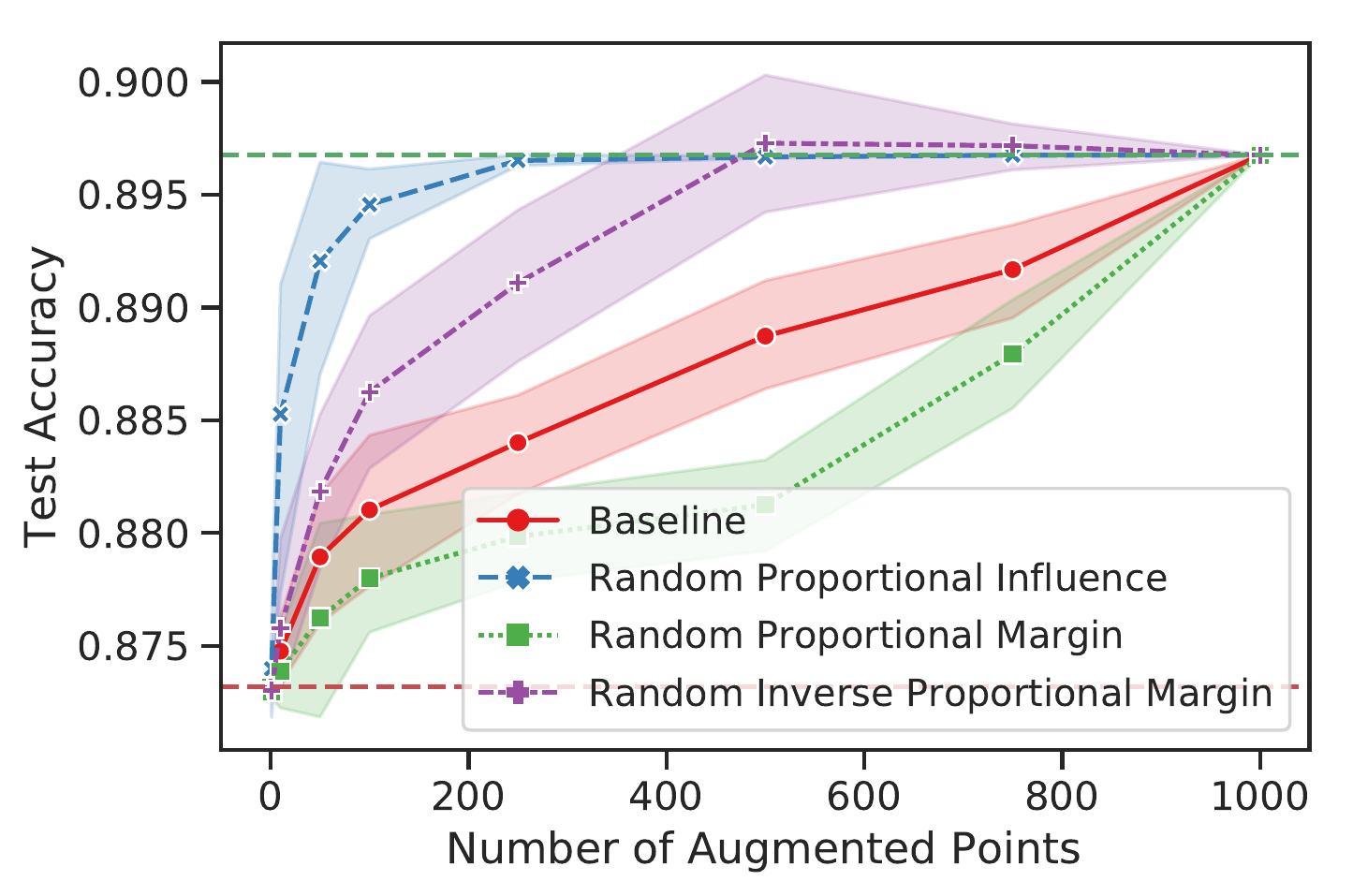}
  }%
  \newline
  \subfloat[MNIST-rotate]{%
  \trimfigleft{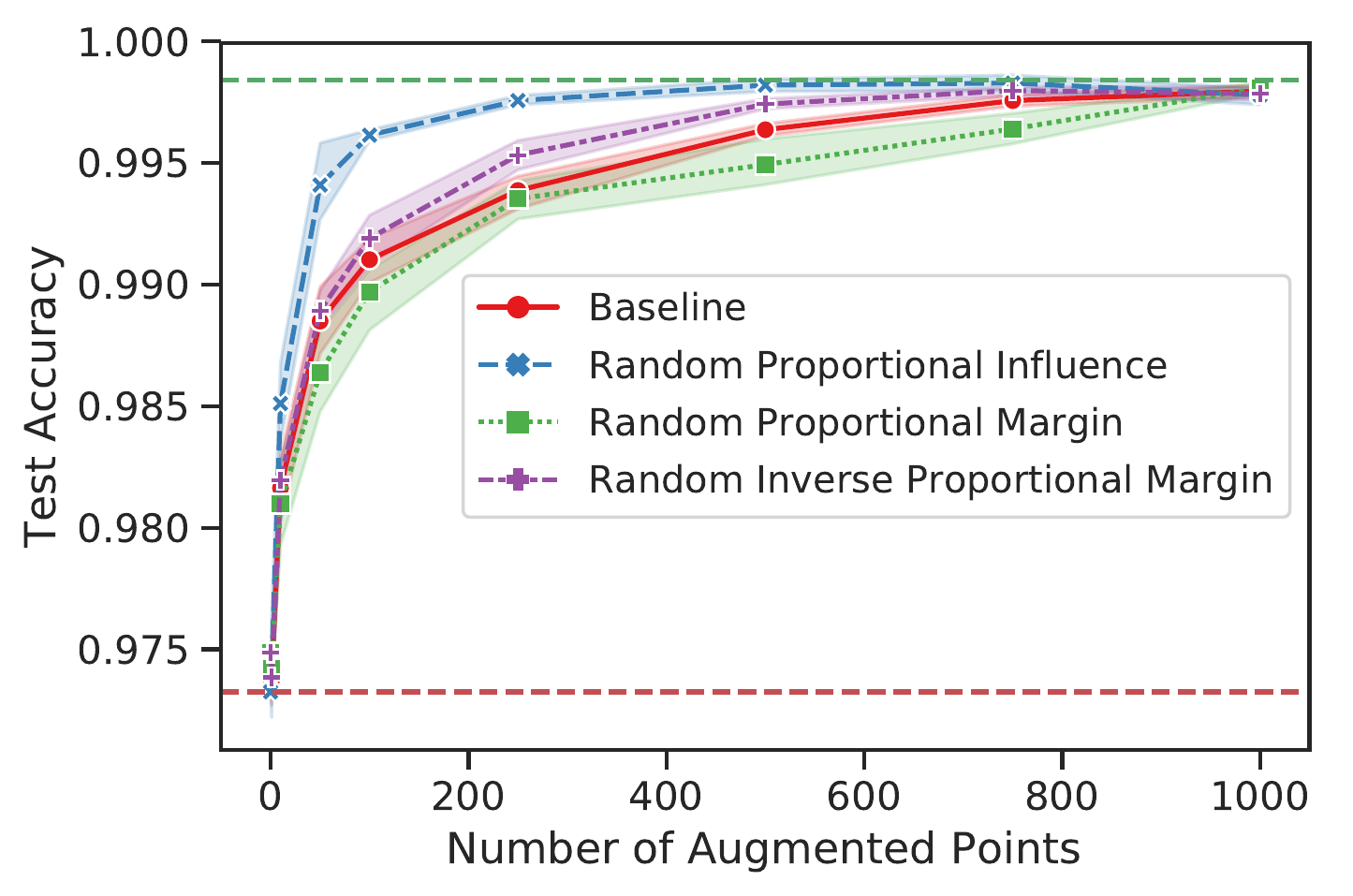}
  }%
    \subfloat[CIFAR10-rotate]{%
  \trimfig{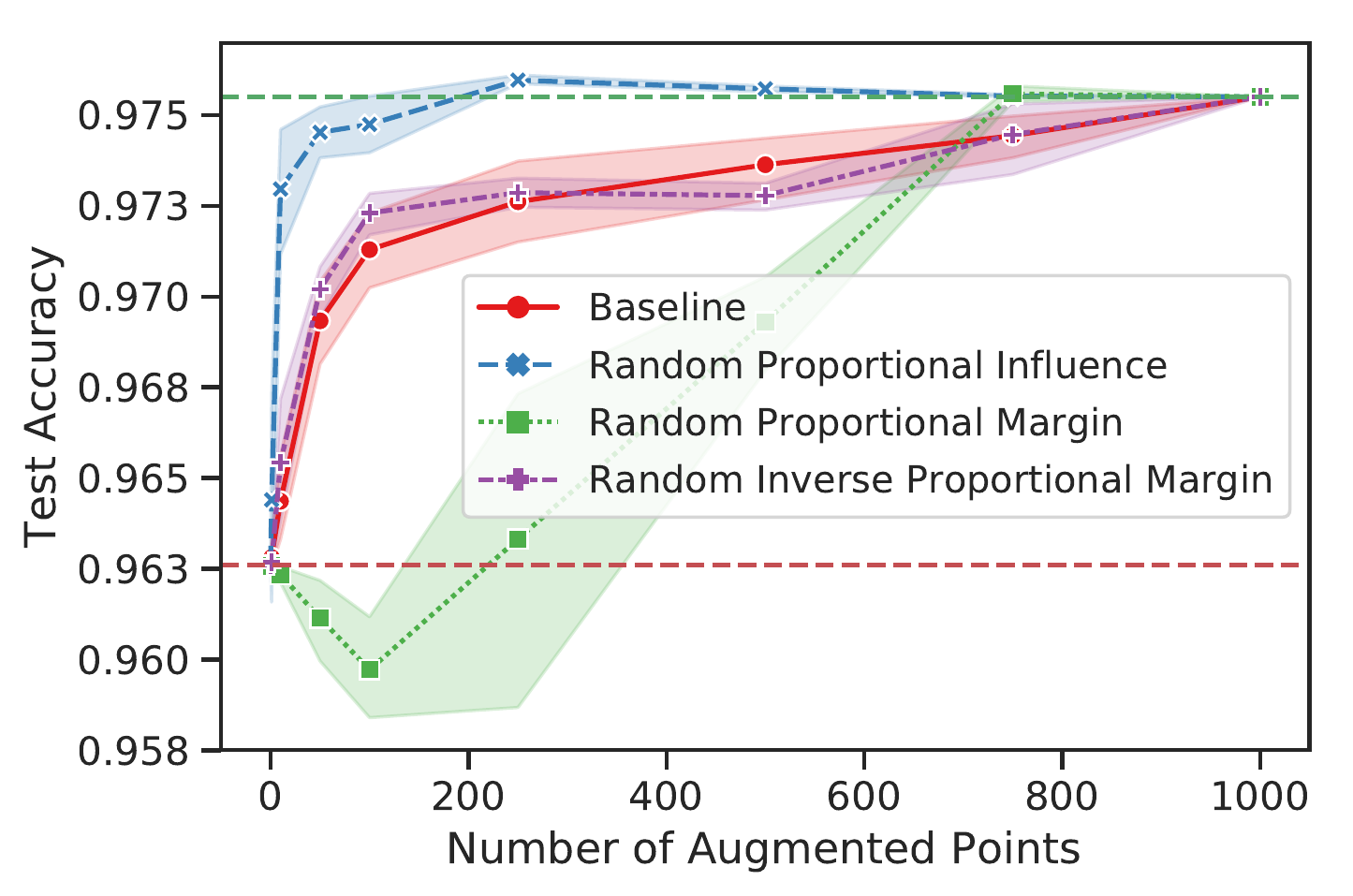}
  }%
    \subfloat[NORB-rotate]{%
  \trimfig{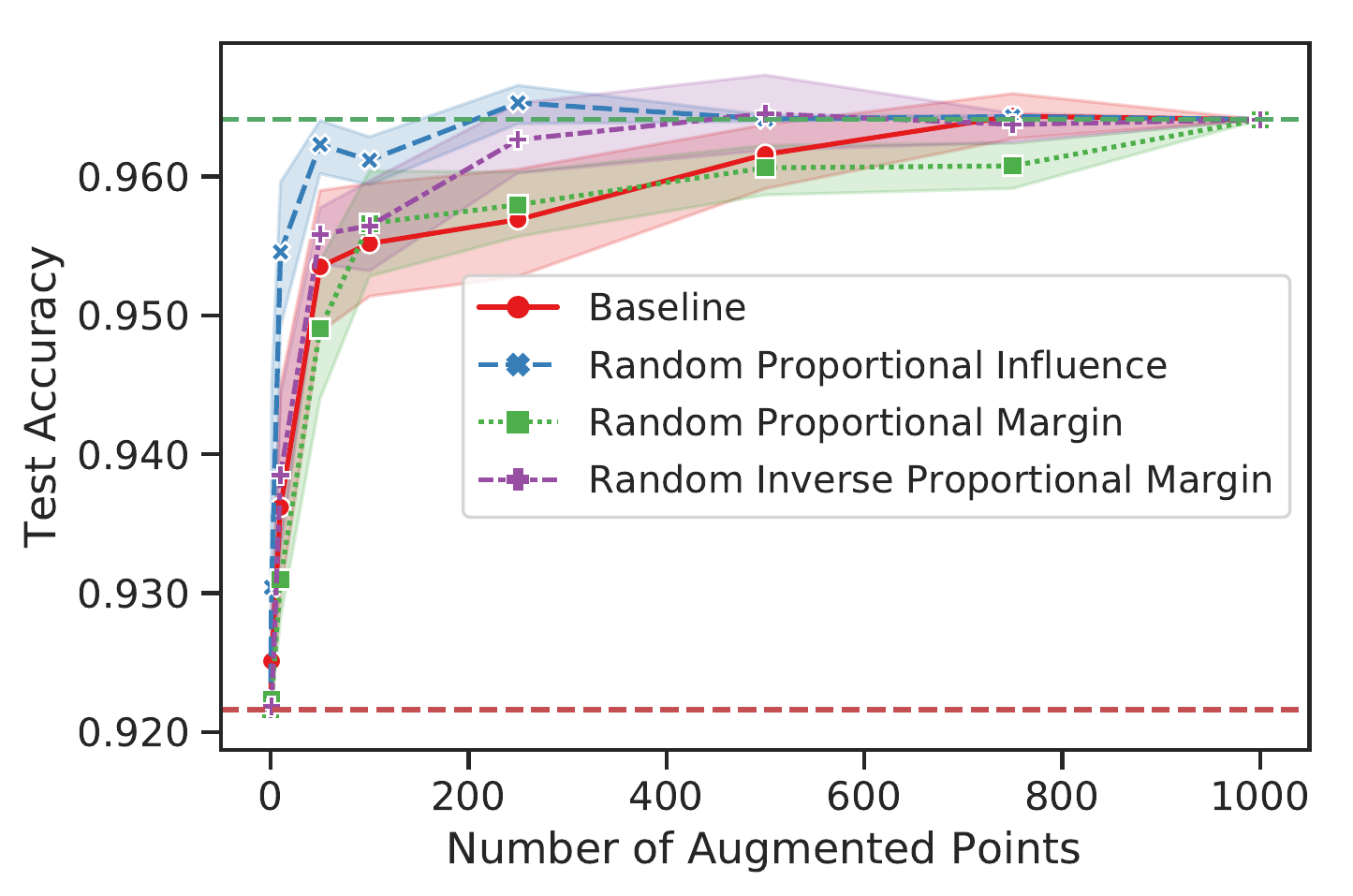}
  }%
  \newline
  \subfloat[MNIST-crop]{%
    \trimfigleft{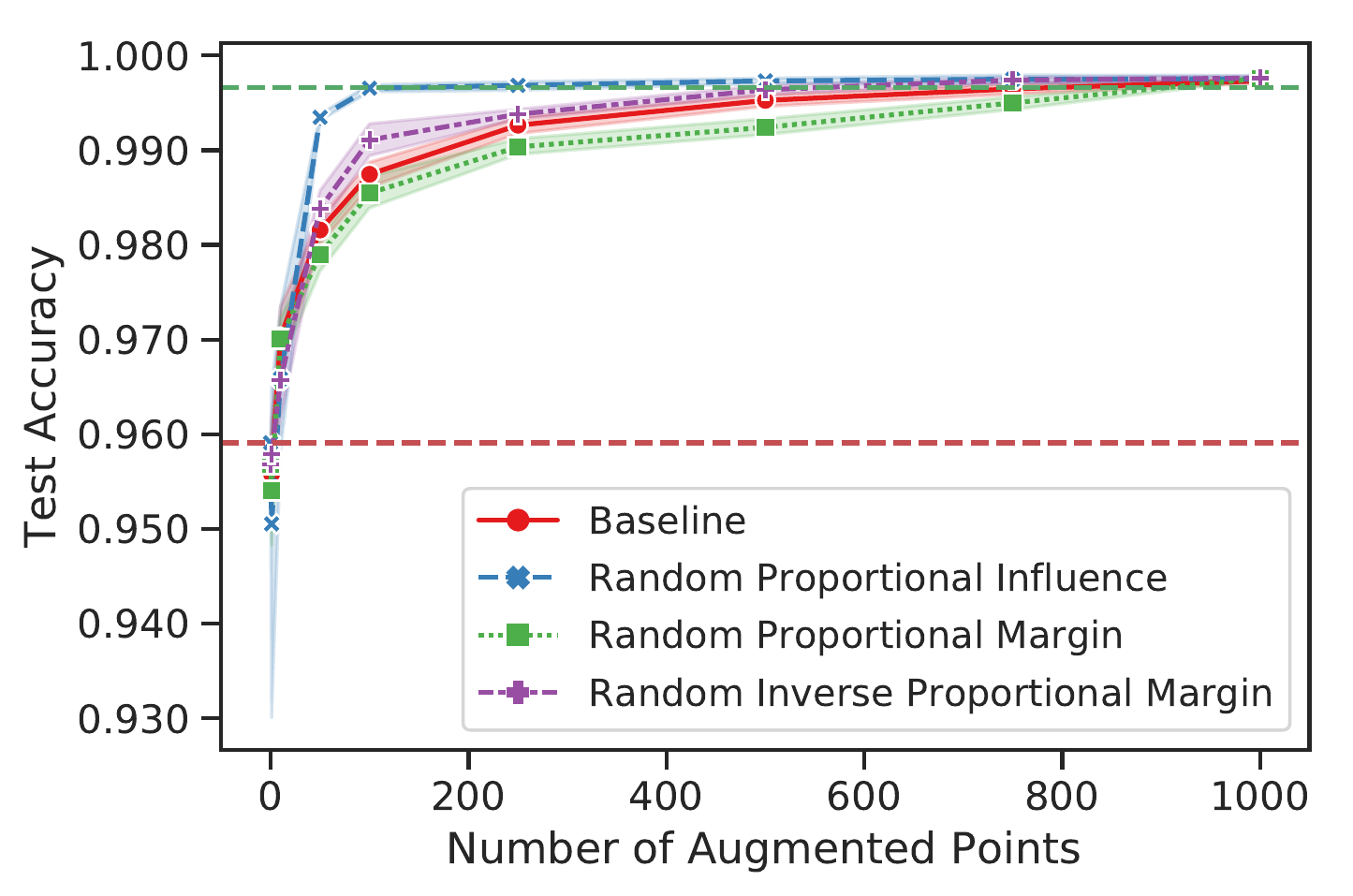}
  }%
  \subfloat[CIFAR10-crop]{%
    \trimfig{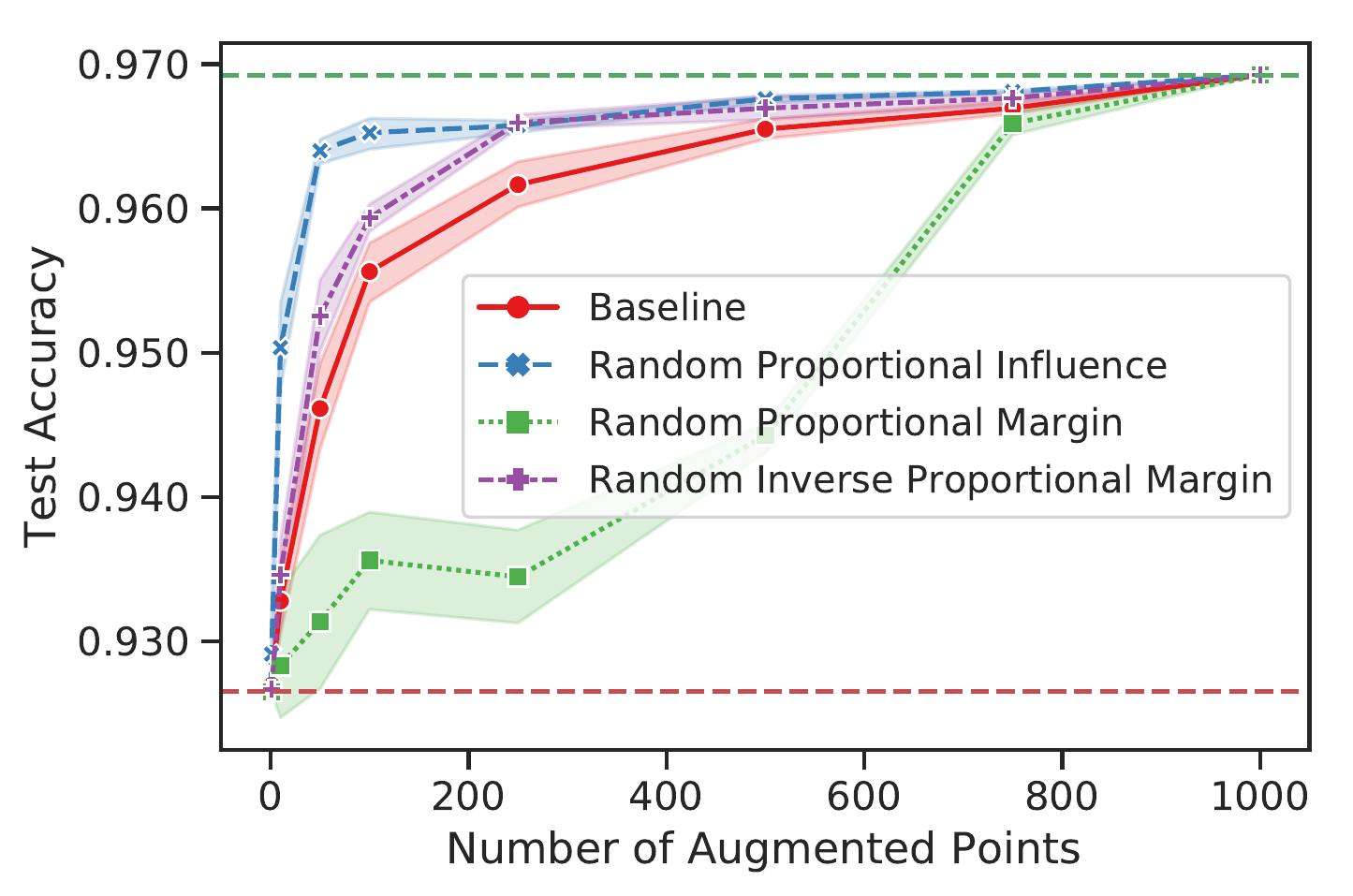}
  }%
  \subfloat[NORB-crop]{%
  \trimfig{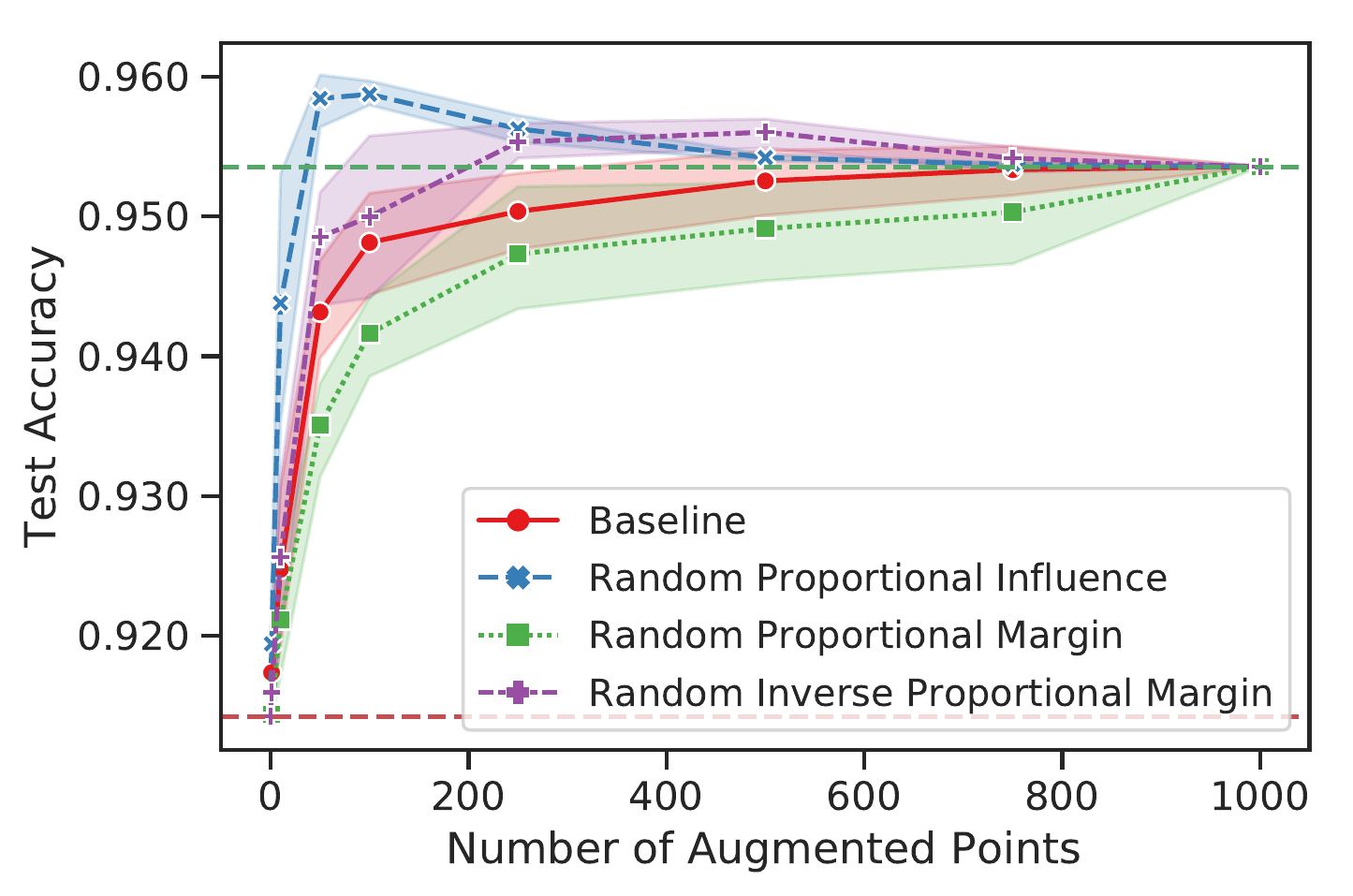}
  }%
  \caption{The performance of randomized policies using SVM margin.}%
\label{fig:all_mod_SVM_margin}%
\end{center}
\end{figure}

\section{Stratified Sampling with Clustering}%
\label{sec:sample_clustering}
Here we present results of performing augmentation with cluster-based stratified sampling.
One motivation of this method is to see how enforced diversity helps the
augmentation process, since points with high influence or loss seem diverse.
The subgroups are derived by using k-means clustering, where we fix the number of clusters to be equal to the
sample size.
We explore two policies: one where a point is selected at random from each cluster (Baseline Clustered),
and another where the point is selected with probability proportional to its influence value (Random Proportional Influence Clustered).

The baseline using clustering outperforms the standard (uniform random sampling) baseline in
many tasks.
We speculate that this is due to the forced increase in diversity.
However, the standard influence policy outperforms the clustered baseline policy for all datasets and augmentations, and also outperforms the combined strategy (Random Proportional Influence Clustered) in most tasks.
This suggests that stratified sampling can improve over simple random sampling.
However, other polices,
such as weighting points by influence/loss, seem to provide greater benefits for
augmentation, especially when point quality is aggressively traded off with
point diversity.
Indeed, as we increase the number of clusters, the combined policies approach
random sampling.
Selecting the number of clusters also poses practical issues for this sort of
approach.

\begin{figure}[h!]
\begin{center}
  \subfloat[MNIST-translate]{%
  \trimfigleft{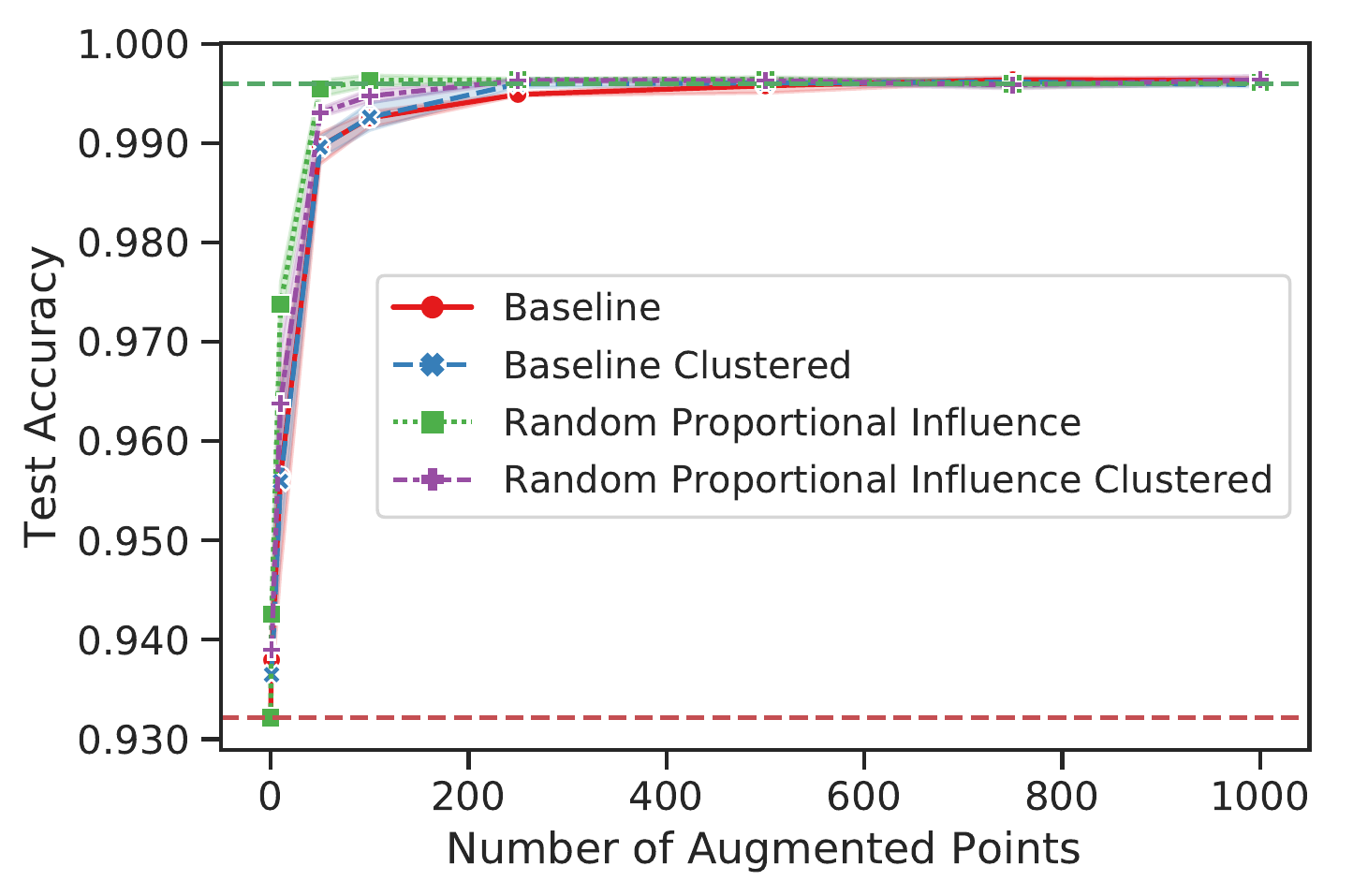}
  }%
   \subfloat[CIFAR10-translate]{%
  \trimfig{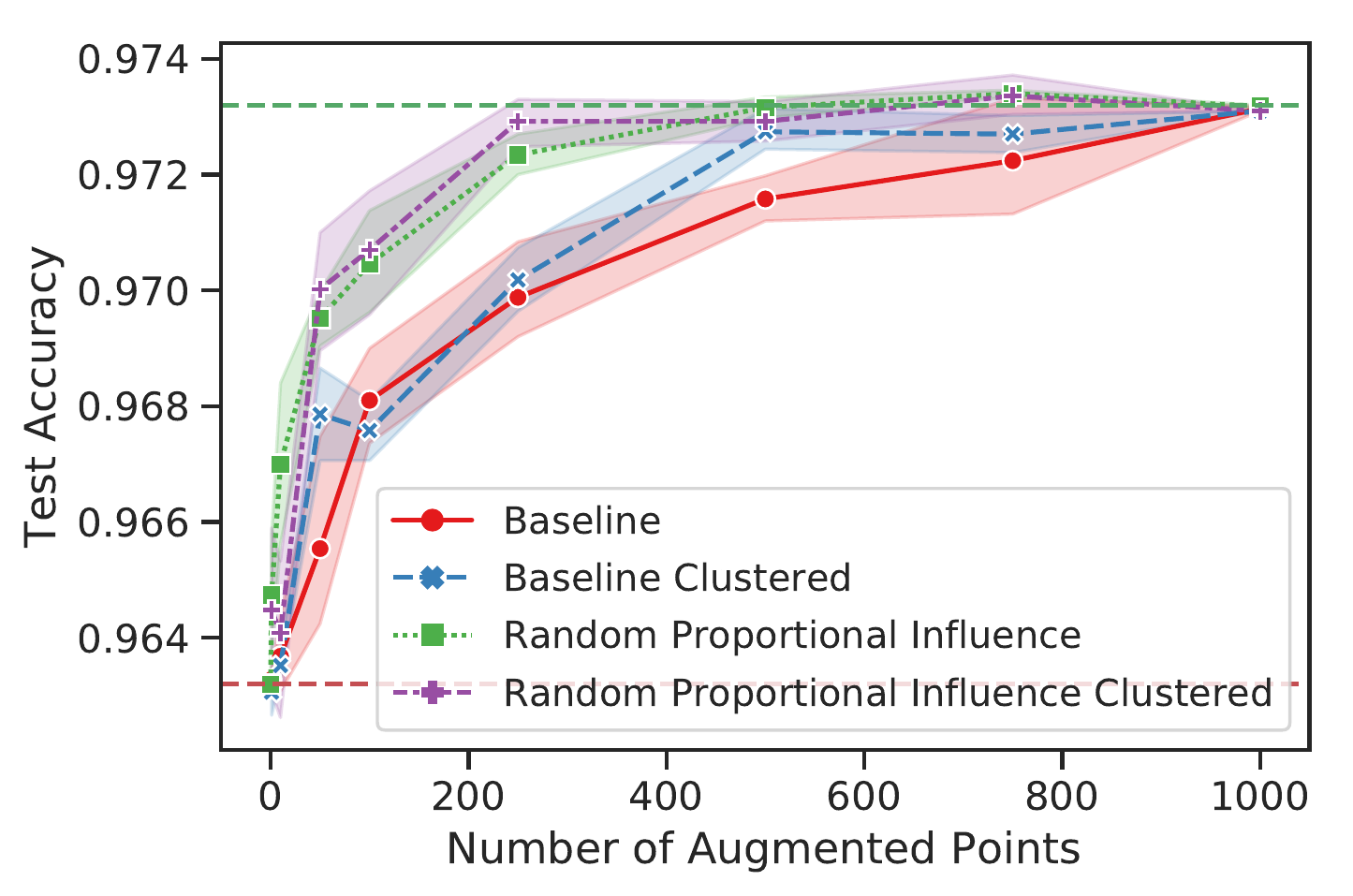}
  }%
    \subfloat[NORB-translate]{%
  \trimfig{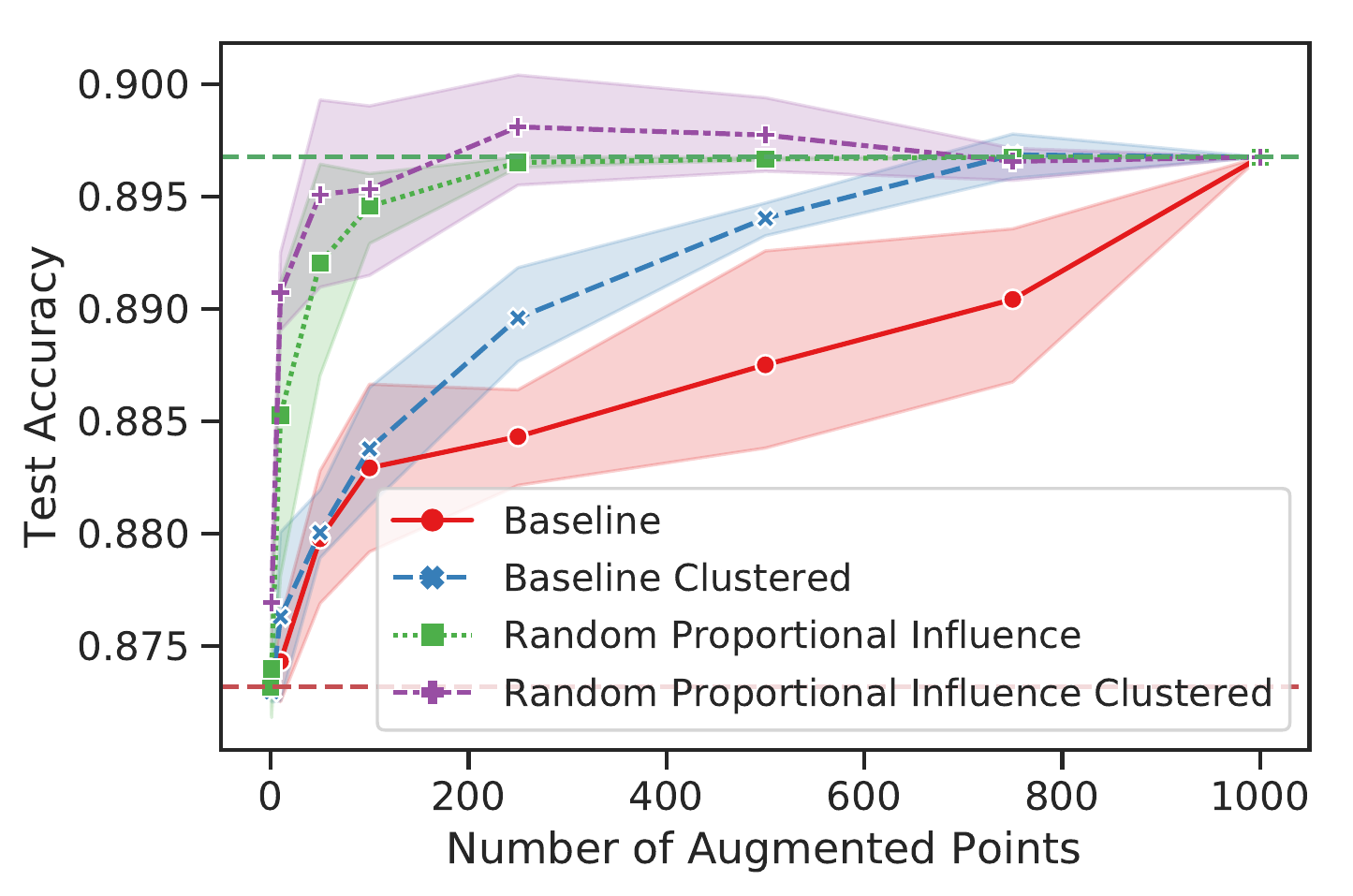}
  }%
  \newline
  \subfloat[MNIST-rotate]{%
  \trimfigleft{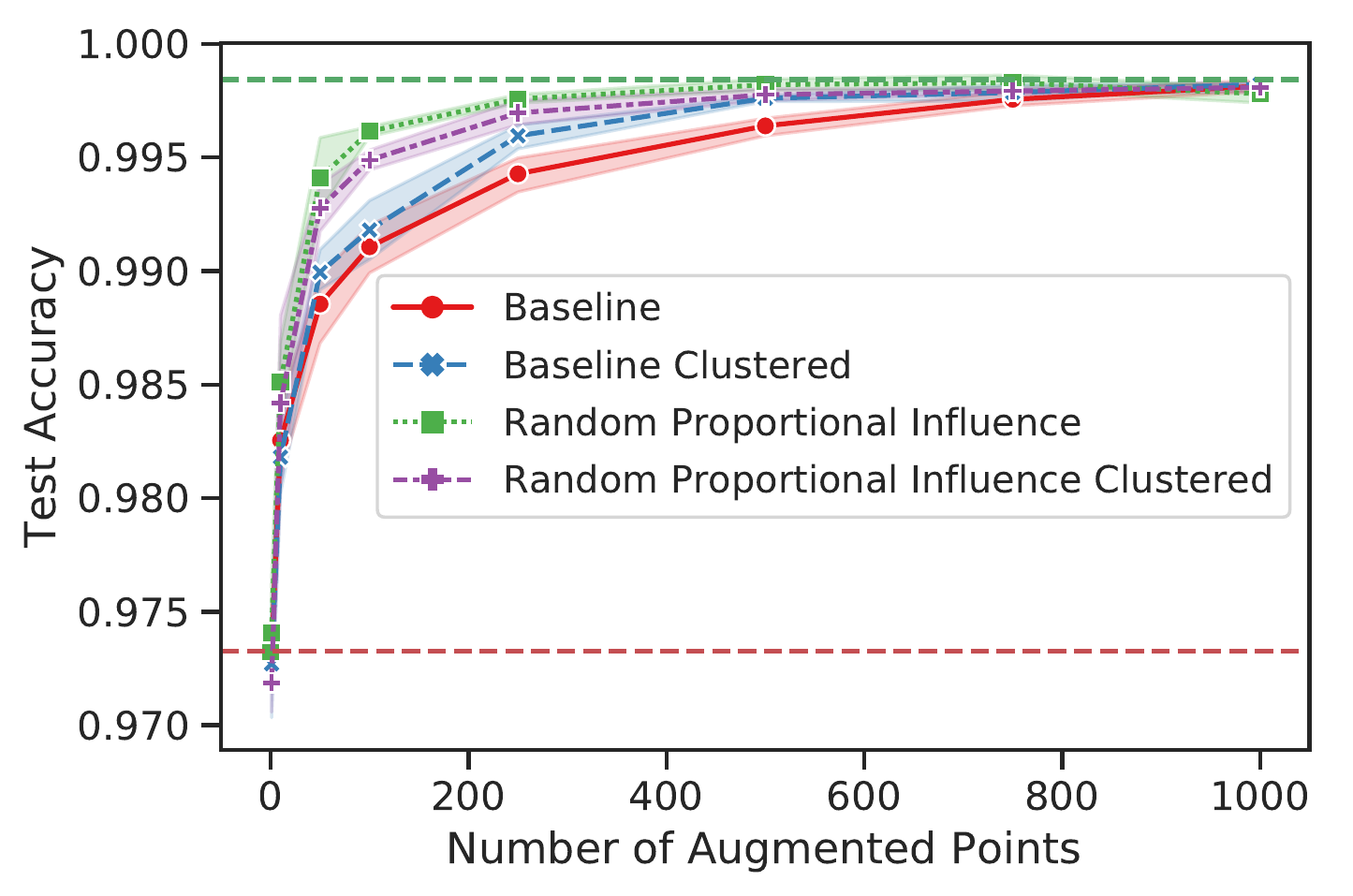}
  }%
    \subfloat[CIFAR10-rotate]{%
  \trimfig{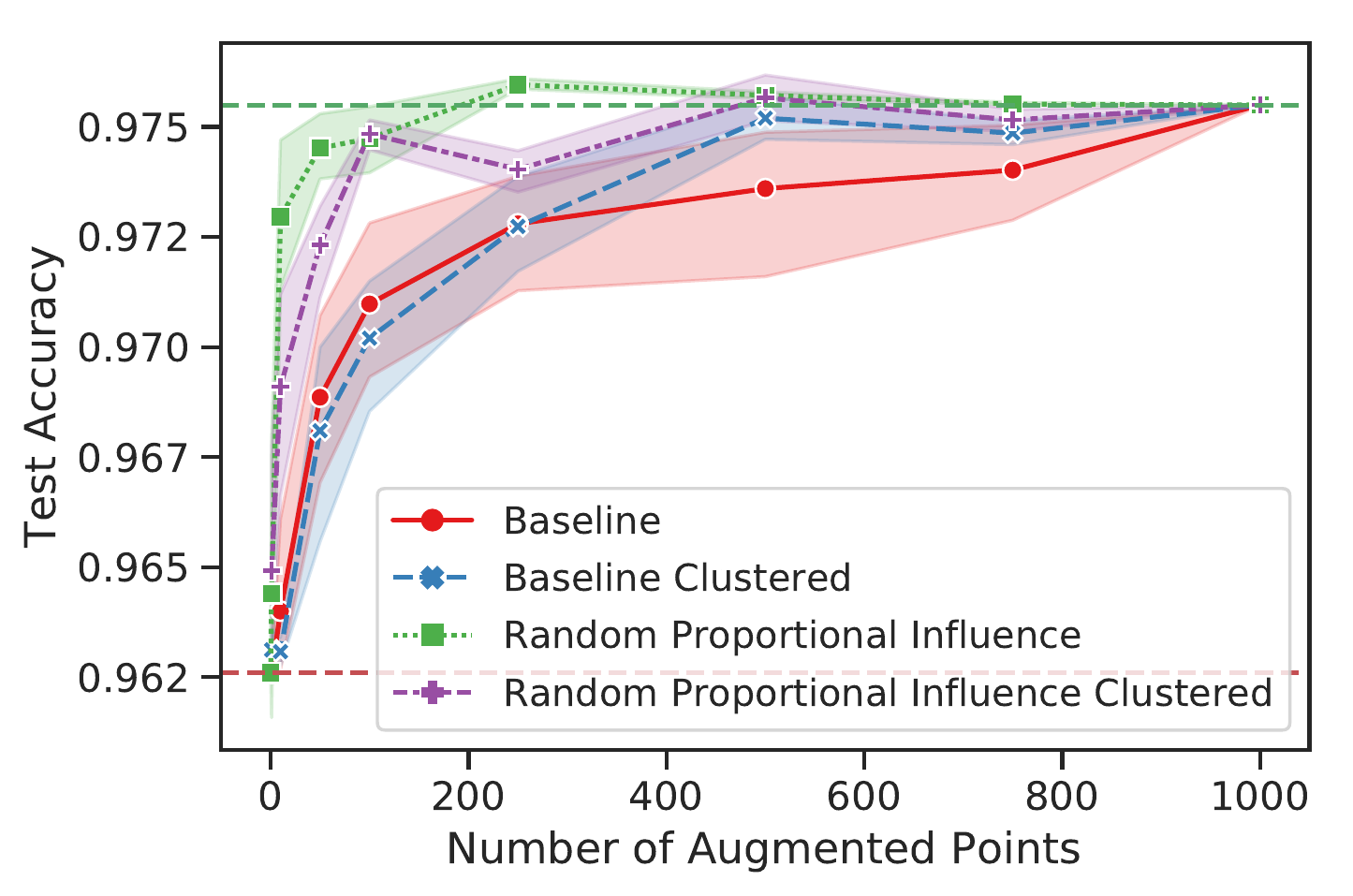}
  }%
    \subfloat[NORB-rotate]{%
  \trimfig{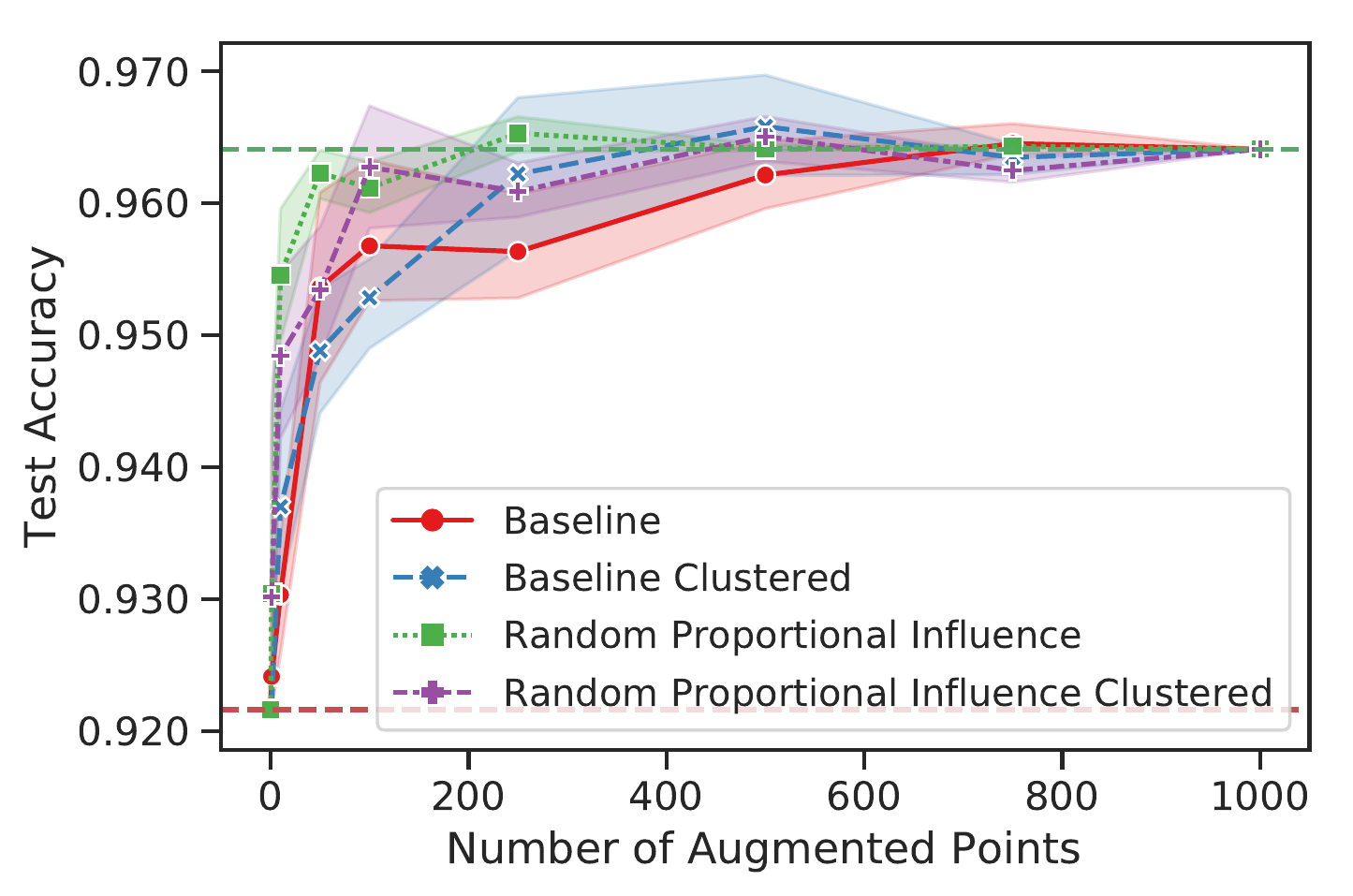}
  }%
  \newline
  \subfloat[MNIST-crop]{%
    \trimfigleft{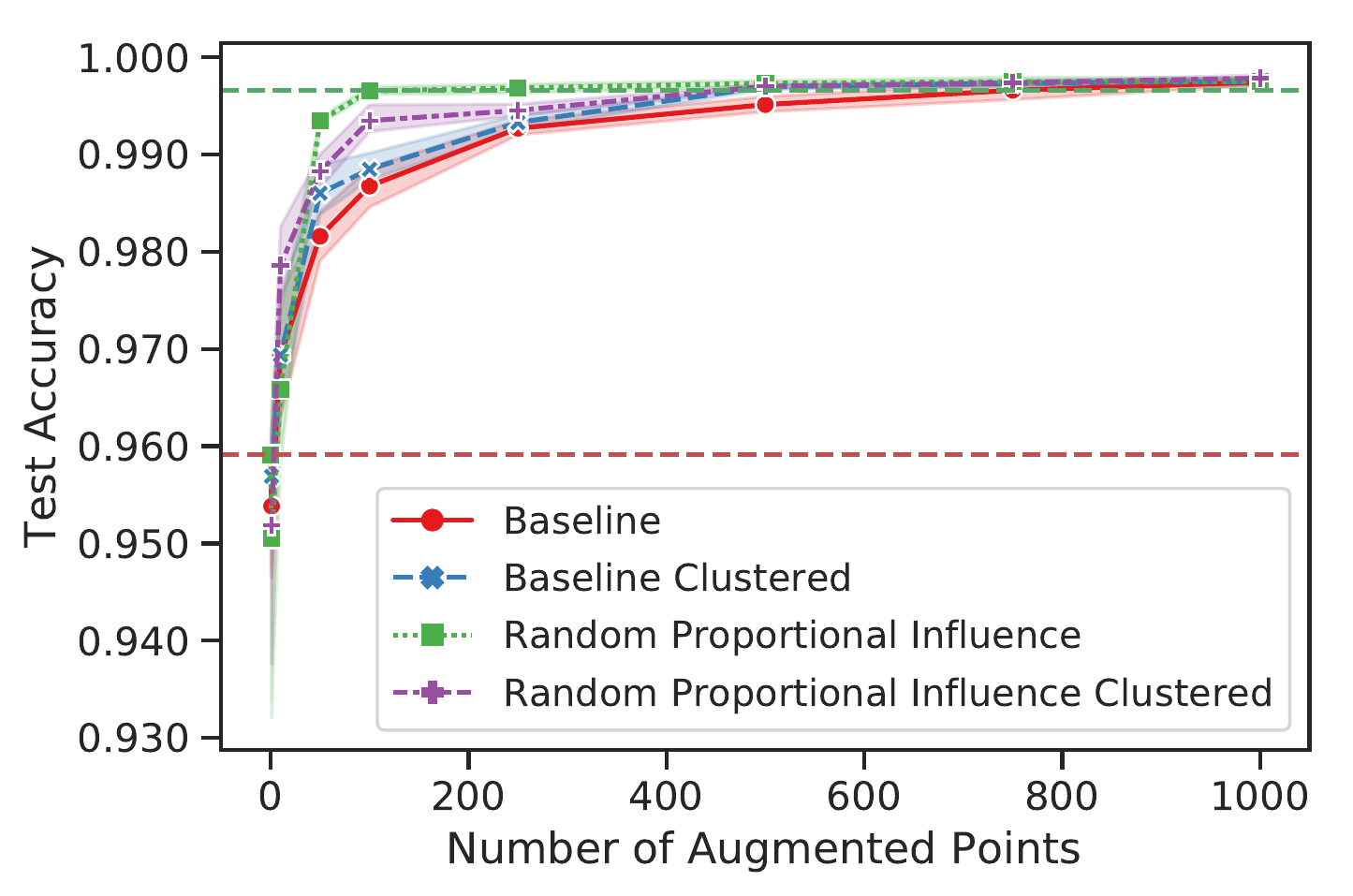}
  }%
  \subfloat[CIFAR10-crop]{%
    \trimfig{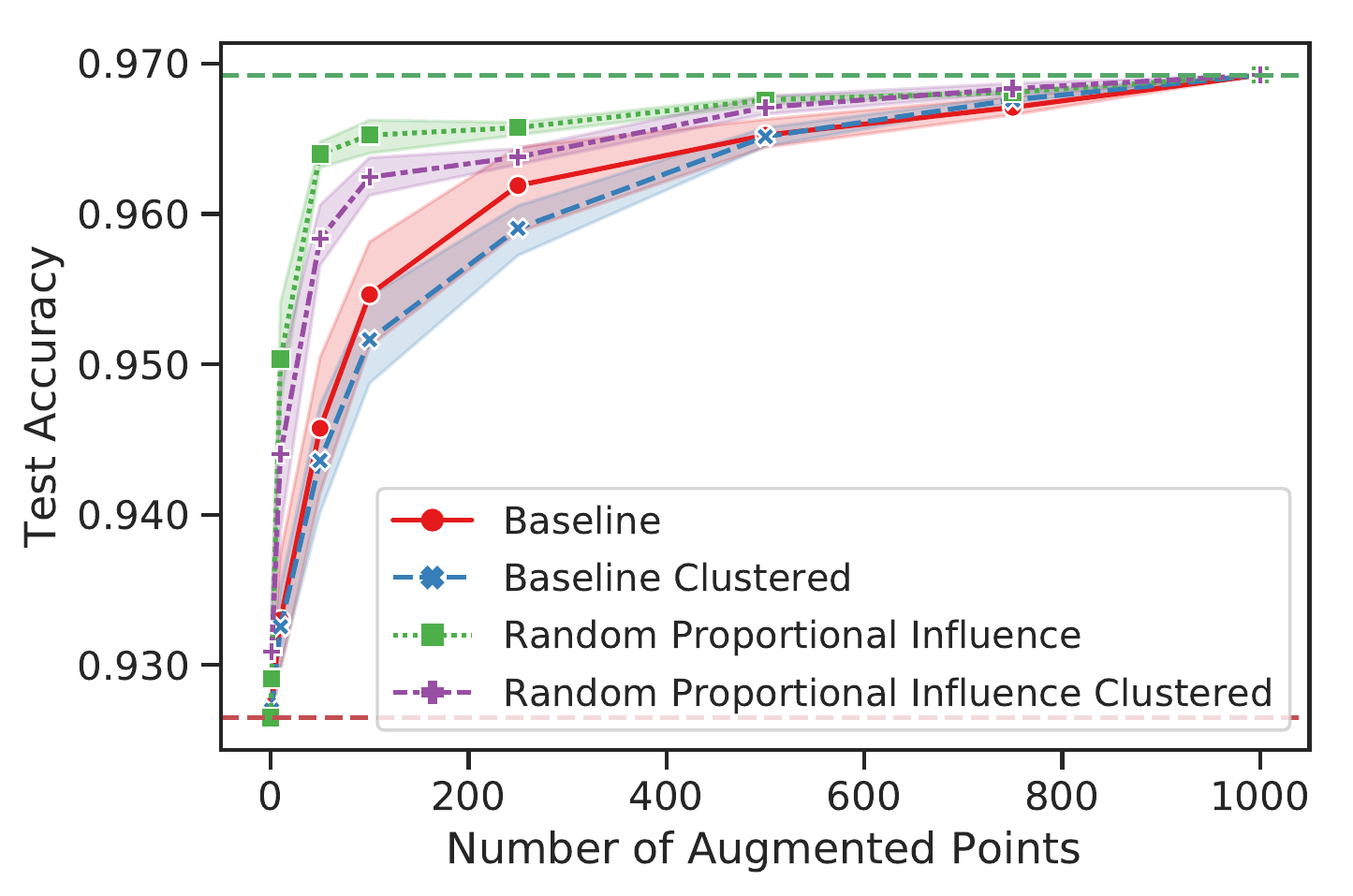}
  }%
  \subfloat[NORB-crop]{%
  \trimfig{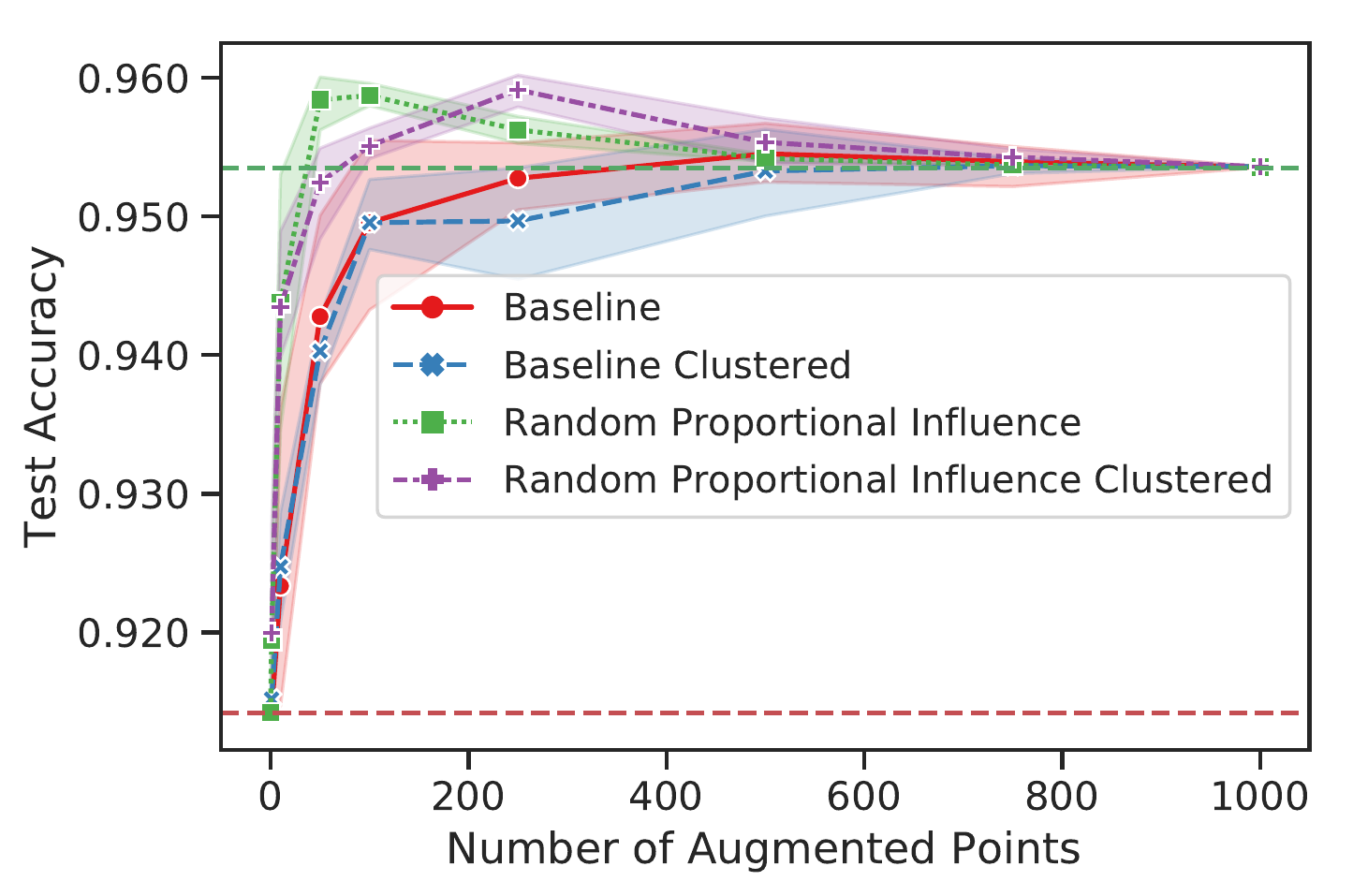}
  }%
  \caption{The performance of randomized policies using standard and clustered
  sampling.}%
\label{fig:all_mod_joined_cluster}%
\end{center}
\end{figure}

\section{Determinantal Point Processes}%
\label{sec:dpp}
One method of subsampling a dataset while encouraging data diversity is with a determinantal point process (DPP),
which is explained further in \cite{kulesza2012determinantal}.
A DPP allows for tractable subset selection with both
diversity and quality criterion.
We use a DPP to select points to augment, mirroring the methodology used
throughout this paper.
One exception is the other explored methods are amenable to greedy extensions in
the number of points augmented, while the DPP algorithm was re-run for every
change in the number of augmented points.
It is worth noting that our experiment setup is not the intended use of a DPP, as we fix the
original training set and concatenate the selected augmented points.
In contrast, a DPP would typically be used to subsample the original training set.
For our experiments, we use a publicly available
implementation\footnote{\url{http://www.alexkulesza.com/code/dpp.tgz}}.

For our tests, we use a $k$-DPP, which is a DPP conditioned on the number of
points selected being equal to $k$.
We summarize the notation given in \cite{kulesza2012determinantal} for our
construction of the DPP kernel, $L$.
We set $L_{ij} = q_i \phi_i^\top \phi_j q_j$\footnote{We experienced numerical issues computing the elementary
symmetric polynomials for high $k$, so we scale $\phi_i$ by $1000$.}, where $q_i \in \mathbb{R}^+$ is a
quality term and $\phi_i \in \mathbb{R}^D$ is a diversity term.
In our tests, we use influence for $q_i$ and bottleneck features for $\phi_i$.
We normalized each sample with $||\phi_i||^2=1$ as was suggested in
\cite{kulesza2012determinantal} to make all samples equally likely without
taking diversity into account.

\begin{figure}%
\begin{center}
    \subfloat[CIFAR10-translate]{%
  \trimfig{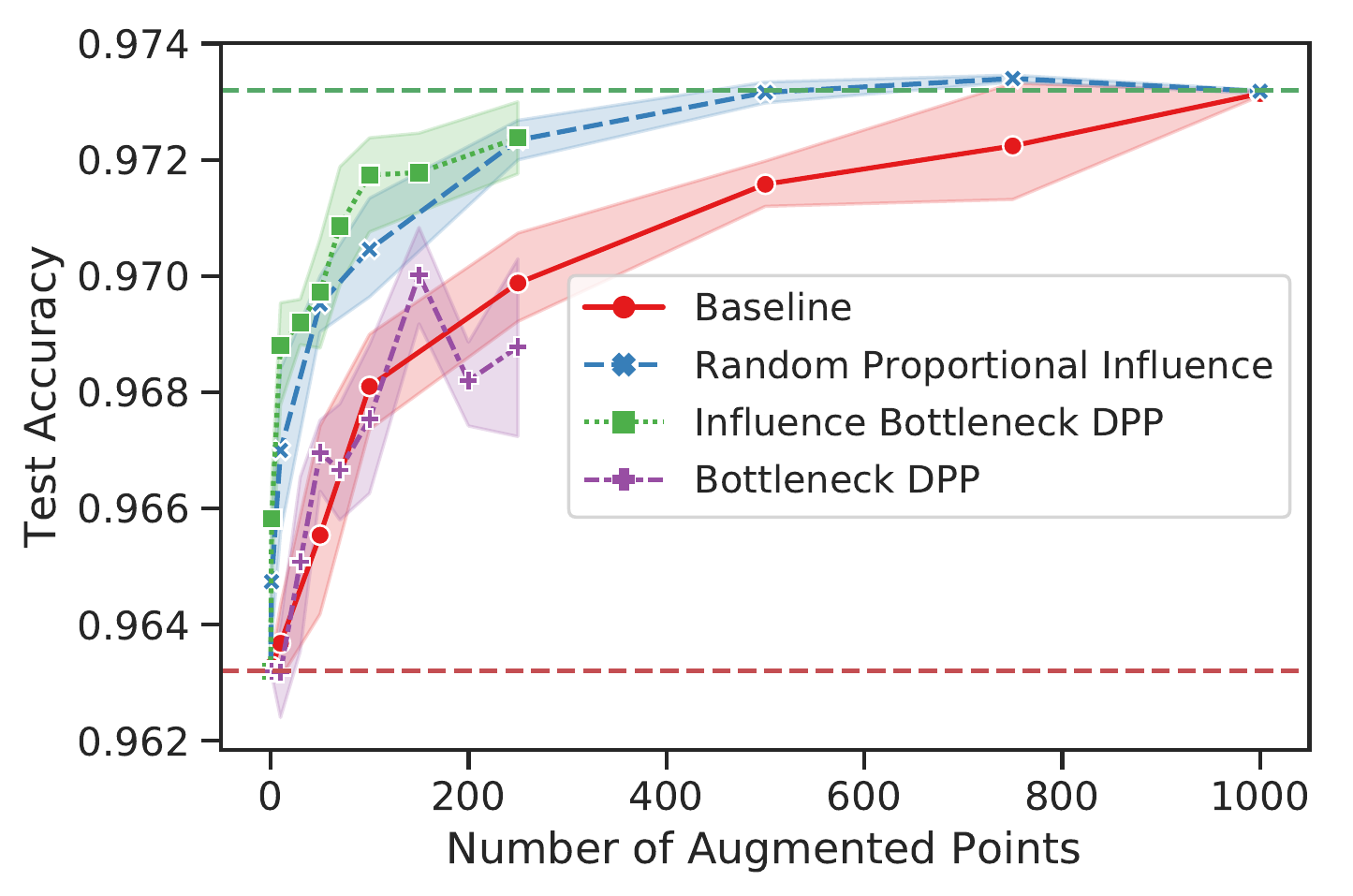}
  }%
    \subfloat[CIFAR10-rotate]{%
  \trimfig{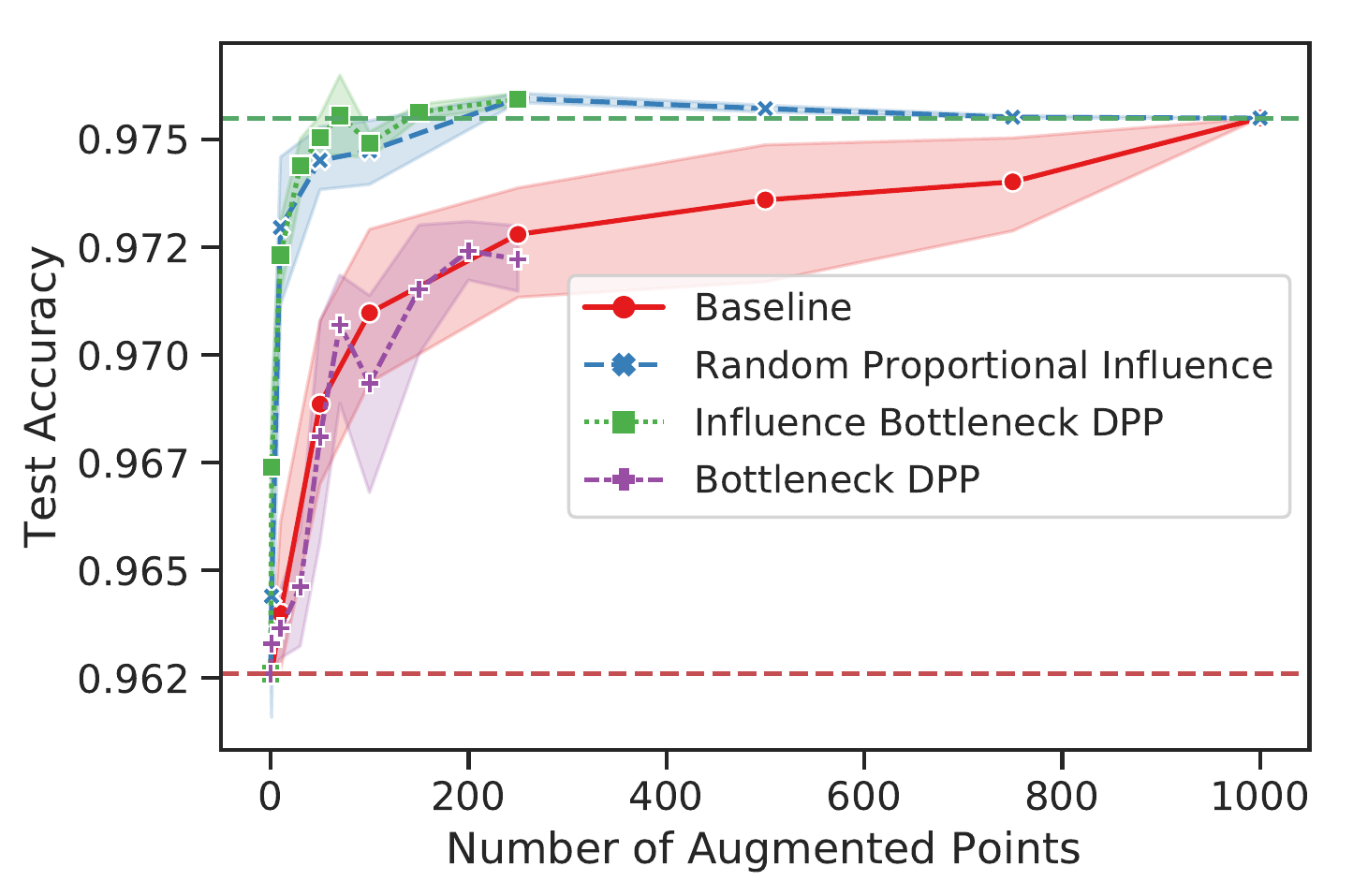}
  }%
    \subfloat[CIFAR10-crop]{%
  \trimfig{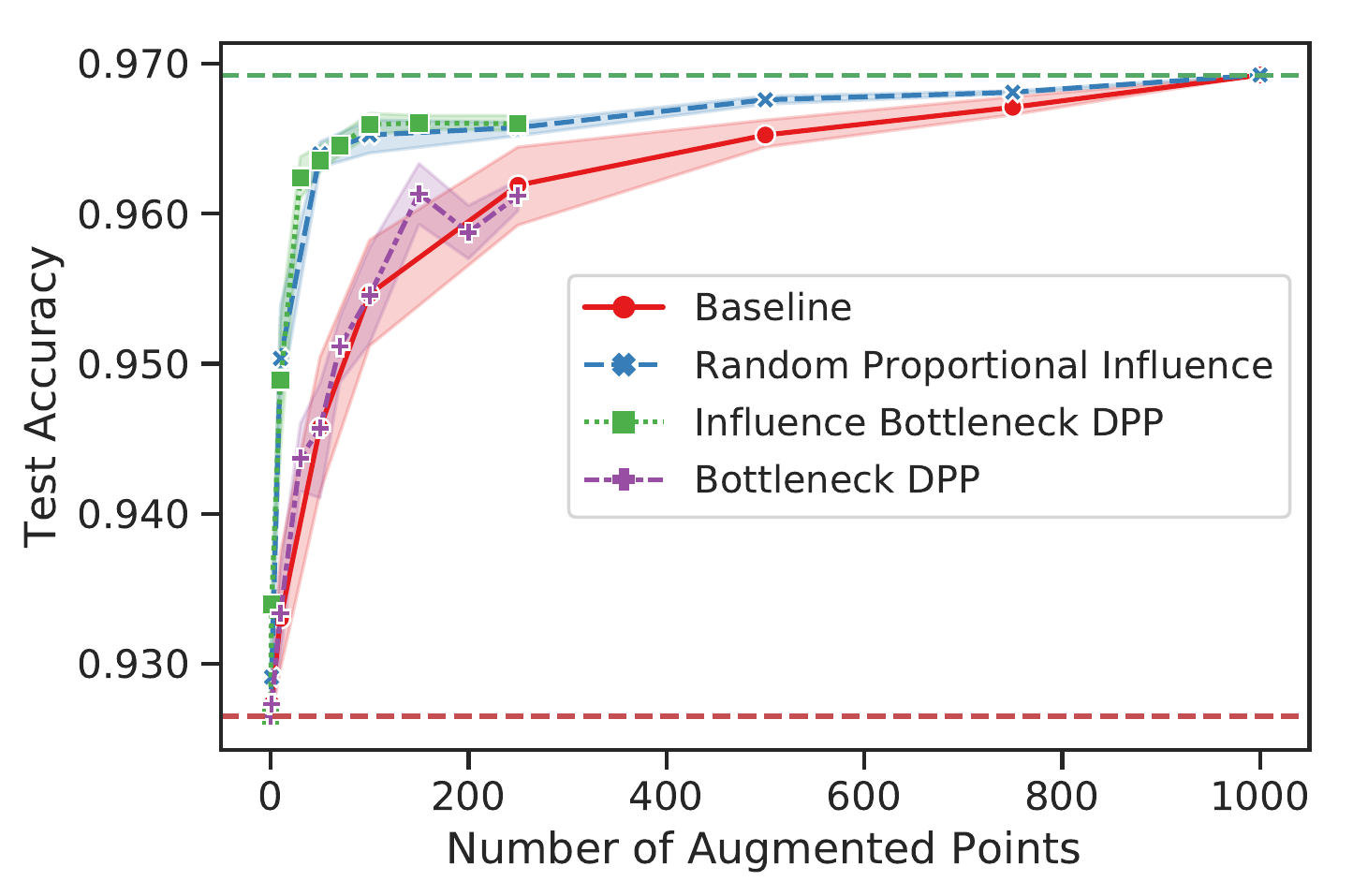}
  }%
  \caption{The performance of CIFAR10 k-DPP policies using bottleneck features or
  a combination of influence and
  bottleneck features in the kernel. Only 250 augmented points were used due to
  the computational expense of samplings a larger amount of points.}%
\label{fig:cifar_dpp}%
\end{center}
\end{figure}

The DPP results are shown in Figure~\ref{fig:cifar_dpp}.
``Influence Bottleneck DPP'' corresponds to using both influence and bottleneck
features in the kernel.
``Bottleneck DPP'' corresponds to using only bottleneck features and therefore
has no idea of the quality of a point.
For the DPP experiments, we ran up to $250$ points augmentations and re-ran the
experiments $5$ times.
The other results included in the plot are from previous experiments.
As can be seen, the influence weighted DPP performance is competitive with the
influence driven approach.
Using solely bottleneck features (i.e., $q_i=1$) for $L$ resulted in poor performance.
The reason we chose to only test up to $250$ augmented points is because
sampling a DPP takes $O(Nk^3)$, where $N$ is the size of the full set used in
the subset selection.
We found this computational performance to be limiting in practice, as $N=1000$ and $k$ approached $1000$.
For low $k$, it may make sense to use DPP methods, but for larger $k$,
alternative approaches or approximations may be required due to computational
budgets.

\section{Effects on Training Performance}%
\begin{figure}
\begin{center}
  \includegraphics[width=0.48\linewidth]{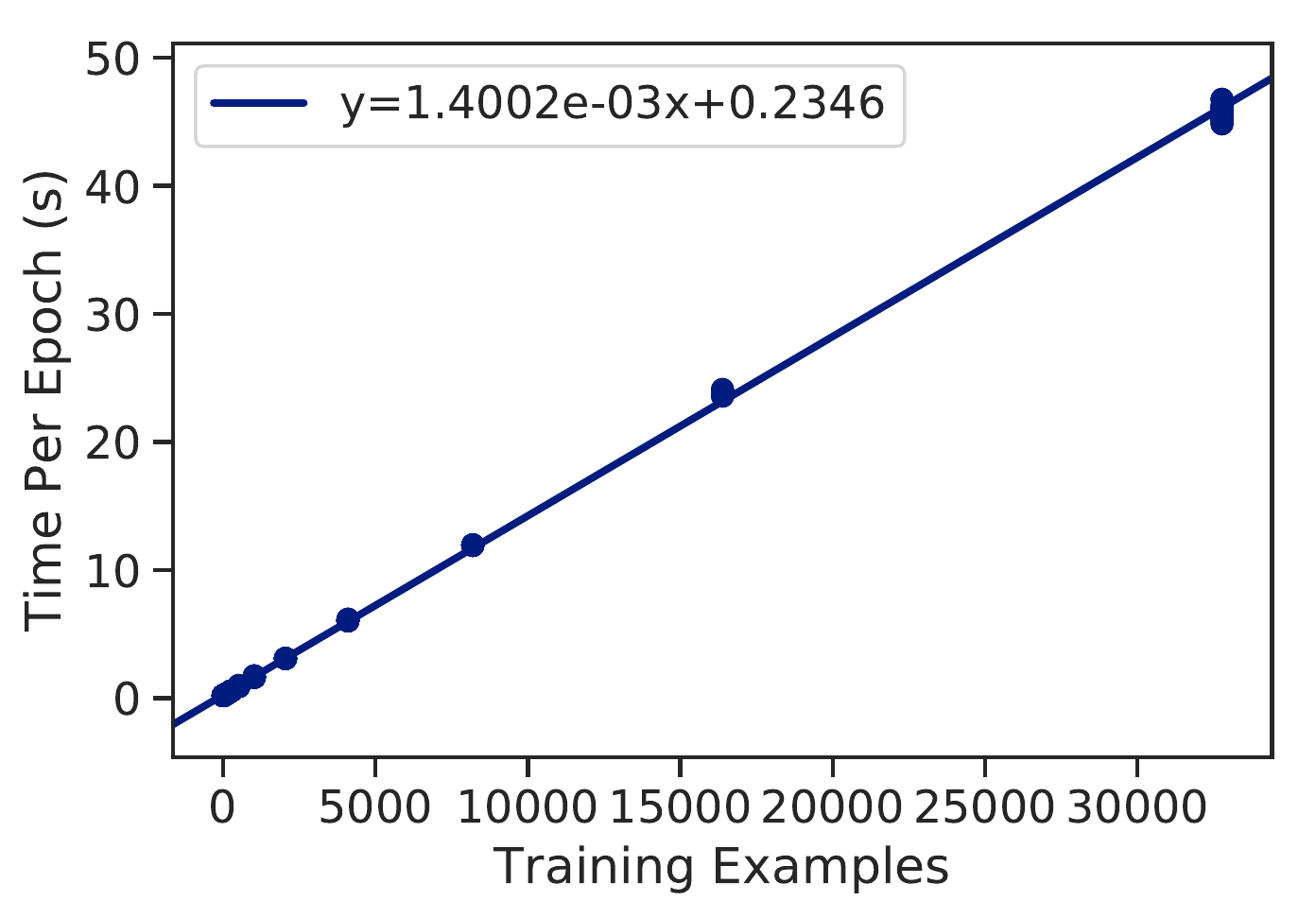}
  \caption{The time it takes (in seconds) to perform a single ResNet50v2 epoch with respect to training
  set size. The training relationship is linear with low variance.}%
  \label{fig:cifar_train_times}%
\end{center}
\end{figure}

In the context of deep learning, we expect performance to scale linearly with an increased dataset.
To highlight
this effect, we train a ResNet50v2 network \citep{he2016identity} using Tensorflow
\citep{tensorflow} version 1.10.1 with a
variable number of training examples obtained from CIFAR10.
The system which was used for the test has an Intel i7-6700k %
and an Nvidia GTX
1080 using CUDA 9.2 and CuDNN 7.2.1.
We ran the test $5$ times to control for variance.
We show the resulting scaling performance in Figure~\ref{fig:cifar_train_times}.
Little deviation was observed from the linear fit.

We can see that the scaling is indeed linear.
If we assume that the number of epochs is fixed, then we can conclude
subsampling would result in a linear decrease in training time.
We observe decreases in training time in our MNIST experiments, where we continously
retrain our feature-extraction model with variable amounts of augmented data.
In a more complicated training regime, such as distributed training, we can
expect this improvement to be greater.
Although it is true that the initial model training may reduce some of these
performance
benefits, it is worth noting that a pretrained model may be used to bootstrap a
different model, and the cost of the initial training can be amortized over many
experiments.
For the CIFAR10 translate task, we tried using the SVM support vectors to select
points to augment for a ResNetv2 model. Our training used a fraction of the
baseline method's total training data (1296/500), but achieved within $5\%$ of the
test accuracy.
Similar trends were observed in other tasks.
Augmenting a subset of the training set has the potential to decrease training
time without significantly compromising model performance.

\end{document}